\documentclass{ieeeaccess}
\usepackage{cite}
\usepackage{amsmath,amssymb,amsfonts}
\usepackage{graphicx}
\usepackage{textcomp}

\usepackage{bm}

\usepackage{adjustbox}
\usepackage{mathtools}
\usepackage{multirow}%

\usepackage{amsthm}%
\usepackage{mathrsfs}%

\usepackage{manyfoot}%
\usepackage{booktabs}%
\usepackage{algorithm}%
\usepackage{algorithmicx}%
\usepackage{algpseudocode}%
\usepackage{listings}%
\usepackage{booktabs}
\usepackage{tabularx}
\usepackage{caption}
\usepackage{multirow}
\usepackage{subcaption}%
\usepackage{float}
\usepackage{pdflscape}

\usepackage{colortbl}
\usepackage{url}
\usepackage{soul}
\usepackage{xcolor}
\usepackage{threeparttable}

\definecolor{best}{RGB}{255,200,200}     
\definecolor{second}{RGB}{255,225,180}   
\definecolor{third}{RGB}{255,245,200}    

\usepackage[percent]{overpic}
\makeatletter
\AtBeginDocument{\DeclareMathVersion{bold}
\SetSymbolFont{operators}{bold}{T1}{times}{b}{n}
\SetSymbolFont{NewLetters}{bold}{T1}{times}{b}{it}
\SetMathAlphabet{\mathrm}{bold}{T1}{times}{b}{n}
\SetMathAlphabet{\mathit}{bold}{T1}{times}{b}{it}
\SetMathAlphabet{\mathbf}{bold}{T1}{times}{b}{n}
\SetMathAlphabet{\mathtt}{bold}{OT1}{pcr}{b}{n}
\SetSymbolFont{symbols}{bold}{OMS}{cmsy}{b}{n}
\renewcommand\boldmath{\@nomath\boldmath\mathversion{bold}}}
\makeatother

\def\BibTeX{{\rm B\kern-.05em{\sc i\kern-.025em b}\kern-.08em
    T\kern-.1667em\lower.7ex\hbox{E}\kern-.125emX}}

\begin{document}
\doi{10.1109/ACCESS.2024.0429000}

\title{Unsupervised Methods for Video Quality Improvement: A Survey of Restoration and Enhancement Techniques}
\author{
\uppercase{Alexandra Malyugina}\authorrefmark{1},
\uppercase{Yini Li}\authorrefmark{1},
\uppercase{Joanne Lin}\authorrefmark{1}, and
\uppercase{Nantheera Anantrasirichai}\authorrefmark{1}
}

\address[1]{Visual Information Laboratory, University of Bristol, Merchant Venturers Building, Woodland Road, Bristol BS8 1UB, UK (e-mails: alex.malyugina@bristol.ac.uk; ub24017@bristol.ac.uk; joanne.lin@bristol.ac.uk; n.anantrasirichai@bristol.ac.uk)}
\tfootnote{This work has been funded by the EPSRC NIA (UKRI241).}

\markboth
{{A. Malyugina }\headeretal: {Unsupervised Methods for Video Quality Improvement: A Survey of Restoration and Enhancement Techniques}}
{{A. Malyugina }\headeretal: {Unsupervised Methods for Video Quality Improvement: A Survey of Restoration and Enhancement Techniques}}

\corresp{Corresponding author: Alexandra Malyugina (e-mail: alex.malyugina@bristol.ac.uk).}

\begin{abstract}
Video restoration and enhancement are critical not only for improving visual quality but also as essential pre-processing steps to boost the performance of a wide range of downstream computer vision tasks. We present an in-depth review of video restoration and enhancement techniques, with a particular focus on unsupervised approaches. {Throughout this survey, we define unsupervised learning broadly to include self-supervised and zero-shot approaches.} We begin by outlining the most common video degradations and their underlying causes, followed by a review of early conventional and deep learning-based methods, highlighting their strengths and limitations. We then present an in-depth overview of unsupervised methods, categorized by their fundamental approaches, including domain translation, self-supervision signal design, {consistency-based methods, degradation-aware methods, and prior-based methods}. We also propose a unified multi-dimensional taxonomy that factorizes unsupervised architectures across three axes: the supervision signal, temporal modeling strategy, and learning paradigm. Additionally, we provide a categorization of loss functions employed in unsupervised video restoration and enhancement and discuss the role of paired synthetic datasets in enabling objective evaluation. Finally, we present the current performance of key applications, identify open challenges, and outline promising directions for future research in this field.
\end{abstract}


\titlepgskip=-21pt

\maketitle

\section{Introduction}

Videos captured under real-world conditions often exhibit various degradations, including poor illumination, motion blur, out-of-focus, noise, compression artifacts, low contrast, flicker, and temporal inconsistencies. These degradations reduce visual quality and, critically, impair downstream computer vision tasks such as object tracking, recognition, segmentation, optical flow, and depth estimation, potentially resulting in false negatives, misclassifications, or system failures, as reported in \cite{Yao:video:2020,Yi:Comprehensive:2024,hill2025automatic}. Consequently, video quality enhancement has become a critical research focus, especially for footage acquired in uncontrolled or adverse environments.

Broadly, two related but distinct objectives are commonly addressed in this context: \emph{restoration} and \emph{enhancement}. Restoration focuses on recovering a clean video from its degraded counterpart by assuming a known or estimated corruption model, such as sensor noise, motion blur, compression artifacts, or missing data. In contrast, enhancement aims to improve the perceptual quality of video, adjusting attributes like dynamic range, contrast, color balance, or sharpness, without requiring explicit knowledge of the degradation process.
Restoration methods typically prioritize fidelity to an undistorted signal, while enhancement techniques focus on improving visual clarity and aesthetic appeal, which are often guided by more subjective criteria \cite{AJAFERNANDEZ2024145, Anantrasirichai:AI:2022}. In practice, the boundary between these two categories is often obscure: enhancing contrast may incidentally reduce blur, and deblurring may improve perceptual sharpness. Consequently, many modern methods are designed to address both tasks simultaneously, particularly in scenarios where degradations are complex or poorly characterized.

Traditionally, both restoration and enhancement tasks have been tackled using handcrafted filters and domain-specific priors \cite{celebi2015color, yu2020image}. More recently, learning-based approaches, particularly those employing deep neural networks, have achieved significant progress in a wide range of \textit{image} restoration and enhancement tasks. Many such tools are now available in both open-source and commercial software \cite{anantrasirichai2025artificial}. On the other hand, \textit{video}-based AI restoration and enhancement methods are comparatively underdeveloped. Their advancement is restricted by a larger volume of unknown parameters, a shortage of training data, and substantial computational and memory requirements. A common workaround involves applying image-based models frame-by-frame, but this often introduces temporal inconsistencies \cite{Zhang:learning:2021}. Furthermore, the majority of existing methods for both image and video quality enhancement rely on paired training data, with degraded and reference sequences aligned spatially and temporally. Acquiring such data is expensive and often impractical in real-world scenarios such as surveillance, low-light natural history recordings, and historical archives, where high-quality ground truth is unavailable or cannot be reliably reconstructed.

This has motivated growing interest in unsupervised approaches that learn from unpaired or unlabeled data by leveraging inherent structures in video, such as temporal consistency \cite{cao2025zeroshot}, spatial redundancy \cite{Huang_2021_CVPR}, motion coherence \cite{liu2025appearance}, and statistical priors \cite{Wang_2019_CVPR}. Without relying on explicit ground truth, unsupervised methods are particularly well suited for domain-adaptive video quality enhancement, where degradations are complex, unknown, or data-specific.

While numerous surveys exist on \textit{image} restoration and enhancement, dedicated reviews of \textit{video}-specific approaches remain limited. Existing works on video restoration~\cite{rota2023video}, low-light video enhancement~\cite{Li:low:2022}, and underwater video enhancement~\cite{hu2022overview} {and recent comprehensive reviews on self-supervised image denoising \mbox{\cite{zhang2026unleashing}} primarily focus on supervised frameworks, specific loss functions, or are strictly confined to image-based methodologies.} Critically, these surveys overlook the expanding literature on unsupervised and weakly supervised video restoration and enhancement. Our paper aims to fill that gap by providing a focused review of methods that operate without paired ground truth, which is an increasingly important paradigm for real-world applications where annotated video data is difficult, costly or impractical to obtain.

\begin{figure}[ht!]
    \centering
    \includegraphics[width=\linewidth]{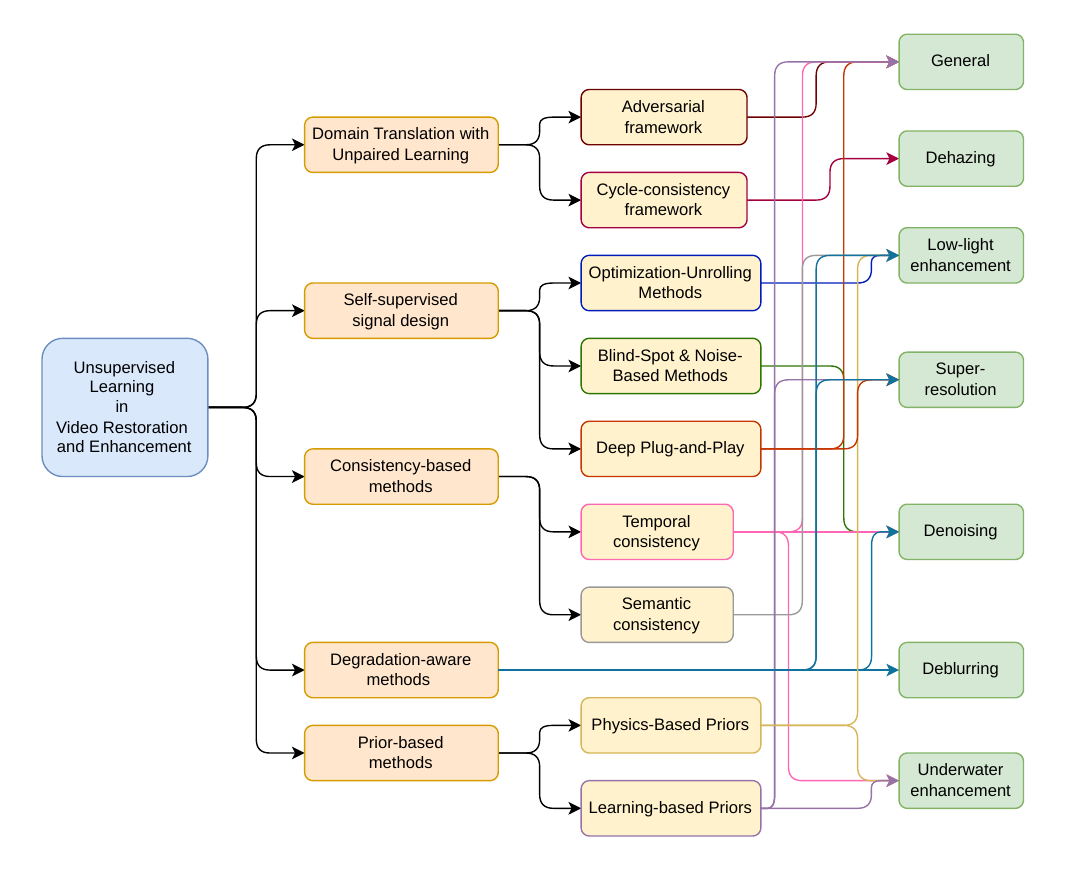}
    \vspace{1pt}
    \caption{Hierarchical taxonomy of unsupervised, self-supervised, and zero-shot learning categories for video restoration and enhancement. The diagram illustrates the methodological framework discussed in Section \ref{sec:unsupervised}, categorizing the literature into five primary strategies. In the final stage of the hierarchy, each arrow represents a specific method discussed in the survey and the colors of the arrows denote methodological subclasses. {Note that unlike other high-level strategies, degradation-aware methods are not further subdivided, as they uniformly rely on the singular analysis-by-synthesis principle to derive self-supervision.}}
    \label{fig:hierarchical_taxonomy}
\end{figure}

In this survey, we present a structured overview of unsupervised methods for video restoration and enhancement. To provide context, we first briefly review classical and supervised approaches. We first introduce a high-level, strategy-centric taxonomy that groups current literature into five primary methodological categories based on their core operational strategies (as illustrated in Fig. \ref{fig:hierarchical_taxonomy}). To address the taxonomic overlap inherent in complex modern networks, we subsequently propose a unified multi-dimensional taxonomy that factorizes these architectures across three orthogonal design axes: the supervision signal ($S$), temporal modeling strategy ($T$), and learning paradigm ($L$) (as shown in Fig. \ref{fig:multidim_taxonomy} and Table \ref{tab:video_taxonomy}). In Section~\ref{sec:loss}, we further summarize and categorize the loss functions commonly employed in unsupervised video restoration and enhancement. Section~\ref{sec:syndata} describes the synthetic datasets used for the objective evaluation of unsupervised methods and discusses the associated challenges and open research questions. We then analyze and compare the performance of unsupervised and supervised approaches in Section~\ref{sec:performance}. Finally, Section~\ref{sec:challenges} outlines the remaining challenges and open problems in this field. Our aim is to provide a critical synthesis of the existing literature and to establish a comprehensive foundation for future research on video restoration and enhancement under unpaired supervision.

\subsubsection*{Literature Search Methodology}
{We adopt an integrative approach to synthesize major developments in unsupervised video restoration and enhancement. We selected literature published between 2017 and early 2026 from leading computer vision and machine learning conferences, such as CVPR, ICCV, ECCV, NeurIPS, AAAI, IEEE TPAMI, IEEE TIP. Search keywords included "unsupervised video restoration," "self-supervised video denoising/deblurring/dehazing," "zero-shot video enhancement," and "unpaired video enhancement". Source selection prioritized technical originality, relevance to unpaired learning, and impact on specific degradation tasks. After initially screening approximately 
800 
documents, we retained over 
200 
sources for analysis and explicitly classifying 21 representative methods within our proposed multi-dimensional taxonomy.
}

\section{Basics of Video Restoration and Enhancement}
\label{sec:basics}

Video restoration and enhancement are inherently ill-posed inverse problems, often admitting multiple valid solutions due to information loss during capture. These tasks become particularly challenging when attempting to reconstruct the original content from severely degraded observations. A common formulation expresses degradation as 
\begin{equation}
    I_\text{obv} = H I_\text{idl} + n,
    \label{eqn:degradationmodel}
\end{equation}
 where $I_\text{idl}$ denotes the ideal (clean) signal, $I_\text{obv}$ is the observed video, $H$ represents an unknown distortion function, and $n$ is additive noise. Many existing methods simplify this by assuming a space-invariant point spread function for $H$ and Gaussian zero-mean noise for $n$. However, such assumptions fail to generalize to real-world scenarios involving complex environmental degradation, e.g., low-light conditions, dynamic non-stationary media, light scattering and absorption, which lead to a mix of blur, noise, color distortion, halo artifacts, and contrast reduction.

Moreover, real-world video restoration is not only ill-posed but also blind, as the degradation model and parameters are typically unknown. The noise characteristics, blur kernels, and degradation severity are often uncertain, necessitating blind restoration techniques that jointly estimate both the clean content and the underlying corruption model. This results in highly non-convex optimization problems, further compounding the difficulty of accurate video enhancement in uncontrolled environments.



\subsection{Common Degradation Problems in Videos}
Real-world videos frequently exhibit a range of artifacts resulting from different types of degradation, necessitating enhancement or restoration. In this subsection, we discuss the most common degradation types encountered in practice: blur, noise, low contrast, and low-light conditions.

\subsubsection{Blur} Videos captured with moving cameras or objects often exhibit \textit{motion blur}, particularly under long exposure times or high-speed movement. In video, motion blur manifests as smearing along moving edges and typically varies across frames. It is often temporally uncorrelated between successive frames \cite{su2017deep}. Common causes include handheld camera shake, which results in global blur, and fast-moving objects, leading to local blur. Unlike still-image blur, video blur benefits from temporal redundancy: adjacent frames are often blurred differently, allowing sharp details in one frame to compensate for losses in another. This property underpins many video deblurring strategies \cite{xiang2025deep}.

\begin{figure}[t]
    \centering
    \begin{minipage}[t]{0.47\linewidth}
    \centering
    \includegraphics[height=4cm]{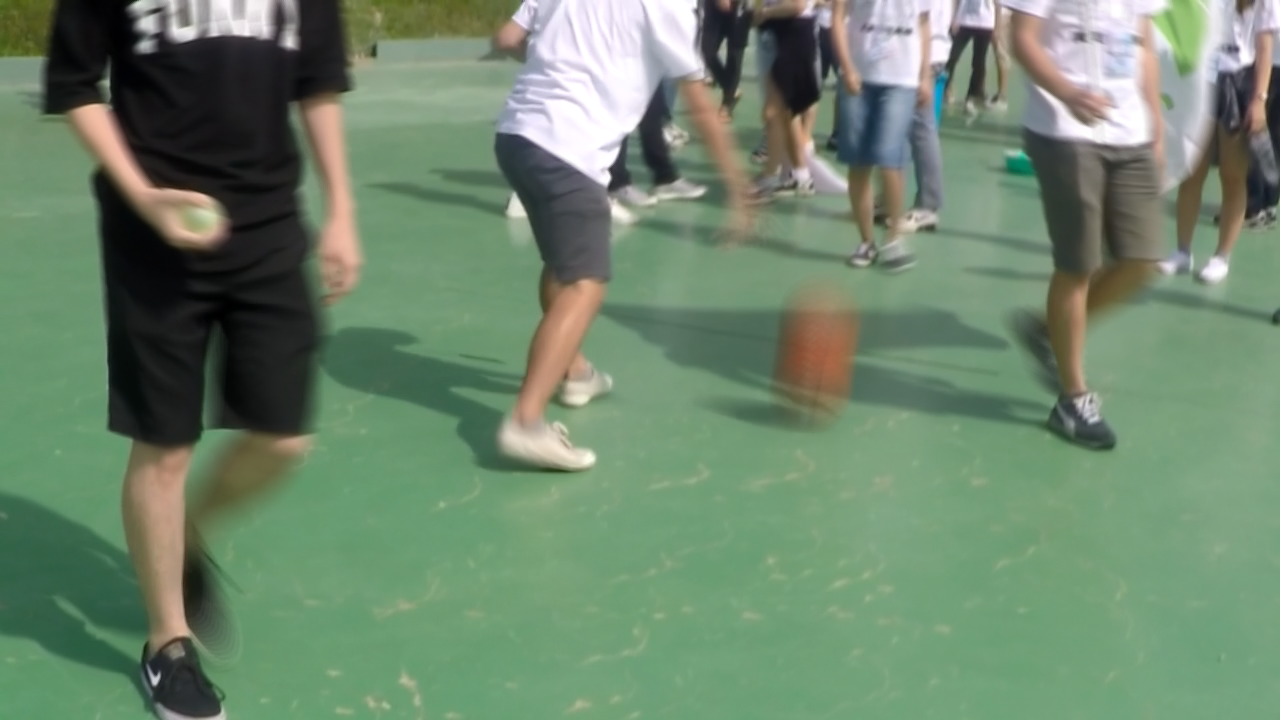}
    \caption{A video sequence fragment from GoPro Dataset \cite{Nah_2017_CVPR}, showing regions affected by motion blur}
    \label{fig:motion}
    \end{minipage}
    \hfill
    \begin{minipage}[t]{0.42\linewidth}
    \centering
    \includegraphics[height=4cm]{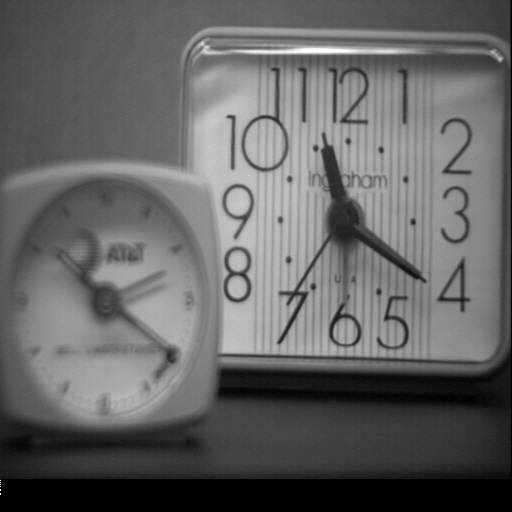}

    \caption{An example of out-of-focus blur}
    \label{fig:outfocus}
    \end{minipage}
\end{figure}

\textit{Defocus blur} (or \textit{out-of-focus blur}) arises when scene elements fall outside the camera’s depth of field, resulting in a loss of sharpness \cite{Morris:dabit:2025}. \textit{Atmospheric blur}, induced by air turbulence resulting from temperature variations across regions, distorts light propagation, particularly over long distances, and is commonly encountered in outdoor surveillance \cite{hill2025deep} and remote sensing applications \cite{richards2022remote}. Example of motion blur  and out-of-focus blur are shown in Fig. \ref{fig:motion} and Fig. \ref{fig:outfocus}, respectively.

\subsubsection{Noise} 
\begin{figure}[t]
    \centering
    \includegraphics[width=\linewidth, trim=0 1170 0 0, clip]{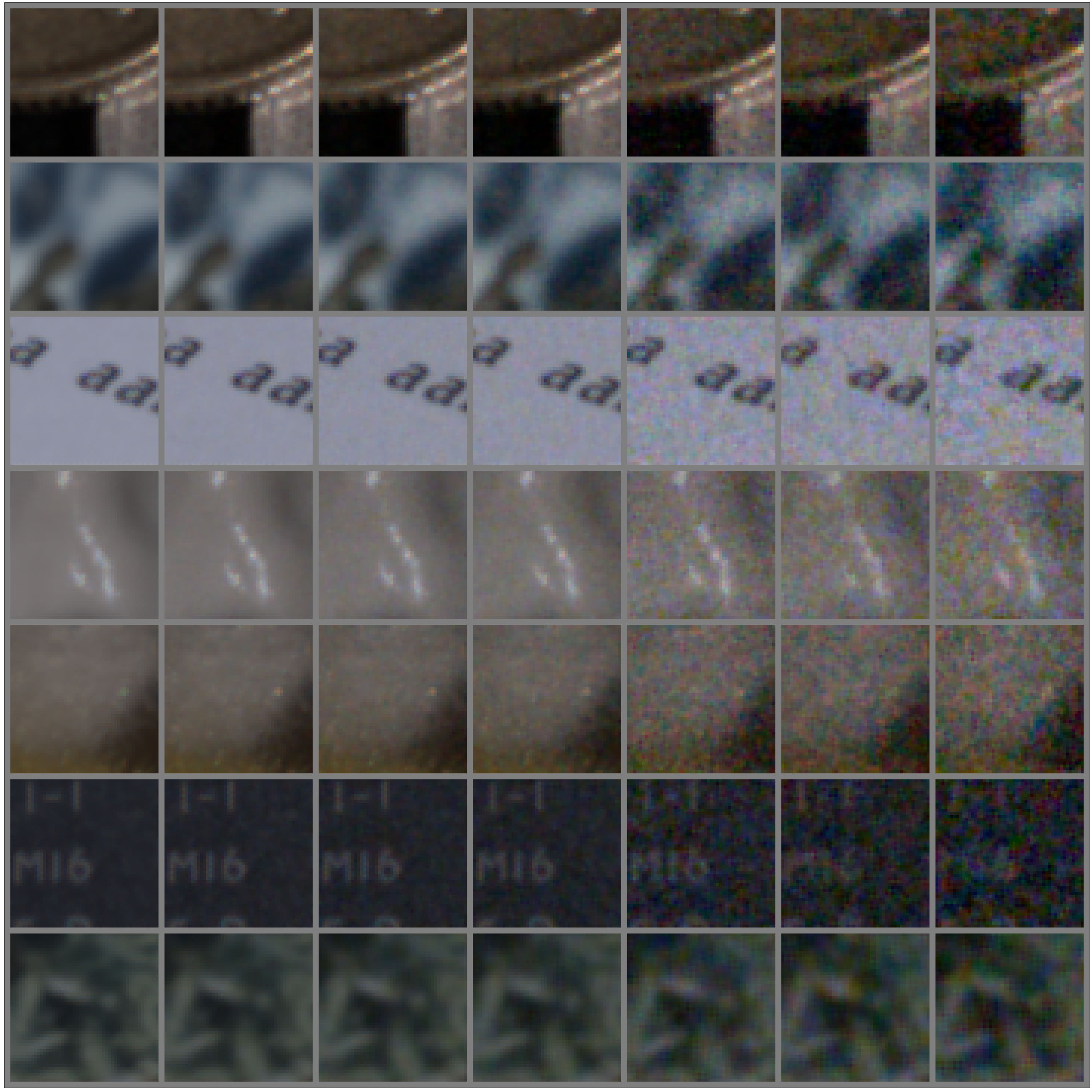}
    \caption{Examples of sensor noise at different ISO levels (ISO = 1600, 6400, 12800, 64000, 102400, 160000); image patches from \cite{malyugina2024topological}}
    \label{fig:noise}
\end{figure}

In digital images and videos, noise manifests as random variations in pixel intensity or color, leading to degradation of fine details. In video, electronic noise is particularly prevalent, primarily originating from the camera sensor and associated circuitry. Typical sources include photon shot noise, read noise, and dark current noise \cite{blanksby1997noise}, along with quantization noise (from analog-to-digital conversion) and transmission noise (from lossy or noisy communication channels). Noise is often modeled using a Gaussian distribution, or a Poisson–Gaussian mixture to account for both photon and electronic contributions \cite{foi2008practical}. Under well-lit conditions, noise tends to be minimal. However, in low-light environments, cameras increase sensor gain (e.g., through higher ISO settings), which amplifies both the signal and noise, often resulting in significantly noisier output. In video sequences, noise can appear as high-frequency grain, stemming from photon shot noise, as well as banding or artifacts introduced by readout and compression. Unlike motion blur, noise is typically temporally uncorrelated, assuming no in-camera temporal averaging or denoising. Examples of noisy patches are shown in Fig.~\ref{fig:noise}, where the right images were captured under lower-light conditions. To maintain a consistent brightness level, higher ISO settings were applied, which introduced additional noise. Consequently, the darker scenes exhibit significantly higher noise due to the increased ISO sensitivity required in low-light environments.

\subsubsection{Continuous Scattering Degradations} 
\begin{figure}[t]
    \centering
    \begin{minipage}[t]{0.47\columnwidth}
    \centering
    \includegraphics[height=2.7cm, trim={0 3cm 0 1cm}, clip]{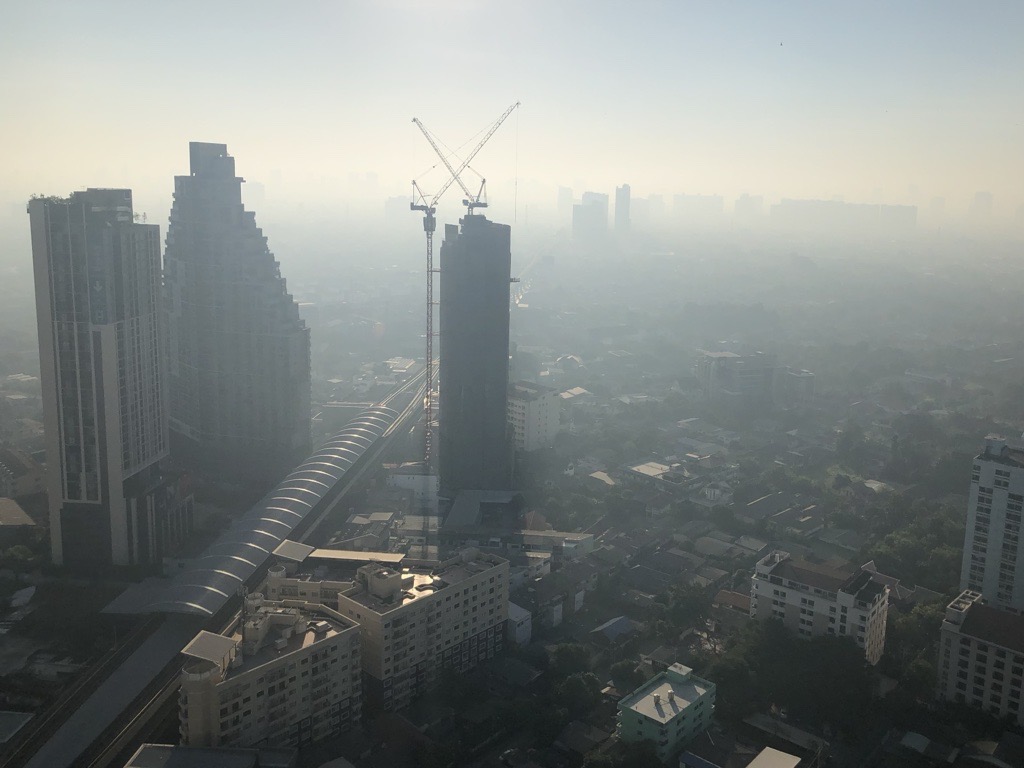}
    \caption{Low contrast and grey veil (haze) caused by atmospheric particles.}
    \label{fig:haze}
    \end{minipage}
    \hfill
    \begin{minipage}[t]{0.5\columnwidth}
    \centering
    \includegraphics[height=2.7cm, trim={12cm 0 0 0}, clip]{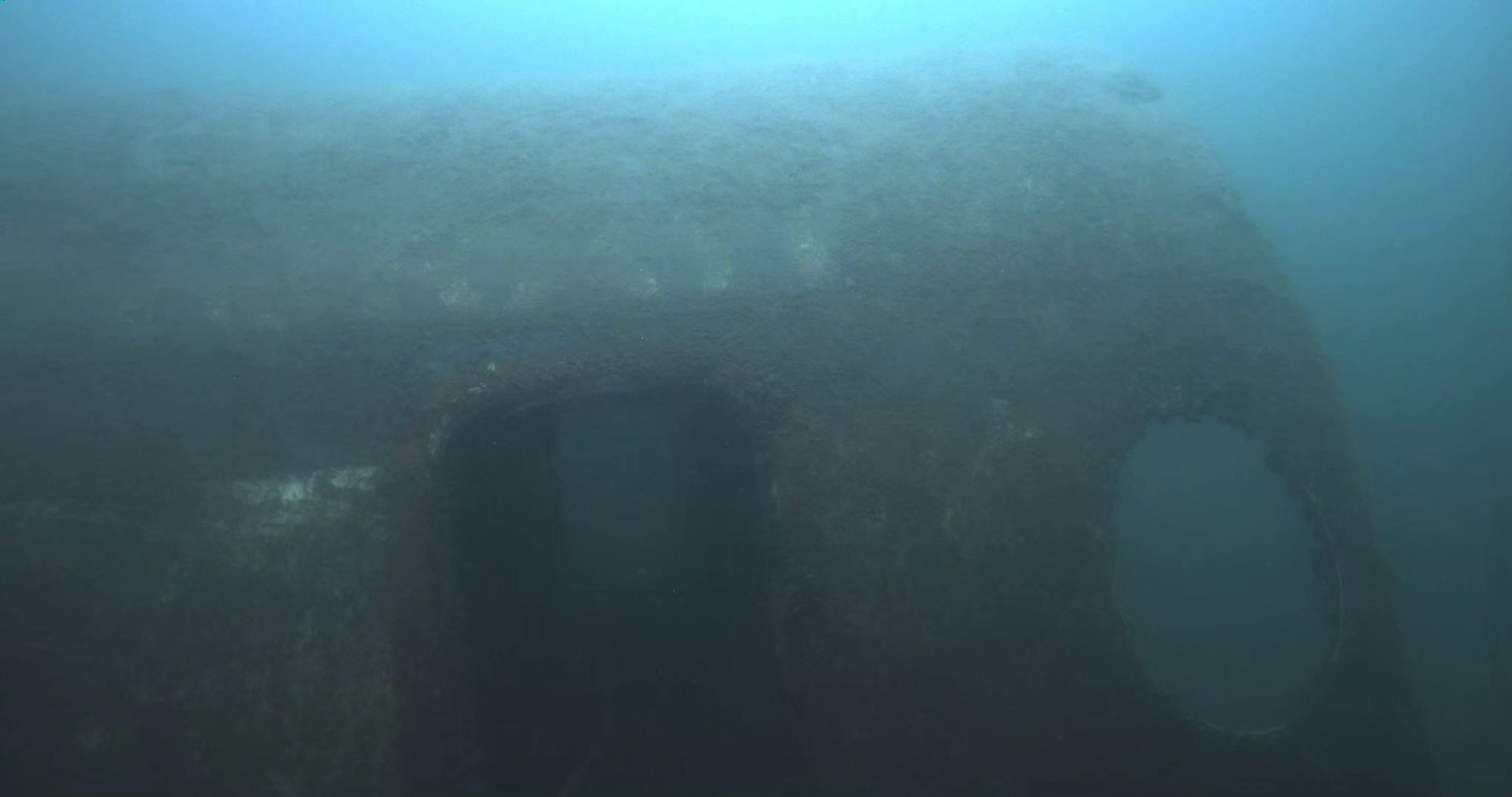}

    \caption{Low contrast and blurry underwater footage}
    \label{fig:underwater}
    \end{minipage}
\end{figure}

Outdoor videos captured in fog, mist, or polluted environments suffer from two primary phenomena \cite{nayar1999vision}: \textit{attenuation}, the loss of scene radiance due to absorption, and \textit{airlight}, the additive scattered light from the atmosphere. Haze in digital images and videos is caused by suspended particles in the air, such as dust, smoke, and microscopic water droplets. These particles absorb and scatter light rays traveling from scene objects to the camera \cite{GOYAL2024102151}. This process reduces contrast and alters color perception, particularly over long distances.
In fluid environments, similar degradation is modeled by the Jaffe–McGlamery model \cite{jaffe1990computer, mcglamery1980computer}, which accounts for both absorption and scattering in participating media such as seawater. This model decomposes the captured image into three components: direct transmission, forward scattering, and backscattering. The scattering introduces a veiling luminance (or veiling light) that overlays the scene, obscuring textures and fine details. Consequently, hazy videos exhibit reduced visibility, loss of texture in distant regions, and desaturated colors. In underwater videos, these effects are further compounded by wavelength-dependent light attenuation, which causes severe color imbalance due to the differential absorption of red, green, and blue wavelengths, resulting in a strong bluish or greenish cast.
The classic degradation model is shown as described in Eq. 
\ref{eqn:light} \cite{kim2013optimized}: 
\begin{equation}
I(x) = J(x)\,t(x) + A\,[1 - t(x)],
\label{eqn:light}
\end{equation}
where $I$ is the observed image intensity, $J$ the true scene radiance (i.e., the desired output), $A$ the global atmospheric light (ambient veil color), which is typically assumed to be constant, and $t(x)$ the transmission map, representing the portion of scene radiance that reaches the camera after attenuation by atmospheric particles. The transmission map generally decreases with increasing scene depth. Dehazing involves estimating both $A$ and $t(x)$ in order to recover the clean image $J(x)$. In video, haze can vary over time due to changes in weather conditions, illumination intensity and direction, or camera motion (e.g., moving into or out of a foggy region). Video degradation, caused by haze and underwater, is depicted in Fig. \ref{fig:haze}  and Fig. \ref{fig:underwater}, respectively.

\subsubsection{Transient Particulate Degradations} This refers to 
degradations caused by discrete, moving particles that pass through the camera’s line of sight only briefly. For example, rain and snow are weather-induced degradations that, unlike haze, involve transient particles that cause visible streaks, flicker, or occlusions across video frames. While haze is primarily caused by the scattering of light due to aerosols suspended in the atmosphere, rain and snow degrade video quality through the direct interaction of falling particles with the camera's line of sight. These effects are dynamic, spatially non-uniform, and temporally inconsistent, making them distinct from haze and often addressed separately in the literature \cite{ren2017video}. In contrast, haze generally produces a veiling glare that reduces contrast, biases color, and obscures fine scene details.

\subsubsection{Complex Degradations} 
\begin{figure}[t]
    \centering
    \includegraphics[width=\linewidth]{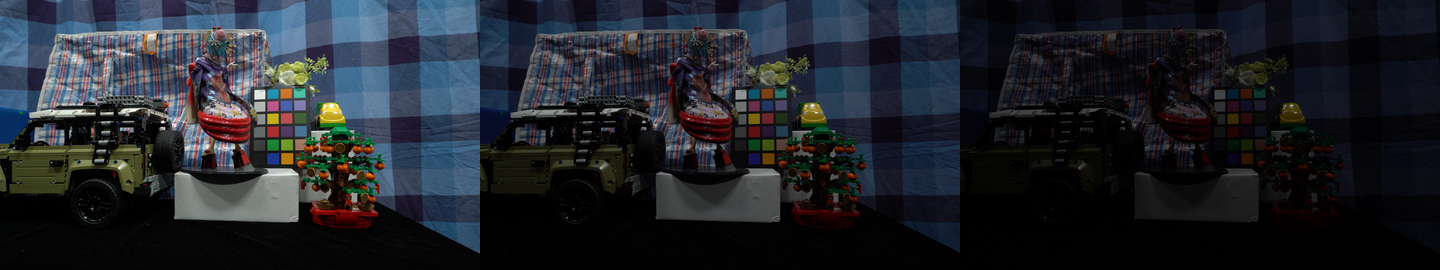}
    \caption{An example of two levels of real low-light video degradation: normal light, light intensity at 20\% and 10\%; video frames from BVI-lowlight dataset \cite{bvilowlight2023}}
    \label{fig:lowlight}
\end{figure}

In real-world scenarios, videos are often affected by multiple degradation factors simultaneously. For instance, videos captured in low-light environments typically suffer from under-exposure, low contrast, and significant noise. Limited photon availability results in dark images where details are obscured in shadows. To compensate, cameras may increase exposure time, risking motion blur, or raise sensor gain (ISO), which introduces noise. Consequently, low-light videos commonly exhibit a combination of issues: low brightness, poor contrast, color shifts from white balance errors, and amplified noise. Fundamentally, these degradations stem from a low signal-to-noise ratio (limited light leads to weak pixel intensities and relatively high noise levels). Camera post-processing that brightens the image also amplifies noise. Moreover, automatic exposure adjustments occurring frame-by-frame often cause temporal instability, resulting in flickering or stalled appearance. The combined effect of noise, motion blur, quantization artifacts, and temporal fluctuations makes low-light video enhancement a particularly challenging task that demands a careful trade-off between noise suppression and detail preservation. An example of low-light footage is shown in Fig. \ref{fig:lowlight}.

\begin{figure}[t]
    \centering
    \includegraphics[width=0.8\linewidth]{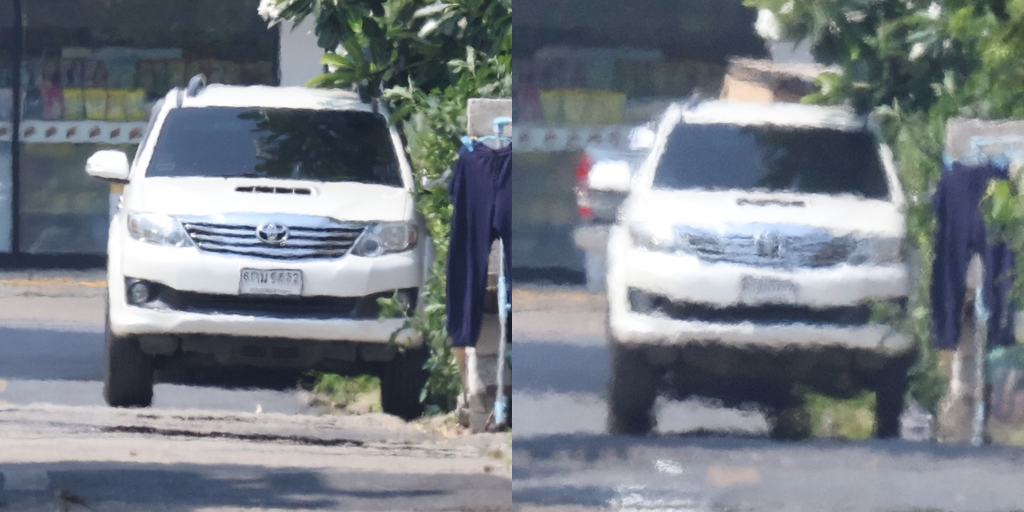}
    \caption{An example of two levels of atmospheric turbulence video degradation}
    \label{fig:atmospheric}
\end{figure}

Another example is atmospheric turbulence condition which involves distortions such as fog or haze, which reduce image contrast, and turbulence caused by temperature variations or aerosols. The increase in temperature difference between the ground and air manifests as changes in the interference pattern of refracted light. Under strong turbulence, video imagery suffers not only from blurring and scintillation, which is small-scale intensity fluctuations, but also from shearing effects, where different parts of objects appear to move in different directions. This complex degradation is commonly referred to as heat haze, with an example shown in Fig.~\ref{fig:atmospheric}.

\begin{table}[!t]
\centering
\caption{Common video degradations and their characteristics.}
\begin{tabularx}{\columnwidth}{p{1.5cm} p{2.3cm} X}
\toprule
\textbf{Degradation} & \textbf{Primary Causes} & \textbf{Key characteristics} \\
\midrule
Blur &
Relative motion (camera/object) during exposure; long exposure, defocus, atmospheric blur &
Loss of sharpness, streaking and smear, detail loss; varies per frame and spatially\\
\addlinespace
Noise &
Sensor physics (photon shot, read, thermal and dark current noise, high ISO gain)  quantization, transmission  &
Random intensity/color variations; mostly spatially independent but can be low frequency; mostly temporally uncorrelated; causes loss of fine details and adds undesired texture\\
\addlinespace
Transient particulate degradations &
Precipitation particles crossing the line-of-sight &
Bright/dark streaks or blobs; transient, spatially sparse occlusions; temporally inconsistent, causing flicker \\
\addlinespace
Continuous Scattering Degradations &
Atmospheric (dust, smoke, fog) or water particles scattering/absorbing light &
Low contrast, white/gray veil over entire frame; global light color cast; tends to be depth-dependent; temporal changes tied to weather/ambient conditions or camera motion through environment\\
\addlinespace
Low-Light &
Insufficient illumination; short exposure; small aperture &
Dark frames with low contrast and color distortion; high noise due to low SNR; auto exposure changes cause flicker; potential motion blur with higher exposures; details hidden in shadows\\

\addlinespace
Atmospheric Turbulence &
Refractive-index fluctuations from temperature gradients or aerosols over long paths &
Geometric distortions (warping, shearing), local blur and scintillation; strong temporal variation; reduces sharpness and contrast \\
\bottomrule
\end{tabularx}
\end{table}

\subsection{Traditional Video Enhancement and Restoration Methods}
\label{sec:trad}

We refer to non-learning-based methods as traditional methods. Traditional video restoration and enhancement techniques can be broadly categorized into model-driven and rule-based approaches, depending on their underlying design principles. Model-driven methods incorporate prior knowledge about the video acquisition process, noise characteristics, or motion dynamics, and are often inspired by perceptual models of human vision. Notable examples are Retinex theory for low-light enhancement or atmospheric scattering degradation model (Eq. \ref{eqn:light}). Another common approach involves constructing objective functions composed of a fidelity term and regularization terms that encode natural image priors (e.g., local smoothness, non-local similarity, and sparsity). These optimization problems are typically solved using iterative convex optimization methods such as ADMM \cite{7744574}. These model-based methods offer principled formulations and are often interpretable, making them appealing in contexts where explainability is important.

In contrast, rule-based methods enhance video quality using predefined, often handcrafted rules, without explicitly modeling the underlying image formation process or human visual perception. Common examples include global histogram Equalization, fixed gamma correction, linear contrast stretching, and brightness scaling. Tone mapping is also frequently implemented in this manner, using static curves or compression functions such as logarithmic or sigmoid mappings. These methods typically apply the same transformation uniformly, regardless of scene content. While this makes them computationally efficient and straightforward to implement, their lack of adaptability limits their effectiveness in handling diverse or complex video conditions.

Different image processing methods have been developed to address specific forms of degradation. For instance, noise in images and videos has been traditionally addressed through spatial, temporal or spatio-temporal filtering. The early spatial filtering techniques including median, Gaussian and Wiener filter \cite{lee1980digital,ghazal2008structure} are computationally efficient but suffer from severe blurring and loss of fine details. They also do not utilize temporal information inherent in video sequences.
Temporal filters, on the other hand, exploit the redundancy present in consecutive frames \cite{dubois2003noise, hsia2015high,yahya2015videonoise}. However, they can introduce artifacts and blur and are efficient only in relatively static videos. Spatio-temporal methods are aiming to combine the strengths of both approaches, utilizing information both in  spatial and temporal domains. Statistical filters such as extensions of Wiener and Kalman filters can be applied only under specific assumptions (such as stationarity, linearity or normality of the process) which in the case of real world videos rarely hold. Another example of spatio-temporal filtering approach is extensions of BM3D (Block Matching and 3D filtering) \cite{dabov2007image} for videos that searches for similarities across frames \cite{vbm3d, li2019enhanced, maggioni2012video}. These methods tend to outperform early filtering techniques, however, they have high computational complexity, can introduce block artifacts and their performance of these methods falls short in high noise level scenarios.

Traditional video dehazing methods often rely on atmospheric scattering model which describes hazy images as a combination of attenuated scene radiance and atmospheric airlight (see Eq.~\ref{eqn:light}) \cite{zhang2011video}. The techniques based on Dark Channel Prior (DCP) \cite{he2010single, lin2013dehazing} assume that most outdoor images have at least one color channel with very low values in the areas excluding sky. Methods utilizing this assumption have been proposed for images \cite{zhang2011video, song2015improved,lee2016review} and videos \cite{lv2010realdehazing,adidela2021single,park2018fast}. However, DCP struggles in scenes with high brightness and with large sky areas, causing color distortion and incorrect estimation of transmission map. Applying methods suited for images to video frames causes flickering and temporal inconsistencies. 

Conventional deblurring techniques aim to invert the blurring process that is mathematically modeled as convolution of a sharp image with Point Spread Function (PSF). They address it via deconvolution (e.g. Wiener deconvolution \cite{biswas2015deblurring} or Richardson-Lucy deconvolution \cite{richardson1972bayesian}). However, these methods require a known PSF and assumptions about noise and signal that do not necessarily hold. Additionaly, if PSF is inaccurately estimated, they can introduce artifacts such as noise and residual blur. Blind deconvolution does not require the knowledge of PSF \cite{cai2009blind}, but it often struggle with accurate frame alignment and separating motion blur from the actual movement.

Traditional methods for atmospheric turbulence mitigation rely on so called ``lucky regions'' or ``lucky images'' --- the cleanest and sharpest frames or frame regions that appear in a video sequence due to changing nature of atmospheric turbulence \cite{aubailly2009automated,lou2013video}. These frames are then aligned and fused  to produce a clearer image. These techniques are only efficient for scenes with little or no motion as it is difficult to distinguish between the turbulence and genuine motion at the registration stage. The same limitations apply to multi-frame fusion techniques \cite{zhu2011stabilizing,anantrasirichai2013atmospheric}. 

For conventional low-light enhancement, histogram-based methods like Histogram Equalization (HE) and its improved adaptive modifications, Adaptive Local Histogram Equalization (ALHE) \cite{kim1998contrast} and Contrast Limited Adaptive Histogram Equalization (CLAHE) \cite{zuiderveld1994contrast} were introduced. They involve manipulations with histograms that are aimed at improving global contrast, however they can result in over-enhancement and in case of CLAHE might introduce noise, result in loss of detail and blocky artifacts. 

Based on models of  human vision, Retinex theory \cite{land1971lightness} states that an image $I(x,y)$ can be decomposed into two components: illumination $L(x,y)$ and reflectance $R(x,y)$, using the relation $I(x,y)=L(x,y)R(x,y)$. This model assumes that intrinsic properties of the scene (such as color and texture) come from reflectance component, while most undesirable brightness variations stem from illumination, which can be estimated and suppressed to enhance local contrast and reveal finer details. Traditional low-light enhancement is achieved by estimating and boosting illumination component while preserving and enhancing the reflectance component. Building upon this theoretical framework, LIME \cite{10.1145/2964284.2967188} estimates illumination through minimization of an objective function incorporating a structure-aware L1 smoothness prior.
Single-scale Retinex (SSR) methods typically use Gaussian filter to estimate the illumination component from the input image \cite{jobson1997properties}. The Multi-Scale Retinex methods \cite{jobson1997multiscale} combine the results from SSRs at different scales that ensure better dynamic range compression and enhancement results. 

Traditional video restoration and enhancement methods, while being foundational to the field, share common limitation: their strong reliance on specific hand-crafted priors or explicit models that may not hold universally and often fail in the conditions of severe degradations. They require careful parameter tuning, struggle with dynamic scenes and tend to introduce artifacts and new degradations.
These limitations has driven interest towards data-driven approaches. They can learn to distinguish noise from signal and model complex scene motion without predefined rules, shifting video enhancement from explicit modeling to implicit learning of degradation from the data.

\subsection{Supervised Deep Learning Based Video Restoration and Enhancement Techniques}

Compared to traditional techniques, supervised deep learning implicitly learns complex priors directly from paired data, making it the dominant paradigm for video restoration. Recent surveys comprehensively cover supervised approaches across specific tasks, including denoising~\cite{sheeba2019denoisingreview}, deblurring~\cite{article}, dehazing~\cite{ayoub2024dehazingreview}, low-light enhancement~\cite{Li:low:2022, ye2024lowlightsurvey, zheng2022lowlightreview}, deraining~\cite{wang2022derainingsurvey}, and turbulence mitigation~\cite{hill2025deep}, alongside broader multi-task overviews~\cite{lhiadi2025deepehnhreview, rota2023video}. Readers are referred to these works for scenarios where paired ground truth data is available or synthetic data is used for supervised training.

\subsubsection{Convolutional Neural Networks}



{Convolutional Neural Networks (CNNs) are foundational for spatial feature extraction, often employing encoder-decoder structures like UNet~\mbox{\cite{ronneberger2015u}}. Temporal processing in CNNs typically involves sliding windows, 3D convolutions, or explicit motion compensation. Notable architectures include FastDVDnet~\mbox{\cite{9156652}} for efficient real-time denoising without explicit motion estimation, RViDeNet~\mbox{\cite{yue2020supervised}} for raw video processing using deformable convolutions, and PDVD~\mbox{\cite{li2023progressive}} for dehazing.} The evolution of CNN methods shows a trend towards more sophisticated temporal modeling, incorporating elements like deformable convolutions and attention mechanisms.

\subsubsection{Recurrent Neural Networks}

{Recurrent Neural Networks (RNNs) naturally model sequential dependencies and propagate temporal information across frames to ensure consistency. Bidirectional RNNs are particularly common. Prominent frameworks like BasicVSR~\mbox{\cite{chan2021basicvsr}} and BasicVSR++~\mbox{\cite{chan2022basicvsr++}} set benchmarks in video super-resolution and enhancement by leveraging extended bidirectional propagation and flow-guided deformable alignment. Similarly, dual-level recurrent architectures have been effectively adapted for temporally coherent video deraining (FCRVD)~\mbox{\cite{yang2019frame}}.}

\subsubsection{Transformer-Based Networks}
Transformer models, with their self-attention mechanisms, excel at capturing long-range dependencies in video sequences, potentially outperforming CNNs with limited receptive fields and RNNs with sequential processing bottlenecks. {The Video Restoration Transformer (VRT)\mbox{~\cite{10462902}} utilizes multi-scale architectures and mutual self-attention tailored for joint motion estimation, alignment, and feature fusion across multiple frames. While Transformers consistently achieve state-of-the-art structural fidelity, managing their high computational and memory costs remains an ongoing architectural challenge.}


\section{Unsupervised Learning in Video Restoration and Enhancement}
\label{sec:unsupervised}

Unsupervised methods reduce the reliance on labor-intensive data collection and annotation, helping to address the limitations posed by the scarcity of paired training data, particularly for dynamic scenes captured under adverse conditions or affected by complex, poorly understood degradations. Models trained on diverse unlabeled data often develop representations that are more robust to content and degradation variability than their supervised counterparts. These approaches are especially favorable in cases where degradation processes are poorly understood or difficult to simulate or characterize. By learning directly from the input data, they can adapt to the unique properties of each video, making them well suited for in-the-wild enhancement tasks.

\subsection{Terminology and Learning Paradigms}

{In literature on image and video restoration and enhancement, the terminology used to describe learning without paired ground truth is not always consistent. In particular, the terms \textit{unsupervised},\textit{self-supervised} and \textit{zero-shot} can refer to different aspects of the method and therefore should not be regarded as strictly disjoint learning paradigms.

Acquiring authentic ground-truth video data is often impractical for dynamic scenes and adverse real-world degradations. Most importantly, none of the paradigms discussed in this review rely on explicit paired ground truth during the final optimization phase, distinguishing them strictly from supervised methodologies. 

Operationally, we define \textit{unsupervised learning} as any framework optimized without paired targets. Within this category, \textit{self-supervised} methods derive surrogate training signals directly from the degraded input by exploiting internal video statistics, generating pseudo-clean targets or employing synthetic degradation models (e.g., masking or re-degradation). \textit{Weakly supervised} approaches make use of unaligned collections of degraded and unpaired clean data, typically enforcing structural preservation via cycle consistency. \textit{Zero-shot learning} and \textit{test-time optimization} adapts parameters strictly to a single degraded instance during inference without requiring prior dataset exposure. Frameworks incorporating pre-trained foundation models are classified by their optimization scheme: if network weights are updated at test time for a specific degraded input, the method is operationally zero-shot; if a pre-trained model acts purely as a static regularizer during training, the framework remains within the broader unsupervised category.

Because the boundaries between these paradigms frequently blur in practice, we first group the existing literature into five high-level, strategy-centric categories (Sections \mbox{\ref{sec:domain-transl}} -- \mbox{\ref{sec:prior}}). These categories classify techniques by their primary design intent: domain translation, self-supervised signal design, consistency-based methods, degradation-aware methods, and prior-based methods. To address the taxonomic overlap inherent in complex modern hybrid networks, we subsequently formalize these methodologies within a unified multi-dimensional taxonomy (Section \mbox{\ref{sec:orth_taxonomy}}), factorizing architectures by their supervision signal, temporal modeling strategy, and learning paradigm.}

\subsection{Domain Translation with Unpaired Learning}
\label{sec:domain-transl}
Unpaired adversarial training is a common strategy to bridge the domain gap between degraded and clean videos. They learn a mapping from degraded to clean domains without paired examples, typically via \textit{cycle‐consistency} and \textit{adversarial objectives}. Cycle-consistency frameworks, inspired by CycleGAN \cite{zhuUnpairedImageToImageTranslation2017} and DualGAN~\cite{8237572}, employ dual-generator architectures to enforce content preservation: a generator $G$ maps low-quality video frames to high-quality outputs, while an inverse generator $F$ attempts to reconstruct the original input from the enhanced output. This cycle-consistency constraint ensures that the restored frames retain the original content, even in the absence of paired ground truth. 

An adaptation of the CycleGAN structure has been proposed for low-light enhancement in \cite{Contextual:2021}, where the brightness and color characteristics of a low-light video (domain A) are mapped to those of a reference frame (domain B). Fuoli et al. \cite{fuoliEfficientRecurrentAdversarial2023} propose an adversarial framework, comprising a lightweight recurrent generator to propagate spatio-temporal information across frames, a degrader network with the same architecture, and a single joint recurrent discriminator that evaluates the realism of the enhanced video. By enforcing adversarial loss, the generator learns to produce outputs with both natural textures and temporal coherence. Similarly, Patil et al. \cite{patilUnpairedRecurrentLearning2025} address real-world video dehazing in an unpaired setting by coupling a dehazing network with a haze synthesizer. Their architecture integrates a Mixed Multi-Level Attentive (MMA) module for robust feature extraction and a Recurrent Multi-Attentive (RMA) module for temporal alignment, all within a compact 1.8M parameter model. The network learns to remove haze from input videos and reintroduces it via a backward branch, with a cycle-consistency loss enforcing structural preservation between the original and re-hazed frames.

Diffusion Models (DMs) have also been applied to domain translation tasks using cycle-consistency constraints. However, because they rely on training two generators and require large amounts of training data, DM-based methods have so far been primarily developed for images rather than videos \cite{li2023diffusion}.

\subsection{Self-Supervised Signal Design}
\label{sec:ss-signal}

Self-supervised  signal design techniques obtain their loss functions from the properties, transformations or inherent structure of degraded input data itself. These methods exploit internal redundancies and constraints within the data. {They do not require paired task-specific ground truth; depending on the formulation, pretrained priors or unpaired external data may still be incorporated.} 

\subsubsection{Blind-Spot and Noise-Based Methods}
\begin{figure*}[ht!]
    \centering
   
    \begin{subfigure}{\textwidth}
        \centering
        \begin{overpic}[width=\textwidth]{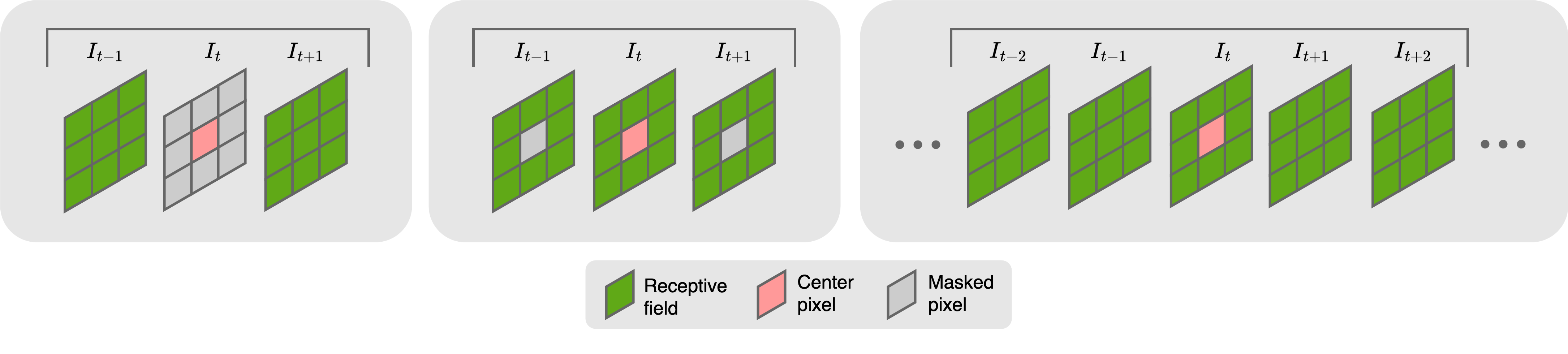}
        \put(73,23){RDRF \cite{wangRecurrentSelfSupervisedVideo2023}}
        \put(8,23){MF2F \cite{dewilSelfsupervisedTrainingBlind2021}}
        \put(34,23){UDVD \cite{shethUnsupervisedDeepVideo2021}}
        \end{overpic}
        \caption{}
        \label{fig:rdrd}
    \end{subfigure}
    
    \begin{subfigure}{0.61\textwidth}
        \centering
        \includegraphics[height=6cm]{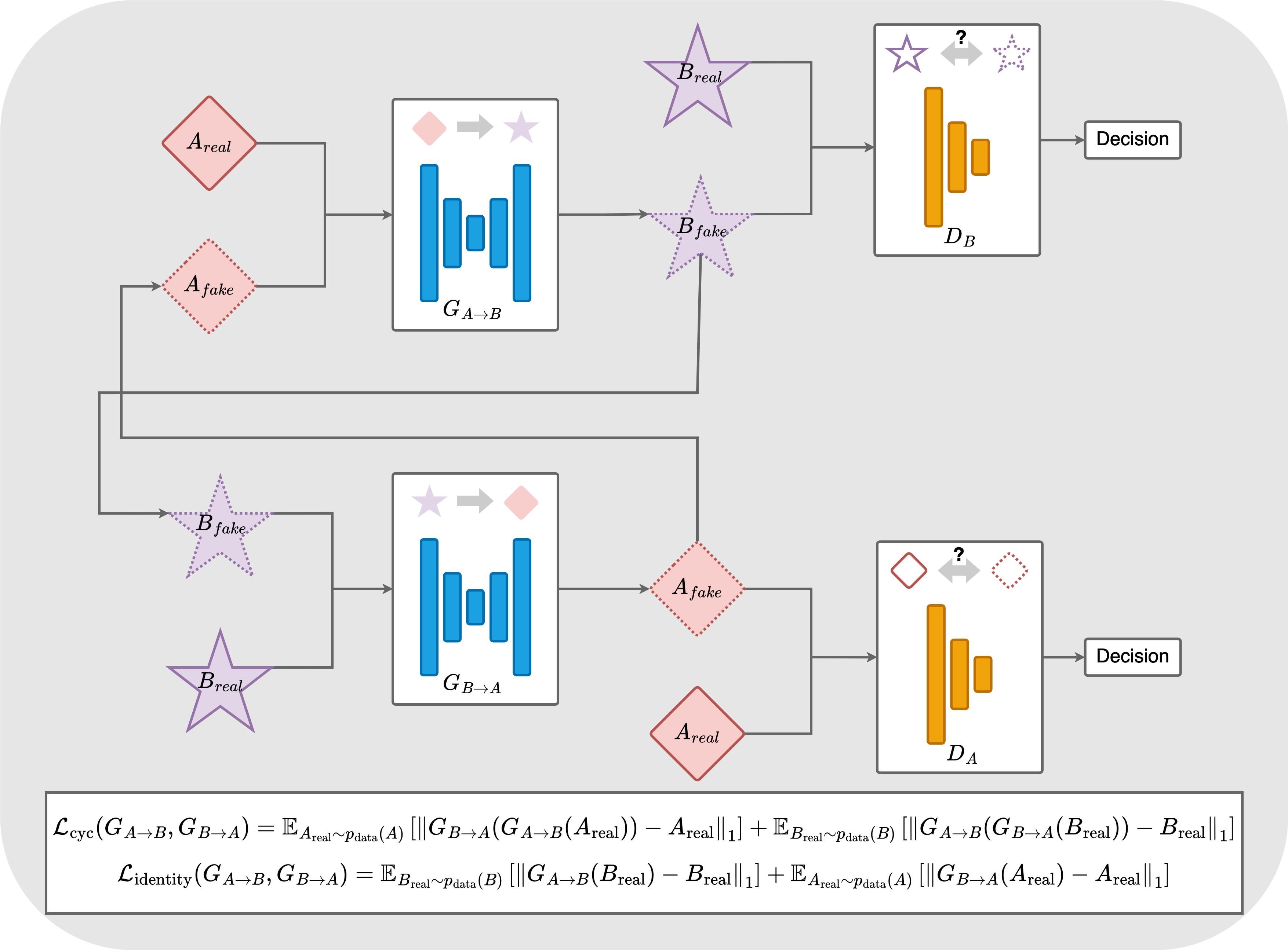}
        \caption{}
        \label{fig:cyclegan}
    \end{subfigure}\hfill
    \begin{subfigure}{0.37\textwidth}
        \centering
        \includegraphics[height=6cm]{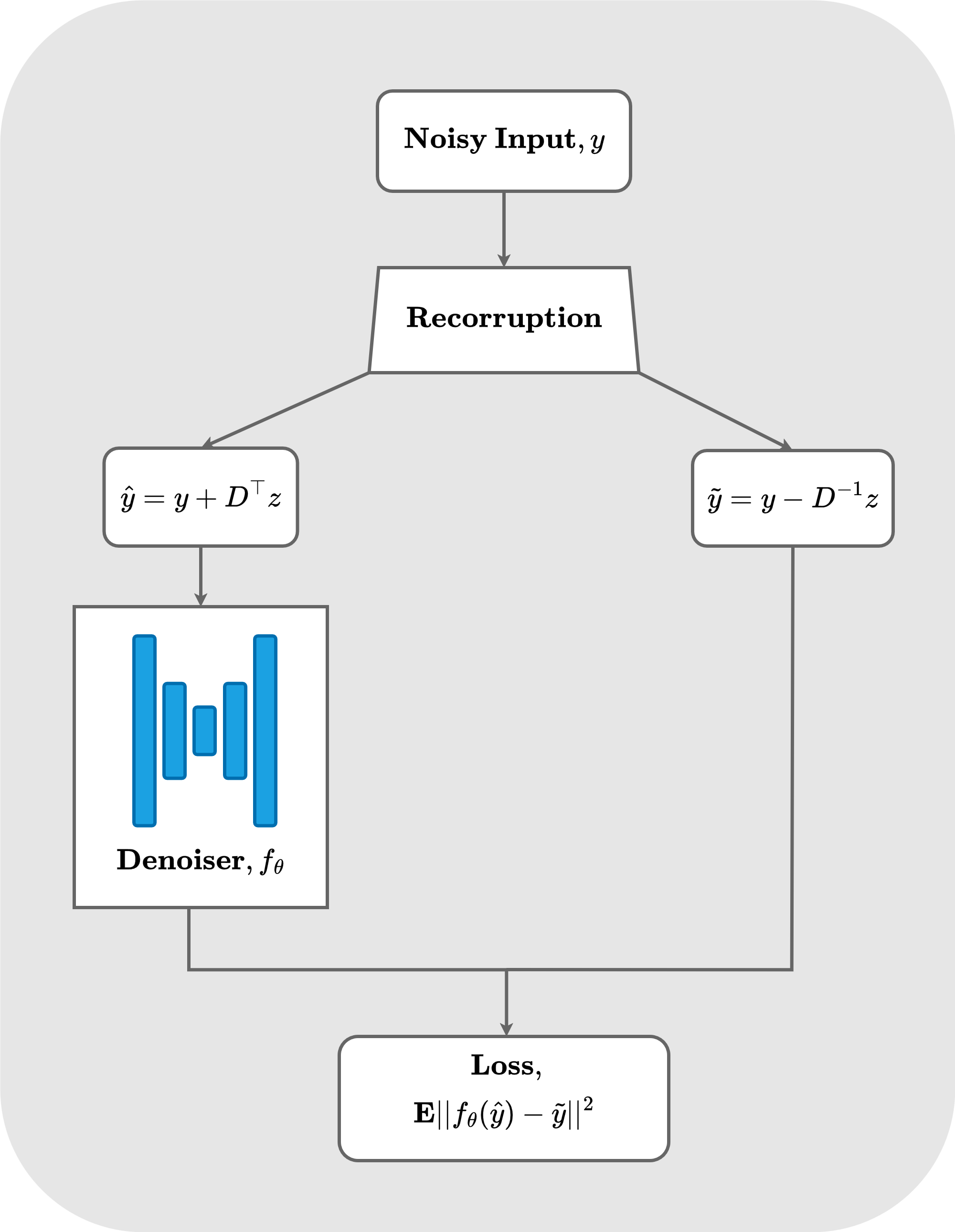}
        \caption{}
        \label{fig:rec2rec}
    \end{subfigure}
    
    \caption{Key design concepts in recent unsupervised video‐enhancement pipelines. (a) Spatio-temporal receptive fields. {The schematics illustrate the masking constraints and multi-frame receptive fields for MF2F, UDVD, and RDRF: MF2F’s motion-aligned warping draws information from closest frames; UDVD employs a spatial-temporal blind-spot mask around the center pixel; RDRF utilizes a denser receptive field and uses a recurrent blind-spot framework to leverage long-range temporal context.}
(b) CycleGAN backbone. Two generators $G_{A\to B}$,$G_{B\to A}$, and two discriminators $D_A,D_B$ are tied together by cycle-consistency and identity losses, enabling translation between domains $A$ and $B$ without paired ground truth.
(c) Recorrupt-to-Recorrupt (R2R) training loop. The method involves generating two uncorrelated re-corrupted versions of the same input and learning to map one to the other, mimicking supervised denoising without clean targets.}
    \label{fig:main}
\end{figure*}

When enhancing videos degraded by noise, a key challenge is learning from the noisy data alone. \textit{Blind-spot networks} have emerged as a powerful self-supervised tools for this scenario. They are designed such that when predicting a pixel’s intensity, the network has no access to that specific pixel in the input, relying only on its neighbors. This prevents the network from trivially copying the noisy input value, forcing it to learn denoising from context. Fig. \ref{fig:main} demonstrates the key methods in this area.

Sheth et al. \cite{shethUnsupervisedDeepVideo2021} introduced an Unsupervised Deep Video Denoiser (UDVD) that extends the blind-spot principle to the video domain. Their CNN is trained exclusively on noisy videos by masking out each target pixel and relying on adjacent pixels across spatial and temporal neighborhoods to predict it. By randomly shifting input frames or using multiple noisy frames of the same scene, UDVD obtains supervision signals similar to Noise2Noise \cite{lehtinen2018noise2noise}, effectively treating one noisy observation as the target for another. This approach achieved denoising performance close to supervised methods on various real-world datasets, despite never seeing a clean video. A crucial benefit of the video setting is redundancy: even though any single pixel is noisy, neighboring frames often carry a slightly different noise realization for the same scene point, which the network can exploit. UDVD and similar blind-spot methods demonstrate that, given sufficient video data, a model can calibrate itself to remove noise without ground truth, learning the noise distribution in the process.

Building on this idea, Dewil et al. \cite{dewilSelfsupervisedTrainingBlind2021} proposed Multi-Frame2Frame (MF2F), a self-supervised blind training scheme for multi-frame video denoising that also performs noise level estimation. In their approach, the network is trained to map a stack of consecutive noisy frames to a denoised output for the central frame. By assuming the noise is uncorrelated across frames, any one frame can serve as a target for its neighbors. In practice, they employ a strategy where one frame is reconstructed from its adjacent frames (which are treated as references); the loss penalizes differences between the network’s output and the actual noisy target frame. Since the only consistent signal across frames is the true underlying image (noise is random), the network learns to recover that image while the noise averages out. Furthermore, MF2F integrates a noise estimation sub-task. The network learns the noise variance by predicting inter-frame differences, which facilitates adaptive filtering of varying (heteroscedastic) noise levels. This blind, self-supervised joint denoising and noise-level learning yields a robust video denoiser that requires no clean data, effectively learning the noise model directly from the input video sequence itself.

While blind-spot methods avoid using the target pixel, they can suffer from information loss since masked pixels are never directly seen. Wang et al. \cite{wangRecurrentSelfSupervisedVideo2023} address this limitation by proposing a recurrent self-supervised denoising framework with a {R}icher and {D}enser {R}eceptive {F}ield (RDRF). Their method fully exploits both the reference frame and neighboring frames by aggregating information from a larger spatio-temporal neighborhood, while adhering to the blind-spot constraint. RDRF employs a recurrent architecture that processes a video  in both forward and backward directions, integrating information from distant frames to enlarge the temporal receptive field. A key innovation is a training scheme that maintains the self-supervised mask constraint even in a recurrent setup (which is non-trivial, as naive recurrence could leak a pixel’s own information from one time step to the next). This is achieved through offset blind-spot masks and merge features from multiple timesteps, ensuring that each output pixel is influenced by many context pixels, including those from past and future frames, but never by itself at the same timestep. The result is improved detail preservation and lower noise, as the network can utilize more of the video’s content. Empirically, RDRF outperforms earlier blind-spot video denoisers, achieving cleaner results with fewer texture artifacts, thanks to its denser utilization of available data.

Recently, a SpatioTemporal Blind-spot Network (STBN) \cite{chen2025spatiotemporal} has been proposed to enhance video denoising through bidirectional blind-spot feature propagation, ensuring accurate temporal alignment without leaking target pixel information. To address the instability of motion estimation in noisy conditions, the framework incorporates an unsupervised optical flow distillation mechanism, which guides the learning of flow from noisy videos using enhanced outputs. Furthermore, STBN replaces conventional bilinear interpolation in the warping operation with nearest-neighbor interpolation, which better preserves spatial correlations and the original noise distribution.  

In \cite{zhengUnsupervisedDeepVideo2023}, Zheng et al. use an untrained deep network that requires only the single test noisy video for its training process. A core contribution is the development of a self-supervised loss function, an extension of the ``Recorrupted-to-Recorrupted (R2R)" idea for images and videos (ER2R and VER2R). This loss function works by re-corrupting the noisy input frames with additionally simulated noise (from the same distribution as the measurement noise) to generate training pairs. This re-degradation scheme allows the loss to act as an unbiased estimator of its supervised counterpart, making it applicable to general random noise beyond just Gaussian noise. The method employs a two-stage training approach: first, a spatial denoising module is pre-trained using the ER2R loss to facilitate frame alignment; subsequently, the entire video denoising network, which includes a motion-compensation module and a lightweight temporal attention module for fusing temporal neighbors and handling misalignments, is trained using the VER2R loss.

\subsubsection{Optimization-Unrolling Methods}
Optimization-unrolling methods (sometimes called unfolding) treat image and video enhancement as an inverse problem and solve it with an iterative optimizer (e.g. Alternating Direction Method of Multipliers (ADMM) \cite{7744574} and Half Quadratic Splitting (HQS) algorithm~\cite{392335}), “unrolling” those iterations into a trainable network whose blocks represent an optimization step. During the training, the loss function is composed of the optimizer’s internal data-fidelity term and handcrafted regularizers.
A number of methods based on unrolled optimization were developed for image enhancement \cite{lyu2024enhancing,wu2022uretinex}.

To enhance low-light video, UDU-Net \cite{zhuUnrolledDecomposedUnpaired2025} unrolls the optimization process into a deep network that decomposes the video signal into spatial and temporal factors, which are then updated iteratively. The spatial component (Intra-subnet) corrects exposure and contrast using priors inspired by expert photography adjustments, effectively learning a mapping towards well-exposed statistics such as histograms of professionally retouched images. A human perceptual feedback mechanism is also incorporated to prevent under- or overexposure, giving the user some control over the brightness level. The temporal component (Inter-subnet) ensures that the illumination changes are smooth over consecutive frames. A discriminator is incorporated to ensure that the enhanced videos match the illumination distribution of expert-retouched footage.  Experimental results demonstrate that UDU-Net outperforms previous unsupervised learning methods by more than 3dB in PSNR on the SDSD \cite{wang2021seeing} dataset.

DUST~\cite{10479079} suppresses noise in videos by incorporating the assumption that the signals have a sparse representation in a dictionary, and the optimization algorithm is derived and unfolded into a deep unfolding Transformer encoder architecture. 

\subsubsection{Deep Plug-and-Play}

Deep Plug-and-Play (PnP) is a hybrid framework that integrates deep learning with classical optimization techniques, such as those used in model-based approaches for solving inverse problems \cite{ryu2019plug}. Specifically, it replaces the prior term in an optimization problem with a pre-trained image denoiser. The motivation behind PnP is to combine the advantages of both model-based and learning-based approaches. Model-based restoration methods are task-agnostic, relying on a specified degradation operator $H$ in the formulation $I_\text{obv} = H I_\text{idl} + n$ (Eq. \ref{eqn:degradationmodel}), and do not require training, making them adaptable across tasks. However, they depend on hand-crafted priors, and they are typically slower as it solves the problem with iterative algorithm, like ADMM \cite{7744574} or FISTA \cite{9913822}. In contrast, learning-based methods require extensive training but offer fast inference and strong performance due to end-to-end optimization. Hybrid approaches, such as deep plug-and-play methods, combine the strengths of both paradigms by using learning-based denoisers as learned priors within iterative model-based frameworks.

PnP is not exactly a form of learning, but rather an unsupervised inference or optimization framework that can incorporate learned components. The classic optimization form is given by 

\begin{equation}
\min_{x} \frac{1}{2} \| Hx - I_\text{obv}\|^{2} + \lambda R(x),
\end{equation}

where $x$ is an estimate of the ideal image $I_\text{idl}$, $\lambda$ controls the trade-off between the data fidelity term
$\frac{1}{2} \|Hx - I_\text{obv}\|^{2}$ and the regularization term $R(x)$, which serves as a prior. In PnP, the proximal operator associated with $R(x)$ is replaced by a pretrained denoiser. For applications in image restoration and enhancement, we refer the reader to \cite{9454311,10004791} for more details.

For video restoration and enhancement, Yuan et al. \cite{9495194} adopt a PnP-ADMM framework, leveraging FastDVDnet \cite{9156652}, a deep video denoising network, as the learned prior. Zerva et al. \cite{10401873} similarly employ a PnP-ADMM strategy but incorporate DnCNN \cite{7839189}, a pretrained image denoiser, alongside optical flow for motion compensation. Their approach primarily aims for video super-resolution. While deep PnP frameworks offer flexibility and strong generalization, they remain less popular in practice compared to standard image restoration networks due to their inherently iterative nature, which results in slower inference and increased computational cost.

\subsection{Consistency-Based Methods}
\label{sec:consistency}

Consistency constraints in video restoration force stable outputs across video frames or semantics. Methods based on consistency enhance video by enforcing agreement across related signals, such as neighboring frames or semantic content. Temporal consistency ensures smoothness across time, typically by penalising differences between frames after motion compensation. Semantic consistency uses a pretrained model to preserve high level features such as object identity or segmentation across enhancement. Consistency losses act as regularizers and are usually complemented by priors or data-fidelity terms.

\subsubsection{Temporal consistency} Unsupervised video restoration methods must ensure temporal consistency to prevent flickering or jitter across frames. Common approaches involve enforcing alignment between restored frames by accounting for motion dynamics. 
Yeh et al. \cite{yehDiffIR2VRZeroZeroShotVideo2025} enforce temporal consistency in a zero-shot manner using a hierarchical latent warping strategy. Their DiffIR2VR-Zero framework applies a pre-trained diffusion-based image restoration model in frame-wise fashion, while introducing intelligent warping of latent features between keyframes and their neighboring frames to maintain coherence. By combining optical flow with feature matching in a hybrid token merging mechanism, they adapt static image restoration model to video, achieving significantly improved temporal stability without retraining. Instead of estimating motion flow, Varghese et al. \cite{Varghese_2023_ICCV} estimate depth from two consecutive frames and use view synthesis to enhance temporal consistency. However, since this work focuses on underwater enhancement, the estimated depth is also used to compute the light transmission map.

An alternative strategy is to incorporate recurrent or plug-in temporal modules into an image restoration network. Fu et al. \cite{fuTemporalPluginUnsupervised2024} present \textit{Temporal-as-a-Plugin} (TAP), an unsupervised video denoising framework that inserts tunable temporal filtering blocks into a pre-trained single-frame denoiser. {The test-set adaptation variant, TAP-T utilizes inter-frame information to denoise consistently across time by fine-tuning these temporal modules on the target video data}. A progressive training schedule is used, where initial frame-by-frame denoising produces pseudo-clean frames, which are then used to iteratively refine the temporal modules, improving the network’s performance on real videos without any ground truth.

\subsubsection{Semantic consistency} Zheng et al. \cite{zhengSemanticGuidedZeroShotLearning2022} propose a Semantic-Guided Zero-shot (SGZ) method for low-light video enhancement. Their SGZ framework uses a pre-trained object recognition or segmentation network to enforce semantic consistency between low-light and enhanced frames. The key  the semantic content of a frame should remain invariant under enhancement. The SGZ network is optimized to maximize the similarity of deep semantic features extracted from the original and enhanced frames, and simultaneously improve the confidence of object detections in the enhanced output. By treating the original and enhanced frames as two views of the same scene, SGZ ensures that the network does not introduce artifacts that alter content. This method  effectively performs high-level contrastive alignment.

\begin{table*}[!tpb]
\centering 
\caption{Summary of Unsupervised Video Enhancement and Restoration Techniques }
\label{tab:summary_tabularx}

\begin{tabularx}{\textwidth}{p{1.6cm} p{2.7cm} X p{1.6cm} }


\toprule
Category & Method & Key idea & Task\\
\midrule

\multirow{3}{\linewidth}{\textbf{Unpaired Domain Translation }} 
    & Fuoli \emph{et al.} \cite{fuoliEfficientRecurrentAdversarial2023} 
    & Lightweight recurrent generator–degrader pair with a joint discriminator learns spatial detail and temporal coherence through adversarial training. 
    & General \\ 
\cmidrule(l){2-4}
    & Patil \emph{et al.} \cite{patilUnpairedRecurrentLearning2025} 
    & Compact recurrent network with MMA and RMA modules performs cycle-consistent dehazing and re-hazing. 
    & Dehazing \\ 

\hline

\multirow{1}{\linewidth}{\textbf{Optimization Unrolling}} 
    & UDU-Net \cite{zhuUnrolledDecomposedUnpaired2025} 
    
    & Unrolled optimizer iteratively updates spatial (expert-photography priors + perceptual feedback) and temporal factors.
    & Low-Light \\ 
\hline

\multirow{2}{\linewidth}{\textbf{Deep Plug-and-Play}} 
    & {PnP (Yuan \emph{et al.})} \cite{9495194} 
    & PnP-ADMM framework that plugs FastDVDnet as the denoiser prior for joint video restoration/enhancement. 
    & General \\ 
\cmidrule(l){2-4}
    & Zerva \emph{et al.} \cite{10401873} 
    & PnP-ADMM with DnCNN denoiser and optical-flow compensation targets video super-resolution. 
    & Super-Resolution \\ 
\hline
\multirow{6}{\linewidth}{\textbf{Temporal Consistency}} 
& DiffIR2VR-Zero \cite{yehDiffIR2VRZeroZeroShotVideo2025} & Tranining-free adaptation of a pre-trained image diffusion model. Uses hierarchical latent warping with optical flow and feature matching. & General \\
\cmidrule(l){2-4}
& Varghese et al. \cite{Varghese_2023_ICCV} & Estimates depth from consecutive frames and uses view synthesis for temporal consistency. & Underwater\\
\cmidrule(l){2-4}
& TAP \cite{fuTemporalPluginUnsupervised2024} & Inserts and fine-tunes temporal filtering blocks into a pre-trained single-frame denoiser. Uses a progressive training schedule. & Denoising \\
\hline

\textbf{Semantic Consistency} & SGZ \cite{zhengSemanticGuidedZeroShotLearning2022} & Uses a pre-trained object/segmentation network to enforce semantic consistency between original and enhanced frames. & Low-Light\\
\hline

\multirow{4}{\linewidth}{\textbf{Degradation-Aware Methods}} & Reblur2Deblur \cite{chenReblur2DeblurDeblurringVideos2018} & Re-blurs the restored frame using estimated optical flow and minimizes the difference to the original blurry input. & Deblurring \\
\cmidrule(l){2-4}
& RFR \cite{leeRestoreRestoredVideo2021} & Fine-tunes a pre-trained denoiser by re-adding noise to its own output (pseudo-clean frames) and comparing to the input. & Denoising \\
\cmidrule(l){2-4}
&
ADST-BPN \cite{10540131}
&
Further downsamples and temporally subsamples the target LR video
to construct internal training pairs for zero-shot adaptation; MCT/SPMC provides motion estimation and compensation.
& Super-Resolution \\
\cmidrule(l){2-4}
& VER2R \cite{zhengUnsupervisedDeepVideo2023} & Uses a "Recorrupted-to-Recorrupted" (R2R) loss by adding more noise to the input frames to create training pairs for an untrained network. & Denoising \\
\cmidrule(l){2-4}
& LightenFormer \cite{lvUnsupervisedLowLightVideo2023} & Unsupervised transformer model that uses inverse S-curve consistency to reconstruct the original low-light inputs. & Low-Light \\
\hline

{\textbf{Physics-Based Priors}} & Zero-TIG \cite{li2025zerotig} & Decomposes frames into reflectance/illumination based on Retinex theory. Uses 11 no-reference losses and temporal feedback via optical flow. & Low-Light \& Underwater \\

\hline

\multirow{1}{\linewidth}{\textbf{Learning-Based Priors}} 
& VDP \cite{NEURIPS2023_6ba85c6f} & Learns a video dynamics prior by predicting the next frame's latent embedding from the current one. & General\\

\cmidrule(l){2-4}
& UnDIVE \cite{Srinath:undive:2025} & Uses generative diffusion-based priors and minimizes the difference between warped enhanced frames in both forward and backward directions. & Underwater Enhancement \\

\hline

\multirow{4}{\linewidth}{\textbf{Blind-Spot / Noise-Based}} & UDVD \cite{shethUnsupervisedDeepVideo2021} & Extends the blind-spot principle to video; predicts a masked pixel using spatial and temporal neighbors from noisy frames. & Denoising \\
\cmidrule(l){2-4}
& MF2F \cite{dewilSelfsupervisedTrainingBlind2021} & Multi-frame blind training scheme that jointly performs denoising and noise level estimation. & Denoising \\
\cmidrule(l){2-4}
& RDRF \cite{wangRecurrentSelfSupervisedVideo2023} & Recurrent architecture that enlarges the spatio-temporal receptive field while maintaining the blind-spot constraint. & Denoising \\
\cmidrule(l){2-4}
& STBN \cite{chen2025spatiotemporal} & Uses bidirectional blind-spot propagation, unsupervised optical flow distillation, and nearest-neighbor warping. & Denoising \\
\bottomrule
\end{tabularx}
\end{table*}

\subsection{Degradation-Aware Methods}
\label{sec:degradation}

Another category of unsupervised methods derives self-supervision from simulating the degradation process and enforcing reconstruction consistency, often referred to as ``analysis-by-synthesis''. The core idea is to apply a differentiable degradation model to the restored output and compare it against the original input, thereby generating a training signal in the absence of ground truth. A representative example is Reblur2Deblur \cite{chenReblur2DeblurDeblurringVideos2018}, which addresses video deblurring using a physics-based motion blur model. After restoring a sharp frame, the output is reblurred using a blur kernel estimated from optical flow between consecutive frames. A self-supervised loss minimizes the discrepancy between this reblurred result and the original blurry input, enforcing physical consistency with the observed motion. However, this method is not fully unsupervised, as it relies on a supervised pre-trained image or video deblurring network and uses a hybrid loss function during fine-tuning, which combines the self-supervised loss with a supervised loss to avoid degenerate solutions.

Similar analysis-by-synthesis strategies have been applied to other video degradation types. In the context of denoising, this involves reintroducing noise to the denoised output and compare it against the original noisy input. Lee et al. \cite{leeRestoreRestoredVideo2021} follow this principle in their ``Restore-from-Restored'' (RFR) approach for self-supervised video denoising. This method is based on Noise2Noise \cite{lehtinen2018noise2noise}  and is an extension of frame-to-frame training approach \cite{ehret2019model} that allows the network to train during the test phase with self-supervision. RFR does not require optical flow estimation due to translation equivariant property of Fully Convolutional Networks (FCNs). A pre-trained denoiser is used to generate a pseudo clean video from a given noisy input, forming synthetic training pairs of \{noisy frame, pseudo-clean frame\}. The denoiser is then fine-tuned by reintroducing noise to its output and minimizing the error to the input. This process effectively treats the network’s own output as a pseudo ground truth, refining the model in offline or iterative manner.

Another example is LightenFormer, proposed by Lv et al.~\cite{lvUnsupervisedLowLightVideo2023}, the first unsupervised transformer-based approach for low-light video enhancement.
Unlike Retinex-based methods, LightenFormer does not explicitly decompose images into illumination and reflectance. Instead, it adopts a learnable S-curve adjustment strategy for illumination enhancement. {Its inverse-consistency loss exploits the invertibility of the S-curve by requiring the enhanced frame, when mapped back through the inverse transformation, to recover the original low-light input. This provides a degradation-aware self-supervision signal without paired ground truth. In addition, a bi-directional denoising loss exploits noise independence across frames, while the spatial-temporal self-attention mechanism models long-range dependencies for temporally consistent enhancement.}


\subsection{Prior-Based Methods}
\label{sec:prior}
In prior-based methods, the enhancement is regularized with explicit knowledge (unlike consistency or self-supervised signal design methods, where supervision stems from implicit information within the signal), that is inspired by the physics models of image formation or learned priors distilled from large datasets. Priors are aimed to reduce artifacts, impose desirable properties and guide plausible reconstructions.

\subsubsection{Physics-Based Priors} 
In the absence of ground truth, integrating domain-specific physics models or priors can meaningfully constrain the solution space for video restoration. Two prominent examples are atmospheric scattering models for dehazing and the Retinex theory for low-light enhancement.

As discussed in \ref{sec:trad}, Retinex theory states that an image can be decomposed into reflectance and illumination. Low-light video enhancement thus involves estimating and adjusting the illumination while preserving reflectance. Several unsupervised methods incorporate this idea or related illumination priors. 
For instance, Zero-TIG \cite{li2025zerotig}, a zero-shot learning approach designed for enhancing low-light and underwater videos, exemplifies this principle by  decomposing input frames into reflectance and illumination components for targeted enhancement. The architecture features two main modules: an enhancement module and a temporal feedback module. Similar to Zero-IG~\cite{shi2024zero}, the enhancement module contains three subnetworks: a low-light denoising network (LD-Net), trained via self-supervision following Noise2Noise~\cite{lehtinen2018noise2noise}; an illumination estimation network (IE-Net); and a reflectance denoising network (RD-Net). To ensure temporal consistency, the feedback module recursively integrates information from previously enhanced frames. This is achieved by applying histogram Equalization to the current pre-denoised frame, estimating optical flow via RAFT between the previous and current frames, and warping the enhanced components of the previous frame to align with the current frame before reintegration. Optimization is driven by a set of 11 no-reference loss functions adapted from prior zero-shot methods. For underwater video enhancement, Zero-TIG further addresses color distortion by adaptively balancing RGB channels during illumination estimation.

In dehazing problems, several priors are commonly employed, including the color ellipsoid prior, color line prior, dark channel prior, color attenuation prior, and haze line prior. These priors are detailed in comprehensive reviews of single-image defogging \cite{pandey2025comprehensive}. However, the extension of unsupervised dehazing or defogging techniques to video remains limited and relatively underexplored.


\subsubsection{Learning-Based and Generative Priors}
As an alternative to using known physics-based or image formation models, deep neural networks have been employed to capture low-level image statistics as implicit priors, a concept first introduced in the Deep Image Prior (DIP) framework \cite{ulyanov2018deep}. Following DIP, Deep Generative Prior \cite{Pan:exploiting:2022} was proposed, where image priors are learned by a GAN trained on large-scale natural images, rather than optimizing a single low-quality image. More recently, DDRM \cite{NEURIPS2022_95504595} employs diffusion generative models as learned priors, using a variational inference objective to estimate the posterior distribution of the inverse problem.

In fully unsupervised scenarios, leveraging pre-trained foundation models as learned architectural priors provides powerful spatial, semantic, and geometric regularizers. For instance, UnDIVE \cite{Srinath:undive:2025} employs generative image priors trained on underwater scenes using Denoising Diffusion Probabilistic Models (DDPM) \cite{ho2020denoising}. Parallel to purely visual representations, multimodal vision-language foundation models provide robust high-level guidance that prevents content hallucination. UPID-EDM \cite{Wen:unpaired:2024} utilizes visual-language priors embedded within the Contrastive Language-Image Pre-training (CLIP) model \cite{radford2021learning} to effectively distinguish between background semantics and dynamic, localized degradation like rain streaks. Similarly, ZVRD \cite{cao2025zeroshot} exploits the semantic consistency of frame-level CLIP embeddings over contiguous video sequences to maintain structural fidelity. 

However, directly applying spatial priors or foundation networks designed for static images frame-by-frame to sequence-based data typically introduces severe temporal flickering and structural instabilities. To mitigate this discrepancy, Shrivastava et al. proposed the Video Dynamics Prior (VDP) \cite{NEURIPS2023_6ba85c6f}, which introduces an internal temporal modeling constraint directly within the latent space. VDP partitions the architecture into two primary networks, $f$ and $g$. The network $g$ acts as a predictive transition function that models the evolution of sequential representations across frames:
\begin{equation}
z_{t+1} = g(z_t),
\end{equation}
where $z_t$ is the latent embedding of the current frame. Concurrently, $f$ functions as a spatial decoder tasked with reconstructing the clean frame $\hat{X}_{t+1}$ from the updated latent space step:
\begin{equation}
\hat{X}_{t+1} = f(z_{t+1}) = f(g(z_t)).
\end{equation}

Another downside of relying exclusively on the global priors embedded in video generative models is limited reconstruction power for complex, localized structures. To overcome this, Chen et al. introduced GenVE \cite{Chen_2025_ICCV}, a generative video enhancement framework that maps global sequence semantics to local textures through a dual-stage pipeline. The method first employs an image diffusion model to upscale a source keyframe, creating a high-quality spatial anchor that preserves global scene semantics. A video controller, integrated into a 3D-UNet backbone and driven by a local-aware cross-attention mechanism, subsequently extracts patch-level textures from this anchor to refine and stabilize texture across neighboring frames. This approach allows pre-trained video generative models to maintain spatial-temporal consistency while preserving high-frequency details under severe degradation.

The five broad strategies described above establish a functional hierarchy, giving the researchers an intuitive starting point by classifying unsupervised video processing by their primary design intent. However, as architectures become more sophisticated, this high-level grouping exposes systematic taxonomic limitations. Modern hybrid networks frequently blur these rigid division. For instance, pairing generative diffusion models with test-time optimization loops and optical flow physical constraints. Attempting to force these multi-faceted frameworks into a single hierarchical label results in severe classification overlap. To resolve this, the structural paradigms must be decoupled from absolute categories and remapped onto an independent coordinate space, motivating the multi-dimensional taxonomy proposed below.

\subsection{Unified Multi-dimensional Taxonomy}
\label{sec:orth_taxonomy}
 While the strategy-centric classifications in Sections \ref{sec:domain-transl} - \ref{sec:prior} categorizes existing literature by operational intent, the complexity of modern unsupervised models challenges rigid boundaries. The existing methods frequently combine self-supervision signals with physical modeling and adversarial steps, leading to taxonomic overlap. The diversity of these approaches requires a structured taxonomy to identify shared design patterns and performance trade-offs. To address the inherent overlap between methodological strategies, we propose a three-dimensional classification design for unsupervised video enhancement, categorizing existing literature across three fundamental design dimensions: the Supervision Signal (S), Temporal Modeling (T), and Learning Paradigm (L). 
 
\begin{table*}[t]
\centering
\caption{Unsupervised video restoration and enhancement methods classified using the proposed factorized taxonomy: supervision signal (S), temporal modeling (T), and learning paradigm (L) alongside with task and degradation types, training and evaluation data and performance limitations.}
\label{tab:video_taxonomy}
\resizebox{\textwidth}{!}{%
\begin{threeparttable}
\small
\begin{tabular}{l l l c c c l l l}
\toprule
\textbf{Method} & \textbf{Task} & \textbf{Degradation} & \textbf{S} & \textbf{T} & \textbf{L} & \textbf{Training Data} & \textbf{Eval Dataset} & \textbf{Key Limitations} \\
\midrule
Fuoli et al.~\cite{fuoliEfficientRecurrentAdversarial2023} & {General} & {Illumination/Color} & S2 & T3 & L2 & {Unpaired collections} & {Vid3oC, Huawei} & {Training instability/hallucination risk} \\
Patil et al.~\cite{patilUnpairedRecurrentLearning2025} & {Dehazing} & {Scattering} & S2 & T3 & L2 & {Unpaired hazy/clean} & {REVIDE,RSA100Haze} & {Hallucination risks} \\
LightenFormer~\cite{lvUnsupervisedLowLightVideo2023} & {Low-Light} & {Illumination} & S3 & T1 & {L1} & {Unpaired collections} & {SDSD/LLIV-Phone} & {High computational cost} \\
UDVD~\cite{shethUnsupervisedDeepVideo2021} & {Denoising} & {Sensor Noise} & S1 & T1 & L1 & {Noisy sequences} & {DAVIS, Set8, Derfs, Vid3oC, Raw} & {Blind-spot info loss} \\
MF2F~\cite{dewilSelfsupervisedTrainingBlind2021} & {Denoising} & {Sensor Noise} & S1 & T1 & L5 & {Consecutive frames} & {Derf, Vid3oC-10, CRVD, FLIR ADAS} & {Fails in heavy motion} \\
RDRF~\cite{wangRecurrentSelfSupervisedVideo2023} & {Denoising} & {Sensor Noise} & S1 & T3 & L1 & {Noisy sequences} & {Set8, DAVIS,CRVD} & {Recurrent processing, receptive field constraints} \\
STBN~\cite{chen2025spatiotemporal} & {Denoising} & {Sensor Noise} & S1 & T2 & L1 & {Noisy sequences} & {Set8, DAVIS, CRVD} & {Flow failure in low SNR} \\
UDU-Net~\cite{zhuUnrolledDecomposedUnpaired2025} & {Low-Light} & {Illumination} & S4 & T2 & L3 & {Unpaired collections} & {SDSD + MIT-Adobe FiveK} & {Relies on rigid priors and BRISQUE feedback} \\
PnP (Yuan)~\cite{9495194} & {General} & {Various} & S3\tnote{a} & T1 & L3 & {Pre-trained prior} & {Simulated + Real SCI } & {Iterative latency} \\
Zerva et al.~\cite{10401873} & {Super-Res} & {Resolution} & S3\tnote{a} & T2 & L3 & {Pre-trained prior} & {Vid4 (Calendar, City)} & {Iterative latency} \\
DiffIR2VR-Zero~\cite{yehDiffIR2VRZeroZeroShotVideo2025} & {General} & {Various} & {S4} & T2 & L4 & {Pre-trained prior} & {REDS4, Vid4, DAVIS for SR; REDS30, Set8} & {Dynamic-scene flicker, Extreme degradation} \\
Varghese et al.~\cite{Varghese_2023_ICCV} & {Underwater} & {Color/Scattering} & S3 & T2 & L1 & {DRUVA} & {DRUVA, SQUID, Sea-thru, RUIE, UIEB} & {Depth estimation errors} \\
TAP~\cite{fuTemporalPluginUnsupervised2024} & {Denoising} & {Sensor Noise} & S4 & T1 & L1 \tnote{b} & {Noisy training videos} & {Set8, DAVIS, CRVD} & {Test-time latency for TAP-T} \\
SGZ~\cite{zhengSemanticGuidedZeroShotLearning2022} & {Low-Light} & {Illumination} & S5 & T0 &{L1}\tnote{c} & {2002 low-light/exposure-varied images} & {Author-captured drone video} & {Frame-by-frame flicker} \\
Reblur2Deblur~\cite{chenReblur2DeblurDeblurringVideos2018} & {Deblurring} & {Motion Blur} & S3 & {T1}\tnote{d} & {L5} \tnote{e} & {Blurry videos} & {MCD, DVD, ISOCHART, WFA} & {Blur model dependency} \\
RFR~\cite{leeRestoreRestoredVideo2021} & {Denoising} & {Sensor Noise} & S3 & T0 & L5 & {Pretrained denoiser + target noisy video} & {DAVIS} & {Test-time latency} \\
VER2R~\cite{zhengUnsupervisedDeepVideo2023} & {Denoising} & {Sensor Noise} & {S1} & {T2} & {L5} & {Single noisy video} & {DAVIS, Set8 and real raw-video data} & {Assumes known noise} \\
Zero-TIG~\cite{li2025zerotig} & {Low-Light, Underwater} & {Illumination/Noise/Color} & S4 & T2 & L5 & {Single degraded video} & {BVI-RLV, BVI-Coral} & {Test-time latency} \\
VDP~\cite{NEURIPS2023_6ba85c6f} & {General} & {Various} & S4 & T3 & {L5} & {Single degraded video} & {DAVIS, UCF-101, Vimeo90K-T} & {High computational cost} \\
UnDIVE~\cite{Srinath:undive:2025} & {Underwater} & {Color/Scattering} & S4 & T2 & L4 & {UIEB + UVE-38k; DDPM prior} & {VDD-C, Brackish, UOT32, MVK} & {Generative hallucination} \\
ADST-BPN~\cite{10540131} & {Super-Res} & {Resolution + Real-world degradation} & {S1}\tnote{f} & {T2}\tnote{g} & L5 & {External + target video} & {Vid4, REDS4, RealVSR, MVSR4\mbox{$\times$}} & {Alignment errors, requires meta-training} \\
\bottomrule
\end{tabular}%

\begin{tablenotes}[flushleft]
\footnotesize

\item[a] {PnP methods combine physical data fidelity with a pre-trained
denoising prior. They are classified as S3 because the observation model
provides the main constraint linking the reconstruction to the measured
data, although S4 is also a plausible interpretation because the
pre-trained denoiser acts as an explicit learned prior.}

\item[b] {Standard TAP trains the temporal modules on noisy training videos
and subsequently applies the resulting model at inference, corresponding
to L1. The test-set adaptation variant, TAP-T, instead updates the temporal
modules directly on the target test videos and would therefore be classified
as L5.}

\item[c] {Although SGZ is described by its authors as a zero-shot method,
the enhancement network is trained offline and no parameter optimization is
performed on the specific test image or video. Under the operational
definition of the L-axis, it is therefore classified as L1 rather than L5.}

\item[d] {The temporal classification of Reblur2Deblur depends on the
underlying deblurring backbone. The DVD-Reblur implementation performs
multi-frame restoration and is classified as T1, whereas the
DeblurGAN-Reblur implementation is essentially frame-based and would be
closer to T0. Optical flow is primarily used to construct the reblurring
self-supervision rather than for forward-pass feature alignment.}

\item[e] {Reblur2Deblur fine-tunes a pre-trained deblurring network on
unlabeled target blurry videos, making L5 the closest category. However,
the method retains pre-trained supervised initialization and may combine
the self-supervised reblurring loss with supervised training samples.}

\item[f] {ADST-BPN/CASTNet uses internal self-supervision during the
zero-shot meta-test stage: the target LR video is further downsampled to
construct an internal input--target pair. It is therefore classified as
S1 for the adaptation stage, although the full framework also uses
supervised external and meta-training.}

\item[g] {T2 refers to the complete CASTNet pipeline, which estimates motion
and explicitly compensates it using the MCT/SPMC modules. If ADST-BPN is
considered only as the isolated enhancement subnetwork, its temporal
processing is closer to T1.}

\end{tablenotes}
\end{threeparttable}
}
\end{table*}

\begin{figure*}[ht]

    \centering
        \includegraphics[height=11cm]{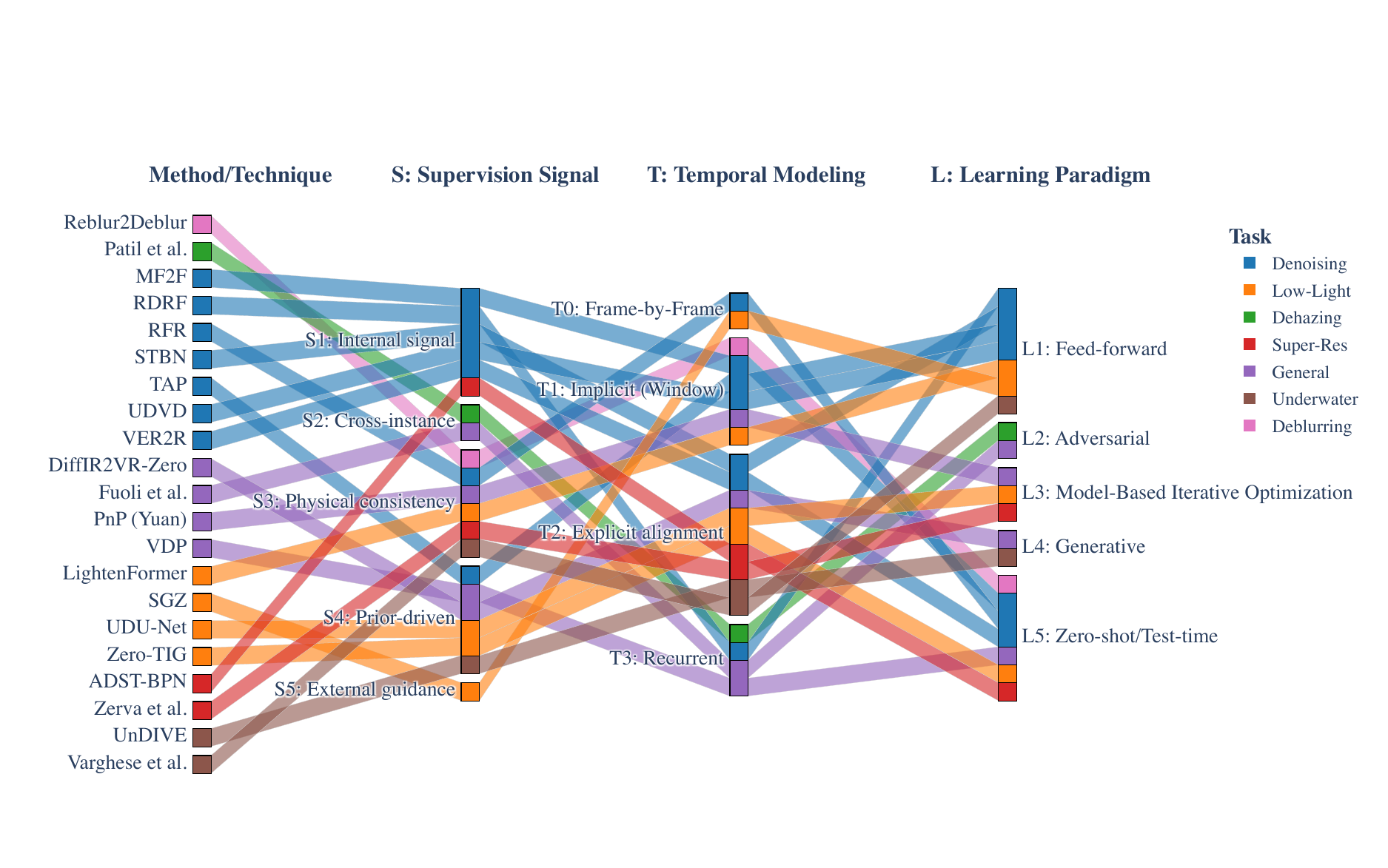}
    \caption{Multi-dimensional taxonomy of unsupervised video restoration methods. Parallel categories plot showing the distribution of representative video restoration techniques across three orthogonal axes: (S) supervision signal, (T) temporal modeling, and (L) learning paradigm. Each polyline corresponds to a method and traces its assignment across the axes, while the color encodes the target application task. The methods are grouped and ordered by task. }
    \label{fig:multidim_taxonomy}
\end{figure*}

\subsubsection{Axis 1: Supervision Signal}
This dimension categorizes the underlying source of the learning signal employed in the absence of
paired ground-truth labels. This categorization isolates the explicit data constraints and mathematical proxies used to bypass missing references. Separating this axis reveals whether a pipeline depends on unaligned external domains, internal structural redundancies, or explicit forward degradation equations. We identify five distinct signal types:
\begin{itemize}
    \item \textbf{Internal Signal (S1):} Supervision is derived directly from the statistical properties or structural redundancies of the degraded sequence itself. This includes blind-spot networks like UDVD \cite{shethUnsupervisedDeepVideo2021} and RDRF \cite{wangRecurrentSelfSupervisedVideo2023}, as well as spatiotemporal consistency frameworks such as STBN \cite{chen2025spatiotemporal}.
    \item \textbf{Cross-Instance/Domain (S2):} Models learn mappings between unaligned collections of degraded and clean video domains, typically utilizing cycle-consistency \cite{zhuUnpairedImageToImageTranslation2017} or adversarial frameworks like in Fuoli et al. \cite{fuoliEfficientRecurrentAdversarial2023}.
    \item \textbf{Physical/Degradation Consistency (S3):} In these methods,a ``re-degradation'' constraint is enforced, where the restored output, when passed through an estimated or explicit degradation model like atmospheric scattering or motion-based blur kernels, must match the original input. Examples include Reblur2Deblur \cite{chenReblur2DeblurDeblurringVideos2018}, which enforces consistency between blurred outputs and input frames, and S2SVR \cite{linUnsupervisedFlowAlignedSequencetoSequence}.
    \item \textbf{Prior-Driven (S4):} These approaches incorporate regularization from explicit human-engineered physical models like Retinex theory \cite{land1971lightness, Li2025TempRetinex} or implicit priors such as Deep Image Prior \cite{ulyanov2018deep} and Video Dynamics Priors \cite{NEURIPS2023_6ba85c6f}.
    \item \textbf{External Guidance (S5):} In these methods, the supervision is provided by pre-trained high-level vision models, such as the semantic-guided zero-shot learning (SGZSL) framework \cite{zhengSemanticGuidedZeroShotLearning2022} or CLIP-based assessments \cite{radford2021learning, wang2022exploring}.
\end{itemize}

\subsubsection{Axis 2: Temporal Modeling}
Methods can be distinguished by their strategy for aggregating inter-frame information to ensure temporal coherence \cite{Lai_2018_ECCV}. Isolating this mechanism shows the fundamental tradeoff between computational latency, motion-tracking accuracy, and frame-to-frame stability. It decouples the mechanical tracking of sequence dynamics from both the supervision format and the optimization loop:
\begin{itemize}
    \item {\textbf{Frame-by-Frame Processing (T0):} Temporal features are not explicitly aggregated during the forward pass. The network executes a purely spatial mapping generating the restored frame independently. Any temporal consistency is enforced implicitly via the optimization objective or loss function during training/test-time adaptation, rather than through architectural temporal modules. Examples include SGZ } \cite{zhengSemanticGuidedZeroShotLearning2022} and RFR \cite{leeRestoreRestoredVideo2021}.
    \item \textbf{Implicit/Window-Based (T1):} Temporal features are aggregated local neighborhoods without explicit motion estimation, commonly used in zero-shot denoisers like TAP \cite{fuTemporalPluginUnsupervised2024} and coordinate-based networks \cite{10647922}.
    \item \textbf{Explicit Alignment (T2):} Relative motion is compensated using optical flow or depth-based warping prior to fusion, as demonstrated in Zero-TIG \cite{li2025zerotig} and unsupervised deblurring frameworks \cite{Chen_2024_CVPR}.
    \item \textbf{Recurrent (T3):} Temporal dependencies are modeled sequentially, propagating information across long-range sequences through hidden states or visual state-space models like BVI-Mamba \cite{huang:bvi:2025}, VDPrior \cite{Shrivastava2024Prior} or UnpairedVD \cite{patilUnpairedRecurrentLearning2025}.
\end{itemize}

\subsubsection{Axis 3: Learning Paradigm}
This axis defines the operational framework and optimization strategy during training and inference. Differentiating these paradigms reveals whether a model maps features through static regression, distribution matching, iterative unrolled optimizations, generative posterior sampling, or direct test-time minimization:
\begin{itemize}
    \item \textbf{Feed-Forward (L1):} Standard end-to-end architectures optimized to minimize consistency-based losses, typical of UDVD \cite{shethUnsupervisedDeepVideo2021} and DUVD \cite{DUVD:2024}.
    \item \textbf{Adversarial (L2):} Training is formulated as a minimax game to ensure the restored output lies on the natural video manifold \cite{jiang2021enlightengan, patilUnpairedRecurrentLearning2025}.
    \item  {\textbf{Model-Based Iterative Optimization (L3):} These methods incorporate explicit optimization structures into restoration pipelines by either mapping iterative solver steps into trainable network stages or replacing explicit regularizers with pretrained denoising priors} \cite{9454311,10401873,zhuUnrolledDecomposedUnpaired2025}.
    \item \textbf{Generative (L4):} Restoration is treated as a posterior sampling task, leveraging pretrained Image Diffusion Models (e.g., DiffIR2VR-Zero \cite{yehDiffIR2VRZeroZeroShotVideo2025}) or generative priors \cite{Srinath:undive:2025, NEURIPS2022_95504595}.
    \item \textbf{Zero-Shot/Test-Time (L5):} These frameworks (e.g., Zero-TIG \cite{li2025zerotig}, TempRetinex \cite{Li2025TempRetinex}) adapt trainable parameters directly to the specific degraded video being processed
during inference. {A method is classified as L5 based on whether it performs optimization at test time.}
\end{itemize}
The complete classification mapping of existing literature using this factorized framework is presented in Table \ref{tab:video_taxonomy}.

\subsubsection*{Handling Taxonomic Ambiguity}
{To resolve the taxonomic overlap inherent in the five high-level strategy categories (Sections \mbox{\ref{sec:domain-transl} - \ref{sec:prior}}) where, for instance, a cycle-consistency approach might conceptually cover both domain-translation and consistency-based methods, we introduce the three-axis taxonomy to factorize these hybrid architectures. The relationship between the two taxonomies distinguishes conceptual intent from structural mechanics: the five broad categories group the literature by its primary design inspiration, whereas the three axes decouple the actual mathematical implementation in attempt to eliminate ambiguity.

To translate a method into the precise three-axis framework, classification is assigned according to the dominant mathematical mechanism driving the network updates. For a composite objective function $\mathcal{L} = \lambda_1 \mathcal{L}_{primary} + \lambda_2 \mathcal{L}_{auxiliary}$, the supervision signal (S) is dictated strictly by $\mathcal{L}_{primary}$, which acts as the core constraint substituting for paired data, isolated via reported ablation studies and the level of novelty.


Finally, while methods may incorporate various temporal modules, their classification on the T-axis strictly reflects the primary mechanism used for inter-frame aggregation during the forward pass. For the learning paradigm (L), dynamic inference mechanisms override static architectural definitions. If a method solves an optimization problem $\theta^* = \arg\min_{\theta} \mathcal{L}(\theta; X_t)$ at inference time by computing gradient updates $\nabla_\theta \mathcal{L}$ on the input sequence $X_t$ to adapt network parameters $\theta$, it is strictly classified as $L_5$ (Zero-shot/Test-time), regardless of whether the underlying network relies on a feed-forward ($L_1$) or generative ($L_4$) structure.}
\subsubsection{Discussion on Multi-dimensional taxonomy}
\label{subsubsec:discussion}

The multi-dimensional taxonomy is illustrated in the parallel categories plot in Figure \ref{fig:multidim_taxonomy}, which shows that supervision signals are closely tied to the specific restoration task. Internal signals (S1) are primarily used in denoising, where methods such as UDVD \cite{shethUnsupervisedDeepVideo2021} and RDRF \cite{wangRecurrentSelfSupervisedVideo2023} exploit spatio-temporal redundancies to distinguish noise from signal. In contrast, physical consistency (S3) and prior-driven (S4) signals are more common in deblurring and low-light enhancement, e.g. Reblur2Deblur \cite{chenReblur2DeblurDeblurringVideos2018} uses a motion-based blur model to enforce consistency and Zero-TIG \cite{li2025zerotig} relies on physics-based illumination priors for regularization.

{Implicit or window-based temporal modeling (T1) can be combined with either feed-forward training or test-time adaptation; standard TAP~\mbox{\cite{fuTemporalPluginUnsupervised2024}} use T1 with L1, while its TAP-T variant performs test-time adaptation and therefore falls under L5. In contrast, SGZ~\mbox{\cite{zhengSemanticGuidedZeroShotLearning2022}} processes video frame-by-frame (T0) using an offline-trained feed-forward model (L1).}
Conversely, recurrent architectures (T3), such as those developed by Fuoli et al. \cite{fuoliEfficientRecurrentAdversarial2023} and Patil et al. \cite{patilUnpairedRecurrentLearning2025}, are often combined with adversarial (L2) learning to propagate information across long sequences and ensure natural textures. Recent research, exemplified by DiffIR2VR-Zero \cite{yehDiffIR2VRZeroZeroShotVideo2025}, indicates an increasing use of generative paradigms (L4) that adapt image-based foundations for video through explicit alignment (T2). Hybrid frameworks further illustrate how different design choices are combined to improve performance: STBN  \cite{chen2025spatiotemporal} synthesizes self-supervision with optical flow for stable denoising, UDU-Net \cite{zhuUnrolledDecomposedUnpaired2025} unrolls physical priors into an optimization framework, and UnDIVE \cite{Srinath:undive:2025} pairs generative diffusion priors with motion-aware warping for underwater scenes.

\section{Loss Functions for Unsupervised Video Restoration and Enhancement}
\label{sec:loss}

The loss function plays a central role in deep learning by quantifying the discrepancy between predicted and target outputs, serving as the objective that guides model optimization. It provides the feedback signal for updating parameters via gradient-based methods. The choice of loss is task-specific, e.g., regression, classification, or restoration, and critically affects both convergence and generalization. Without a well-defined loss, learning would lack direction and evaluative feedback. In unsupervised learning, the design of loss functions is especially critical, as explicit ground-truth supervision is absent. Instead, learning relies on indirect signals derived from structural assumptions, internal consistency, or self-supervision. Loss functions in this setting must capture meaningful constraints, such as temporal coherence, spatial regularity, perceptual similarity, or physical consistency with degradation models. This section reviews the loss functions commonly used in unsupervised video restoration and enhancement.

\begin{figure}
    \centering
    \includegraphics[width=\columnwidth]{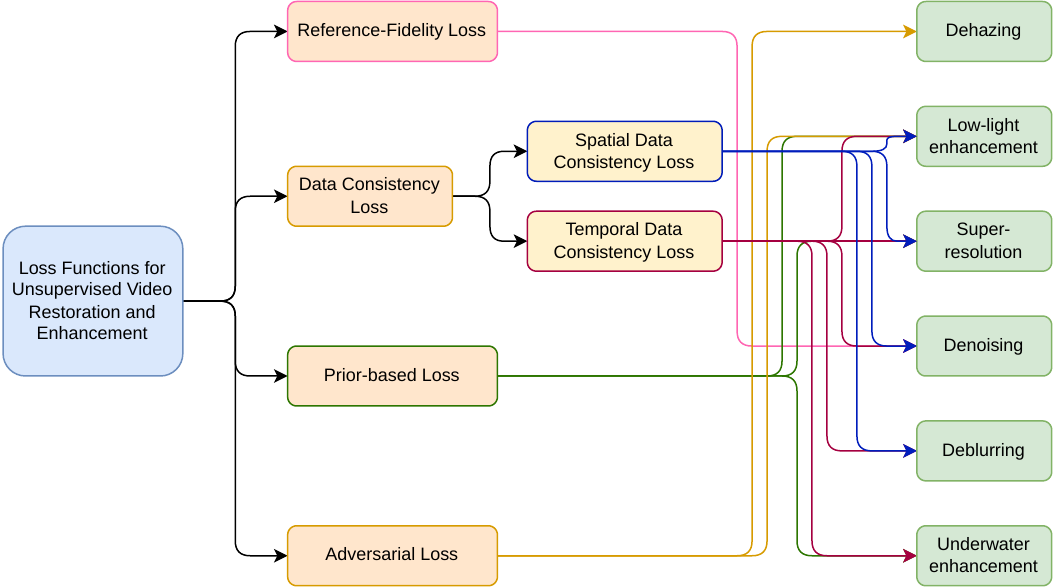}
    \vspace{1pt}
    \caption{Hierarchical taxonomy of loss functions for video restoration and enhancement. The diagram illustrates the losses discussed in Section \ref{sec:loss}, categorizing the literature into four primary strategies. In the final stage of the hierarchy, the arrows indicate which restoration tasks use each loss family. }
    \label{fig:taxonomy_loss}
\end{figure}


\subsection{Pseudo-Reference-Fidelity Losses}
In a supervised setting, a reference-fidelity loss (sometimes also called reconstruction loss) penalizes the pixel-wise difference between ground truth and network's prediction. The most common examples are $\mathcal{L}_1$ (Mean Absolute Error, MAE) and $\mathcal{L}_2$ (Mean Squared Error, MSE) losses:

\begin{multline}
    \mathcal{L}_1 = \frac{1}{N}\sum_{x,y}|I_{pred}(x,y)-I_{gt}(x,y)|, \\ 
    \mathcal{L}_2 = \frac{1}{N}\sum_{x,y}(I_{pred}(x,y)-I_{gt}(x,y))^2,
\end{multline}
where $I_{pred},I_{gt}$ are prediction and groundtruth correspondingly and $N$ is the number of pixels in them.

Although being a fundamental part of supervised learning, reconstruction losses are still sometimes used unsupervised image and video enhancement in cases when the reference is created on-the-fly. For instance, in Recorrupted-to-Recorrupted adapted for videos \cite{zhengUnsupervisedDeepVideo2023}, $\mathcal{L}_2$ loss is calculated between a recorrupted version of an original distorted video and a network's output calculated for a second, independently recorrupted instance of the same distorted video. In \cite{leeRestoreRestoredVideo2021} conventional losses $\mathcal{L}_1$ and $\mathcal{L}_2$ are used to minimize the error between the network output and a pseudo-clean target --- video frames denoised with a pre-trained denoising. Similarly, a reconstruction loss in Temporal As a Plugin (TAP) \cite{fuTemporalPluginUnsupervised2024} minimizes the pixel-wise absolute difference between the video denoiser’s output and the pseudo-clean frames created by denoising followed by re-corruption. 

Reference-fidelity loss functions can provide a pixel-level guidance even in the absence of groundtruth; however, they lack absolute correctness: their performance capped by the quality of the pseudo-reference and the fidelity of the assumed corruption model.

\subsection{Data Consistency Losses}

A \textit{data consistency loss} ensures that the enhanced video remains faithful to certain properties of the input video. Formally, data consistency loss explicitly penalizes the discrepancy between the enhanced video frame $\hat{X_t}$ and the input frame $X_t$ after application of the known operator $A$: in its simple form, it can be writen as $$\mathcal{L}_{DC}=||A(\hat{X_t})-Y_t||_p.$$ The choice of the operator $A$ is motivated by the known imaging process and the specific degradation or transformation applied during data acquisition such as blur kernels in deblurring, optical flow in temporal alignment, or photometric curves in low-light enhancement, so that the data consistency loss faithfully enforces the physical constraints.

Depending on the type of the operator $A$ (per-frame operator or warping that links two frames via observed motion), data consistency losses can be classified into spatial and temporal consistency losses.

\subsubsection{Spatial Data Consistency Losses}

In the super-resolution task, $A$ is simply a downsampling operator~\cite{9286636}. For low-light enhancement
Zheng et al. \cite{zhengSemanticGuidedZeroShotLearning2022} introduce \textit{spatial consistency loss} , that is aimed at preserving local contrast  by penalising differences in the neighboring pixels’ values during enhancement. This helps to preserve edges and avoid distortion of textures.
In Zero-IG \cite{shi2024zero} and Zero-TIG \cite{li2025zerotig}, the self-supervised Noise2Noise-style \cite{lehtinen2018noise2noise} pairwise \textit{residual and consistency loss} terms are calculated between two downsampled noisy versions of the frame, teaching the denoising subnetwork to predict and subtract noise without any clean target. 

\subsubsection{Temporal Data Consistency Losses}
\label{sssec:temp}

The challenging aspect of video enhancement, compared to image-based methods, is making sure the output remains free from temporal artifacts such as flickering or jitter. A common approach is to use optical flow. In UnDIVE \cite{Srinath:undive:2025}, Srinath et al. minimize the difference between the enhanced frame warped via optical flow and the enhancement of the raw frame when warped by that same flow, considering this loss in both forward and backward directions.

To ensure temporal consistency in video denoising, Dewil et al. \cite{dewilSelfsupervisedTrainingBlind2021} introduce multi-frame to frame (MF2F) loss which fine-tunes a video denoiser by using the previous noisy frame as a target for the current frame after applying a warping operator. 

Reblur2Deblur's unsupervised loss is computed as the pixel-wise difference between the original blurry frame  and a reblurred version, that is obtained by applying a motion-based blur kernel, calculated using optical flow between adjacent deblurred frames, to the deblurred frame \cite{chenReblur2DeblurDeblurringVideos2018}. This, although indirectly, promotes temporal consistency and helps to avoid jitter artifacts. 
In ADST-BPN~\cite{10540131}, the difference between two successive frames at both low and high resolutions is used as optimization target.

Data consistency objectives keep solutions physically plausible even without references, however they offer no guidance on ambiguities the forward model cannot resolve, so results can look dull or over-smoothed unless complemented by stronger priors.
 
\subsection{Prior-based Losses}
Unsupervised video enhancing methods often include additional loss terms that encode assumptions about the desirable properties of the latent image. They can depend on additional information from other predictions, such as masks and weights. For instance, Zheng et al. \cite{zhengSemanticGuidedZeroShotLearning2022} introduce \textit{RGB (color) loss} is used to reduce color shifts and is calculated as a Charbonnier distance across color channels, a \textit{brightness loss} is constrains the overall exposure of the output to a desired level and a \textit{total variation (smoothness) loss} that sums squared spatial gradients over RGB channels is used to attenuate residual noise. Similar low-level prior-based constraints are also employed in \cite{lvUnsupervisedLowLightVideo2023} (brightness, smoothness, color) and \cite{Srinath:undive:2025} (smoothness, color).

For a  high-level \textit{semantic loss} Zheng et al. uses a segmentation network: it enforces the enhanced frame to have the same segmented structure as the original one and to preserve scene content and object boundaries. ZVRD \cite{cao2025zeroshot} extracts semantic information using a pretrained CLIP model to ensure semantic similarity between two consecutive frames for video super-resolution task.

For unsupervised low-light video enhancement, additionally to prior-based losses mentioned above, LightenFormer \cite{lvUnsupervisedLowLightVideo2023} uses two unique no-reference losses.  First, an \textit{invertibility loss} leverages the invertibility of the S-curve and ensures that enhancing then inverting a frame recovers the original, which helps to maintain image content consistency during enhancement. Second, a \textit{noise consistency loss} exploits the assumption of noise independence between frames: it encourages the enhancement to not introduce correlated noise across frames.

Prior-based losses penalize implausible content by utilizing statistical information and models of the physical process. However, hand-crafted priors are often too rigid, while learned ones can hallucinate or fail to fit data distribution, their inference is also slower due to iterative character of optimization.

\subsection{Adversarial Losses}
 In a ``vanilla'' GAN, an adversarial loss is a cost function that ensures that the network's generator $G$ produces realistic examples by comparing it against a discriminator $D$ that distinguishes between real and fake data. These two networks play a minimax game, trying to find a solution for the optimization problem $\min_{G}\; \max_{D}\;
\mathcal{L}_{\text{adv}}(D,G)$ \cite{goodfellow2020generative}, where adversarial loss $\mathcal{L}_{\text{adv}}(D,G)$ is defined as follows:

\begin{equation}
\mathcal{L}_{\text{adv}}(D,G)=
 \underbrace{\mathbb{E}_{x\sim p_{\text{data}}}\bigl[\log D(x)\bigr]}_{\text{real term}}
+
\underbrace{\mathbb{E}_{z\sim p_{z}}\bigl[\log\bigl(1-D(G(z))\bigr)\bigr]}_{\text{fake term}}.
\end{equation}

When paired data is unavailable, adversarial loss alone doesn't guarantee an invertible mapping from one domain to another since there are infinite ways to map output. Therefore,
additional loss terms are introduced. CycleGAN \cite{zhuUnpairedImageToImageTranslation2017}, a classic architecture for unpaired domain translation (from domain $A$ to domain $B$), employs two generators $G_{A\to B}$, $G_{B\to A}$ and two discriminators $D_A$ and $D_B$. \textit{Identity loss} and \textit{cycle-consistency loss} ensure content preservation and meaningful translation between input and output (see Fig. \ref{fig:cyclegan}):

\begin{gather}
  \begin{multlined}[b]
    \mathcal{L}_{\text{cyc}}(G_{A \to B}, G_{B \to A}) = \\
    \mathbb{E}_{A_{\text{real}} \sim p_{\text{data}}(A)} \left[ \left\| G_{B \to A}(G_{A \to B}(A_{\text{real}})) - A_{\text{real}} \right\|_1 \right] + \\ 
    \mathbb{E}_{B_{\text{real}} \sim p_{\text{data}}(B)} \left[ \left\| G_{A \to B}(G_{B \to A}(B_{\text{real}})) - B_{\text{real}} \right\|_1 \right]
  \end{multlined} \\[2ex]
  \begin{multlined}[b]
    \mathcal{L}_{\text{identity}}(G_{A \to B}, G_{B \to A}) = \\ 
    \mathbb{E}_{B_{\text{real}} \sim p_{\text{data}}(B)} \left[ \left\| G_{A \to B}(B_{\text{real}}) - B_{\text{real}} \right\|_1 \right] + \\  
    \mathbb{E}_{A_{\text{real}} \sim p_{\text{data}}(A)} \left[ \left\| G_{B \to A}(A_{\text{real}}) - A_{\text{real}} \right\|_1 \right]
  \end{multlined}
\end{gather}

 Adversarial losses aim to push low quality videos onto the manifold of high quality ones. This helps with sharper textures and richer colors than pixelwise losses alone, but it can also make training unstable and make the network to trade data fidelity for style and occasionally hallucinate details.

\section{Synthetic Distorted Data Methods}
\label{sec:syndata}

As noted earlier, unsupervised learning has gained significant attention in video restoration and enhancement tasks due to the scarcity of ground truth data. An alternative approach involves the use of synthetic datasets to enable supervised training. While unsupervised methods do not require ground truth for model learning, objective evaluation, particularly with widely used metrics such as PSNR, SSIM and LPIPS, still relies on paired ground truth data. This emphasizes the continued importance of synthetic datasets.

\subsection{Synthetic Blur}
A prevalent technique for synthesizing motion blur in video datasets involves temporal averaging of frames from high frame rate sequences to simulate the effect of a long camera shutter speed. This approach was effectively utilized by Su et al. \cite{su2017deep} in the creation of DVD dataset and by Nah et al. \cite{Nah_2017_CVPR} for GOPRO dataset. To address the challenges posed by complex, non-linear motion trajectories, the NTIRE 2019 challenge on video deblurring and super-resolution introduced the REDS dataset \cite{nah2019ntire}. This dataset, comprising 300 video clips of 720$\times$1280 resolution, each with 100 frames, advanced the synthesis methodology by employing Convolutional Neural Network (CNN) based frame interpolation. By first upsampling the high frame rate source videos, this technique increases the virtual frame rate, which helps to mitigate unnatural discontinuities or spikes in the subsequent blur trajectory. This process yields a more faithful simulation of motion blur arising from non-linear movements.

Low exposure times causing blur are common in low-light environment. Video captured in low-light conditions often suffer from high noise and low intensity as well as blur in saturated regions. LOL-Blur dataset \cite{zhou2022lednet} addresses this distortions and comprises 200 videos each  60 frames long for joint low-light enhancement and deblurring. The data synthesis pipeline involves simulating low-light effect with an exposure-conditional variant of Zero-DCE \cite{guo2020zero} followed generating blur via frame averaging similarly to previous methods with clipping reverse to produce realistic blur degradation in saturated areas, and finally applying generalized Gaussian filters to imitate defocus blur and adding noise generated by CycleISP \cite{zamir2020cycleisp}.

To simulate a realistic depth-of-field effect for video sequences, DaBiT \cite{Morris:dabit:2025} introduces a focal blur simulator. The method begins by acquiring per-frame depth maps using the DepthAnything model \cite{Yang:DepthAnything:2024}. A spatially-variant Gaussian blur is then applied, modulated by this depth information. The blur intensity, determined by the kernel size, is scaled symmetrically along the depth axis relative to a specified focal plane. The kernel size is zero for pixels within the focal range and increases to a predefined maximum for regions at the furthest depths from the focal point. This process ensures that elements progressively distant from the focal plane are rendered with an increasing level of blur, faithfully replicating a realistic focal blur effect. Alternatively, ELVIS \cite{Lin2026ELVIS} generates blur degradation using multivariate Gaussian distributions. This aims to combine defocus blur with motion blur.

\subsection{Synthetic Noise}

The most common form of noise synthesis is additive Gaussian noise with zero mean and standard deviation $\sigma$, where larger values of $\sigma$ correspond to higher noise levels. This noise is applied independently to each frame of the video sequence. However, while numerous pipelines exist for synthesizing realistic camera sensor noise in static images, dedicated frameworks for video noise synthesis remain largely unexplored. 

Existing methods predominantly focus on modeling noise in RAW image data, with comparatively fewer approaches addressing noise characteristics specific to the sRGB color space. In the domain of RAW image noise synthesis, physics-informed statistical distributions are commonly employed. For instance, Li et al. \cite{li2025dualdn} utilized the well-established Poisson-Gaussian model to simulate the combined effects of signal-dependent shot noise and signal-independent read noise. Further refining these physical models, Wei et al. \cite{wei2021eld} conducted an extensive calibration study across five different camera sensors. Their work proposed replacing the conventional Gaussian distribution with a Tukey-Lambda distribution to more accurately model read noise. In contrast, Zhang et al. \cite{Zhang2023generalrawnoise} introduced a more generalized approach by modeling signal-dependent and signal-independent noise components separately. To ensure the synthesized noise distribution aligns with real-world sensor data, they proposed a novel domain alignment strategy. This strategy leverages both a pre-trained denoiser and a Fourier Transformer discriminator to refine the final combined noise map. A simple method proposed in \cite{malyugina2025marine} generates pseudo noise artifacts characteristic of underwater environments, commonly resembling snowflakes caused by biological debris (marine snow). The approach estimates marine snow by subtracting the median of multiple frames from a frame containing such artifacts. These extracted noise patterns are then added back to clean training data to simulate realistic underwater conditions.

For synthesizing noise in the sRGB domain, learning-based approaches have been developed to model the complex post-processing effects inherent to this color space. Kousha et al. \cite{Kousha2022sRGBflow} introduced a method using normalizing flows, specifically with proposed conditional linear flows. This technique conditions the network on camera metadata, such as ISO and gain settings, enabling it to learn and reproduce the distinct noise characteristics associated with different camera configurations.
In contrast, Fu et al. \cite{Fu2023sRGBncnoise} developed a camera-specific model that eliminates the need for explicit metadata. Their pipeline incorporates a gain estimation network and leverages both local and global image information to effectively model spatially-correlated noise patterns.

More recently, Zheng et al. \cite{zheng2024senmvae} proposed a semi-supervised method based on a variational autoencoder (VAE). Their model, SenMVAE, utilizes separate latent variables, $\mathbf{z}$ and $\mathbf{z_n}$ to disentangle image content from degradation information within the latent space, and used to construct the clean input $\mathbf{x}$ and noisy output $\mathbf{y}$. A key component of their framework is a degradation level prediction network, which allows the generation process to be conditioned on a desired noise intensity, enabling controllable and varied degradation levels.

Despite the plethora of synthetic noise pipelines for images, there are significantly fewer for videos. Addressing this gap, Monakhova et al. \cite{Monakhova2022starlight} introduced a GAN-based pipeline to synthesize noisy image bursts from single, clean source images. Their generator architecture uniquely incorporates learnable parameters to sample noise from physics-informed distributions, which is then processed by a U-Net to model more complex and spatially varying noise characteristics. Lin et al. \cite{lin2025generallowlight} approached the task in an unsupervised approach for general-use; by designing a network which estimates the noise parameters to accurately simulate the noise present in the video by randomly selecting the parameters and generating noisy inputs, and training their network to predict the input parameters.

\subsection{Synthetic Low Light}

Synthesizing low-light content aims to reduce the illumination of the input image/video to simulate the lack of photons in the scene. Naive solutions involve applying linear and gamma transformations:  $y = \alpha x^\gamma + \beta$ (or variants) \cite{7351548,lin2025generallowlight, lv2021agllie} to modify the range of values from the input. However, this results in unrealistic low-light outputs. Yang et al. \cite{yang2021lolv2synth} released a synthetic low-light dataset by gathering normal-light and low-light images from multiple datasets, converting them into YCrCb color space, and applying histogram matching in the Y channel to the normal-light images.

Deep-learning methods have gained popularity as they allow for spatially-varying illumination, which is difficult for traditional methods that do not use contextual information. Triatafyllidou et al. \cite{triantafyllidou2020sidgan} employed two CycleGANs \cite{zhuUnpairedImageToImageTranslation2017} to map videos (domain A) to short-exposure still images (domain C) via an intermediate long-exposure stills domain (domain B). The network that maps domain A to domain B uses unpaired data, while the network that maps domain B to domain C uses paired long/short-exposure images from the DRV dataset \cite{chen2019smid}. Inspired by the low-light enhancement method Zero-DCE \cite{guo2020zero}, Zhou et al. \cite{zhou2022lednet} developed a synthetic low-light network, EC-Zero-DCE, which maps videos from normal-light to low-light by implementing a reversed curve adjustment. This network is conditioned by a parameter to control the illumination level of the output. Recently, the ELVIS \cite{Lin2026ELVIS} framework offers a synthetic low-light pipeline using learning-based degradation parameter estimation, trained on large-scale low-light datasets.

\subsection{Synthetic Light Attenuation} 
\label{ssec:synlightatt}
Both haze and underwater image models are similarly defined as shown in Eq.~\ref{eqn:light}. In hazy environments, the transmission map $t(x)$ is modeled as an exponential decay, $e^{-kd(x)}$, where $d(x)$ denotes the scene depth and $k$ is the atmospheric scattering coefficient. A widely used synthetic dataset, HazeRD \cite{Zhang:HazeRD:2027}, simulates hazy effects based on Eq.\ref{eqn:light}, incorporating the inverse of the color transformation into the channel sensitivities to model spectral absorption. More recent approaches leverage conditional diffusion models. For example, HazeGen \cite{Wang_2025_CVPR} integrates physical scattering priors and is trained on synthetic data to generate more realistic haze-like images. 
Xu et al.~\cite{xuVideoDehazingMultiRange2023} generate hazy video data by applying atmospheric scattering model (Eq. 1) to videos from pre-existing datasets for downstream computer vision tasks: scene understanding \cite{cordts2016cityscapes}, depth estimation \cite{guizilini20203d}, detection and tracking \cite{wen2020ua,zhu2021detection}, segmentation \cite{PontTuset2017DAVIS}, deblurring and super-resolution \cite{nah2019ntire}. To ensure temporal consistency, robust video depth estimation is used. This results in a large-scale synthetic dataset with over 5000 outdoor video sequences for various scenarios.
Zhang et al.~\cite{Zhang_2021_CVPR} introduced a hazy video dataset captured using a haze machine and repeatable camera motion facilitated by a robotic arm. These sequences, however, are limited to indoor environments. 

For underwater imagery, $t(x)$ is dependent to wavelength $\lambda$ as well, $t_\lambda$(x) = 10$^{-\beta_\lambda d(x)}$, where $\beta_\lambda$ is the wavelength-depended medium attenuation coefficient. Scattering in the water medium is typically approximated using the Jaffe–McGlamery model \cite{jaffe1990computer, mcglamery1980computer}, and attenuation coefficients are adjusted to reflect different water types \cite{Li:WaterGAN:2017}. For learning-based approaches, Du et al.\cite{du2024end} generated a synthetic underwater video dataset using the pretrained Underwater Neural Rendering (UWNR) model \cite{Ye_2022_CVPR}, where depth maps are combined with a fixed light field map. The model is trained on the real UIEB dataset~\cite{LiUnderwater:2020}. To ensure temporal consistency, all frames within a sequence share the same light field, derived from a single real underwater image.

\subsection{Synthetic Atmospheric Turbulence}

A common synthetic approach for simulating turbulence distortions applies predefined point spread functions (PSFs), akin to blurring. Hirsch et al. \cite{hirsch2010efficient} introduced spatially varying degradation by applying PSFs to image patches. Chak et al. \cite{chak2021subsampled} extended this by randomly generating motion maps to warp clean images, followed by Gaussian blur. However, many simulation methods model atmospheric turbulence in the phase domain because a medium that is turbulent induces fluctuations in phase. For instance, Hardie et al. \cite{hardie2017simulation} employed a modified Von Kármán spectrum model to better represent phase distortion. Learning-based simulators have also emerged, including CNN-based models \cite{9711075} and GAN-driven approaches \cite{10023498}. However, as reported in the recent review by Hill et al. \cite{hill2025deep}, the simulator proposed in \cite{chimitt2020simulating} achieves superior results. This model uses Zernike polynomials to simulate phase distortions as a function of focal length, frequency, aperture, and a random vector, introducing random tilt to the PSF. While effective for weak turbulence, their simulations tend to overestimate blur and spatial correlation under stronger conditions.

There have been efforts to replicate turbulent conditions using real heat sources. For example, gas hobs were employed in \cite{anantrasirichai2013atmospheric}, with varying numbers of hobs generating different levels of distortion. Similarly, multiple heat chambers were used in \cite{Mao_2022_ECCV} to produce indoor turbulence sequences. However, these real-heat approaches still oversimplify the complexity of atmospheric turbulence. A primary limitation lies in the short propagation distance of light rays through the heated medium, which does not adequately reflect the characteristics of real outdoor turbulence. Moreover, critical environmental factors, such as humidity fluctuations and airborne particles like dust, are not considered. Additionally, the resulting datasets are limited to static scenes, lacking temporal and spatial diversity present in dynamic real-world scenarios.

\section{Current Performance and Discussion}
\label{sec:performance}

This section reviews the performance of existing unsupervised methods for video restoration and enhancement. {While our taxonomy covers a broader range of applications, including video super-resolution and underwater enhancement, we restrict our quantitative benchmarking to the four most established tasks. We omit formal evaluation for super resolution and underwater enhancement because, as discussed later in this section, unsupervised video-specific methods in these domains remain significantly underexplored and currently lack standardized comparative baselines.} Since most approaches are tailored to specific problem settings, we organize the evaluation around four representative tasks: (i) deblurring, (ii) denoising, (iii) low-light enhancement, and (iv) dehazing. Given the limited number of methods developed explicitly for video, we also report results from image-based approaches as a reference. Supervised methods are included only to indicate an approximate upper bound on achievable performance. Through comparative analysis across multiple datasets, evaluation metrics, and methodological categories, several notable trends and insights can be observed.

{Quantitative comparisons across unsupervised methods must be interpreted with caution. Reported metrics often stem from heterogeneous evaluation protocols, differing synthetic degradation models, and varied implementation environments, meaning these tables represent a synthesis of literature rather than a strict, standardized leaderboard. Furthermore, all supervised methods included in these tables operate with full access to paired ground truth; they are presented strictly as approximate performance upper bounds rather than direct competitors to the unsupervised frameworks.}
\subsection{Deblurring}
\label{ssec:deblurring}

The comparative results on the standard GoPro dataset~\cite{Nah_2017_CVPR} are presented in Table \ref{tab:supervised_unsupervised} for video debluring performance. The dataset comprises 3,214 paired blurry and sharp images at a resolution of 1280$\times$720 pixels. As paired ground truth is required, these annotations are used exclusively for supervised methods and quantitative evaluation. Following the standard protocol, 1,111 image pairs are reserved for testing, with the remaining samples used for training.

The supervised methods, consistently achieve higher quantitative performance, reflecting the advantage of direct access to paired sharp–blurred training data. Recent supervised approaches such as Shift-Net+ and BSSTNet achieve the strongest results overall, with BSSTNet attaining the highest PSNR (35.98 dB) and SSIM (0.9792). These gains align with broader trends in the literature, where increasingly sophisticated architectures and loss formulations have led to steady improvements in supervised deblurring performance.
In contrast, unsupervised methods exhibit a wider performance spread and generally lower absolute scores. Early unpaired approaches based on cycle-consistency constraints, such as CycleGAN and DualGAN, achieve relatively modest PSNR and SSIM, indicating the difficulty of learning accurate inverse mappings without explicit supervision. 
While most unsupervised approaches rely on GAN frameworks, UAUD adopts an ADMM-based formulation, highlighting alternative directions for enforcing consistency and regularization in the absence of ground truth. The improved performance of recent unsupervised methods suggests that incorporating stronger priors, optimization constraints, or domain-specific assumptions can partially compensate for the lack of supervision. Although a noticeable gap in PSNR remains between supervised and unsupervised methods, the difference in SSIM is considerably smaller, indicating that structural fidelity is increasingly well preserved despite the absence of supervision.

For the closely related task of video super-resolution, which shares similar challenges with video deblurring, unsupervised methods remain relatively underexplored. As research progresses through 2026, only a small number of approaches explicitly address unsupervised video super-resolution. One line of work adopts a plug-and-play ADMM framework with a pretrained DnCNN denoiser~\cite{10401873}, while another synthesizes downsampled observations during optimization~\cite{10428075}, following a re-degradation principle similar to that discussed in Section~\ref{sec:degradation}. 
Although these methods report improvements over earlier supervised CNN-based baselines such as EDVR~\cite{Wang2019EDVR}, they have not yet been systematically evaluated against more recent architectures, including transformer-based models such as VRT~\cite{10462902}, leaving their relative effectiveness unclear.

\subsection{Denoising}
\label{ssec:denoising}

We follow the standard evaluation protocol for video denoising by injecting synthetic Gaussian noise at varying noise levels and evaluating performance on two widely used benchmarks, Set8~\cite{9156652} and DAVIS~\cite{PontTuset2017DAVIS}. Tables~\ref{tab:set8_davis_unsup} show denoising performance of the current methods. A clear trend emerges across both datasets: performance gaps between unsupervised and supervised methods have narrowed substantially, particularly at moderate to high noise levels. While supervised models such as PaCNet, FloRNN, RVRT, and VRT still achieve the strongest absolute PSNR, several recent unsupervised approaches achieve comparable structural fidelity and perceptual quality, especially on DAVIS. Among the unsupervised methods, STBN consistently achieves the best SSIM across all noise levels and both datasets, indicating superior structural preservation. 

Interestingly, recent supervised video denoising methods predominantly adopt transformer-based architectures, whereas unsupervised approaches remain largely CNN-based. This divergence may be attributed to the nature of the denoising task itself: noise primarily affects low-level signal statistics rather than high-level semantic representations. In the absence of ground-truth supervision, convolutional architectures may therefore be better suited to act as adaptive filtering operators, exploiting local spatial–temporal correlations while maintaining training stability. By contrast, transformer models, which excel at modeling long-range dependencies and semantic context, may require explicit supervision to fully realize their representational capacity, making their unsupervised training more challenging.

\subsection{Low Light Enhancement}
\label{ssec:lowlight}

Table \ref{tab:evaluationlowlight} presents a comprehensive quantitative comparison of unsupervised low-light enhancement methods on two widely adopted benchmarks, BVI-RLV \cite{Lin:BVIRLV:2024} and DID \cite{Fu_2023_ICCV}. Both image-based and video-based approaches are included.

The literature reveals a distinct scarcity of low-light video enhancement frameworks relative to image-based alternatives, primarily due to the architectural demands of dynamic temporal constraints. 
Across datasets, newer unsupervised methods clearly outperform earlier curve-based and GAN-based models, especially on perceptual metrics like LPIPS. Early approaches such as Zero-DCE and EnlightenGAN showed the viability of unsupervised learning but lag behind in structural quality and temporal stability. The best results come from video-based methods like TempRetinex and Zero-TIG, indicating that explicit temporal modeling and illumination consistency significantly improve performance even without supervision.

As perceived brightness is inherently subjective and can substantially influence PSNR values, Histogram Matching (HM) is applied to mitigate this bias and to place greater emphasis on evaluating low-light noise reduction. HM exhibits a systematic and often significant impact on performance across nearly all unsupervised methods. On both datasets, HM leads to notable improvements in PSNR and SSIM, reflecting enhanced global brightness alignment and contrast consistency. Its effect on LPIPS, however, is more nuanced. While stronger models such as TempRetinex and Wakeup-Darkness benefit consistently from HM, weaker approaches often show marginal or inconsistent gains, indicating that global histogram alignment alone cannot compensate for limited model capacity. HM therefore acts as a strong post-processing prior whose effectiveness is highly dependent on the underlying enhancement model.

Similar to denoising task, transformer-based architectures achieve strong performance in supervised low-light enhancement. In contrast, existing unsupervised methods predominantly rely on conventional CNN-based designs. As noted in TempRetinex \cite{Li2025TempRetinex}, incorporating attention mechanisms or replacing convolutional layers with transformer or Mamba blocks~\cite{gu2023mamba} does not yield performance improvements under unsupervised training, while substantially increasing inference cost. Although video-based approaches explicitly enforce temporal consistency through dedicated losses (see Section \ref{sssec:temp}), temporal flickering remains observable, as reflected by the Mean Absolute Brightness Difference (MABD) between consecutive frames.

This suggests that temporal modeling in unsupervised low-light enhancement remains an open challenge. The study in \cite{Lin2025Dynamic}, although conducted in a supervised setting, reveals that recurrent weights applied to previously enhanced frames effectively act as exponentially decaying weights over long temporal histories, thereby reducing heavy noise. This mechanism has the potential to be extended to unsupervised methods, particularly in scenarios where memory is limited and the number of accessible temporal inputs must be constrained.

\begin{landscape}

\begin{table}[t]
\centering
\caption{Comparison of deblurring methods GoPro dataset~\cite{Nah_2017_CVPR}. Red, orange, and yellow denote the best, second-best, and third-best results.}
\label{tab:supervised_unsupervised}
\resizebox{\linewidth}{!}{
{%
\begin{tabular}{lccccr|lcccccr}
\toprule
\multicolumn{6}{c|}{\textbf{Supervised methods}} 
& \multicolumn{7}{c}{\textbf{Unsupervised methods}} \\
\cmidrule(lr){1-6} \cmidrule(lr){7-13}

Method & Year & PSNR$\uparrow$ & SSIM$\uparrow$ & Params (M) / GFLOPs & Runtime (Res, GPU) & Method & Year & PSNR$\uparrow$ & SSIM$\uparrow$ & Architecture & Params (M) / GFLOPs & Runtime (Res, GPU) \\
\midrule

SRN~\cite{tao2018srndeblur}
& 2018 & 30.26 & 0.934& $\sim$10.2 / $\sim$1438 & 200 ms ($1280\times720$, Titan X) 
& CycleGAN~\cite{zhuUnpairedImageToImageTranslation2017}
& 2017 & 22.54 & 0.720 & GAN & 11.38 / 56.8 & $\sim$45 ms ($256\times256$, Titan X) \\

NAFNet~\cite{Chen2022SimpleBaselines}
& 2022 & 33.69 & 0.9670 & -- / 63.3 & 177.1 ms ($1280\times720$, RTX 2080 Ti) 
& DualGAN~\cite{8237572} 
& 2017 & 22.86 & 0.722 & GAN & -- / -- & -- ($256\times256$, Titan X) \\

RVRT~\cite{Liang2022RVRT}
& 2022 & \cellcolor{third}{34.92} & \cellcolor{third}{0.9738} & 13.6 / 88.8 & 300 ms/f ($1280\times720$, RTX 3090) & UID-GAN~\cite{8935440}
& 2020 & 23.56 & 0.738 & GAN & -- / -- & -- (--, --) \\

VRT~\cite{10462902}
& 2024 & 34.81 & 0.9724 & 18.3 / 948 & 2.2 s/f ($1280\times720$, RTX 3090) 
& FCL-GAN~\cite{FCLGAN:2022}
& 2022 & \cellcolor{third}{24.84} & \cellcolor{third}{0.771} & GAN & -- / -- & 11.4 ms ($1280\times720$, RTX 2080 Ti) \\

Shift-Net+~\cite{li2023simple}
& 2023 & \cellcolor{second}{35.88} & \cellcolor{second}{0.9790} & 12.3 / 146 & 45 ms/f ($256\times256$, RTX 3090) 
& UAUD~\cite{10204475}
& 2023 & \cellcolor{second}{26.12} & \cellcolor{second}{0.869} & ADMM & -- / -- & -- (--, RTX 3090) \\

BSSTNet~\cite{10657840}
& 2024 & \cellcolor{best}{35.98} & \cellcolor{best}{0.9792} & 14.1 / 133 & 28 ms/f ($256\times256$, RTX 3090) 
& SEMGUD~\cite{Chen_2024_CVPR}
& 2024 & \cellcolor{best}{29.06} & \cellcolor{best}{0.927} & GAN & -- / -- & -- (--, RTX 3090) \\
\bottomrule

\bottomrule
\end{tabular}
}
}
\end{table}

\begin{table}[tb]
\centering
\caption{Performance comparison of video denoising methods on the Set8~\cite{9156652} and DAVIS~\cite{PontTuset2017DAVIS} datasets. Red, orange, and yellow denote the best, second-best, and third-best results among unsupervised methods, respectively. {Empty cells indicate that the corresponding metrics were not reported by the original authors under these specific evaluation protocols.} $^*$ denotes supervised methods; supervised results are listed for reference.}
\label{tab:set8_davis_unsup}
\resizebox{\linewidth}{!}{%
{%
\begin{tabular}{l|c|cccccccccc|cccccccccc|c|c|c|c|c}
\toprule
& & \multicolumn{10}{c|}{Set8} & \multicolumn{10}{c|}{DAVIS} & & & & & \\
\cmidrule(lr){3-12} \cmidrule(lr){13-22}
Method & Year 
& \multicolumn{2}{c}{$\sigma=10$}
& \multicolumn{2}{c}{20}
& \multicolumn{2}{c}{30}
& \multicolumn{2}{c}{40}
& \multicolumn{2}{c|}{50}
& \multicolumn{2}{c}{$\sigma=10$}
& \multicolumn{2}{c}{20}
& \multicolumn{2}{c}{30}
& \multicolumn{2}{c}{40}
& \multicolumn{2}{c|}{50} & Architecture & Technique & Params (M) & GFLOPs & Runtime (Res, GPU) \\
\cmidrule(lr){3-4}\cmidrule(lr){5-6}\cmidrule(lr){7-8}\cmidrule(lr){9-10}\cmidrule(lr){11-12}
\cmidrule(lr){13-14}\cmidrule(lr){15-16}\cmidrule(lr){17-18}\cmidrule(lr){19-20}\cmidrule(lr){21-22}
& & PSNR$\uparrow$ & SSIM$\uparrow$
& PSNR$\uparrow$ & SSIM$\uparrow$
& PSNR$\uparrow$ & SSIM$\uparrow$
& PSNR$\uparrow$ & SSIM$\uparrow$
& PSNR$\uparrow$ & SSIM$\uparrow$
& PSNR$\uparrow$ & SSIM$\uparrow$
& PSNR$\uparrow$ & SSIM$\uparrow$
& PSNR$\uparrow$ & SSIM$\uparrow$
& PSNR$\uparrow$ & SSIM$\uparrow$
& PSNR$\uparrow$ & SSIM$\uparrow$ & & & & & \\
\midrule
 {PaCNet$^*$~\cite{Vaksman_2021_ICCV}} &  {2021}
&  {37.06} &  {0.960} &  {33.94} &  {0.925} &  {32.05} &  {0.892} &  {30.70} &  {0.862} &  {29.66} &  {0.835}
&  {39.97} &  {0.971} &  {37.10} &  {0.947} &  {35.07} &  {0.921} &  {33.57} &  {0.897} &  {32.39} &  {0.874}
&  {CNN} &  {Supervised learning} & 2.87 & 1.51 & $\sim$30 s/f ($854\times480$, Quadro RTX 8000) \\

 {FloRNN$^*$~\cite{li2022unidirectional}} &  {2022}
&  {37.57} &  {0.964} &  {34.67} &  {0.938} &  {32.97} &  {0.914} &  {31.75} &  {0.891} &  {30.80} &  {0.870}
&  {40.16} &  {0.976} &  {37.52} &  {0.956} &  {35.89} &  {0.944} &  {34.66} &  {0.929} &  {33.67} &  {0.913}
&  {CNN} &  {Supervised learning} & 1.85 & -- & -- (--, --) \\

 {RVRT$^*$~\cite{Liang2022RVRT}} &  {2022}
&  {37.53} &  {--} &  {34.83} &  {--} &  {33.30} &  {--} &  {32.21} &  {--} &  {31.33} &  {--}
&  {40.57} &  {--} &  {38.05} &  {--} &  {36.57} &  {--} &  {35.47} &  {--} &  {34.57} &  {--}
&  {Transformer} &  {Supervised learning} & 12.8 & -- & 0.2 s/f ($1280 \times 720$, --) \\

 {{VRT$^*$}~\cite{10462902}} &  {2024}
&  {37.88} &  {--} &  {35.02} &  {--}
&  {33.35} &  {--} &  {32.15} &  {--} &  {31.22} &  {--}
&  {40.82} &  {--}
&  {38.15} &  {--}
&  {36.52} &  {--}
&  {35.32} &  {--}
&  {34.36} &  {--}
&  {Transformer} &  {Supervised learning} & 18.4 & -- & 1.5 s/f ($1280 \times 720$, --) \\

\hline
MF2F~\cite{dewilSelfsupervisedTrainingBlind2021} & 2021
& 36.01 & 0.9379 & 33.79 & 0.9115 & 32.20 & 0.8831 & 30.64 & 0.8413 & 28.90 & 0.7775
& 38.04 & 0.9566 & 35.61 & 0.9359 & 33.65 & 0.9065 & 31.50 & 0.8523 & 29.39 & 0.7843
& CNN & blind-spot & -- & -- & -- (--, --) \\

UDVD~\cite{shethUnsupervisedDeepVideo2021} & 2021
& 36.36 & 0.9510 & 33.53 & 0.9167 & 31.88 & 0.8865 & 30.72 & 0.8595 & 29.81 & 0.8349
& 39.17 & \cellcolor{third}0.9700 & 35.94 & 0.9428 & 34.09 & 0.9178 & 32.79 & 0.8949 & 31.80 & 0.8739
& CNN & blind-spot \& self-ensemble & -- & -- & -- (--, --) \\

RDRF~\cite{wangRecurrentSelfSupervisedVideo2023} & 2023
& {36.67} & \cellcolor{third}{0.9547}
& {34.00} & \cellcolor{second}{0.9251}
& \cellcolor{third}{32.39} & \cellcolor{third}{0.8978}
& \cellcolor{second}{31.23} & \cellcolor{third}{0.8725}
& \cellcolor{second}{30.31} & \cellcolor{third}{0.8490}
& \cellcolor{third}{39.54} & \cellcolor{second}{0.9717}
& {36.40} & \cellcolor{third}{0.9473}
& {34.55} & \cellcolor{third}{0.9245}
& {33.23} & \cellcolor{second}{0.9032}
& {32.20} & \cellcolor{third}{0.8832}
& CNN & blind-spot & -- & -- & -- (--, --) \\

VER2R~\cite{zhengUnsupervisedDeepVideo2023} & 2023
& \cellcolor{second}{37.55} & --
& \cellcolor{third}{34.34} & --
& \cellcolor{second}{32.45} & --
& {31.09} & --
& {30.05} & --
& {39.52} & --
& \cellcolor{third}{36.49} & --
& \cellcolor{third}{34.60} & --
& \cellcolor{third}{33.29} & --
& \cellcolor{third}{32.25} & --
& temporal attention & Recorrupted-to-Recorrupted & Untrained & -- & -- (--, --) \\

TAP~\cite{fuTemporalPluginUnsupervised2024} & 2024
& \cellcolor{best}{37.76} & \cellcolor{second}{0.956}
& \cellcolor{best}{34.51} & \cellcolor{third}{0.925}
& \cellcolor{best}{32.76} & \cellcolor{second}{0.899}
& \cellcolor{third}{31.21} & \cellcolor{second}{0.873}
& \cellcolor{third}{30.27} & \cellcolor{second}{0.851}
& \cellcolor{second}{39.69} & \cellcolor{best}{0.972}
& \cellcolor{second}{36.62} & \cellcolor{second}{0.948}
& \cellcolor{second}{34.71} & \cellcolor{second}{0.925}
& \cellcolor{second}{33.37} & \cellcolor{third}{0.903}
& \cellcolor{second}{32.36} & \cellcolor{second}{0.884}
& CNN & Recurrent pseudo pairs & 0.42 (plugin) &  -- & 32ms (--, --) \\

UCVD~\cite{10647922} & 2025
& -- & -- & -- & -- & 29.01 & 0.80 & -- & -- & 27.36 & 0.76
& -- & -- & -- & -- & 28.49 & 0.78 & -- & -- & 28.86 & 0.74
& CNN & Coordinate-based NN & 0.35 & 134 & $\sim$1.2 s/f ($960\times540$, RTX 3090) \\

STBN~\cite{chen2025spatiotemporal} & 2025
& \cellcolor{third}{37.24} & \cellcolor{best}{0.9594}
& \cellcolor{second}{34.41} & \cellcolor{best}{0.9322}
& \cellcolor{best}{32.76} & \cellcolor{best}{0.9072}
& \cellcolor{best}{31.57} & \cellcolor{best}{0.8837}
& \cellcolor{best}{30.62} & \cellcolor{best}{0.8608}
& \cellcolor{best}{40.35} & {0.9613}
& \cellcolor{best}{37.67} & \cellcolor{best}{0.9606}
& \cellcolor{best}{36.00} & \cellcolor{best}{0.9454}
& \cellcolor{best}{34.73} & \cellcolor{best}{0.9296}
& \cellcolor{best}{33.70} & \cellcolor{best}{0.9138}
& CNN & Bidirectional blind-spot & 2.8 & 134 & -- (--, --) \\

\bottomrule
\end{tabular}}
}

\vspace{5mm}

  \centering
  \caption{Quantitative comparison on BVI-RLV \cite{Lin:BVIRLV:2024} and DID \cite{Fu_2023_ICCV} datasets. Video- and image-based methods are both included. For unsupervised methods, we report results both without (w/o) and with (w/) Histogram Matching (HM). Red, orange, and yellow denote the best, second-best, and third-best results among unsupervised methods, respectively. {Empty cells indicate that the corresponding metrics were not reported in the original manuscripts.Computational metrics for the available methods were independently benchmarked by the authors on an NVIDIA RTX 4090 at $1200\times700$ resolution.} $^*$ denotes supervised methods; supervised results are listed for reference.}
  \label{tab:evaluationlowlight}
  
  \setlength{\tabcolsep}{2.5pt}
  
  \resizebox{\linewidth}{!}{%
  {%
  \begin{tabular}{c|c@{\hskip 7pt}|c@{\hskip 7pt}|ccccccc@{\hskip 7pt}|ccccccc|c|c|c|c|c}
    \toprule
    \multicolumn{1}{c}{\multirow{3}{*}{Method}} &
    \multicolumn{1}{c}{\multirow{3}{*}{Year}} &
    \multicolumn{1}{c}{\multirow{3}{*}{Type}} &
    \multicolumn{7}{c}{BVI-RLV} &
    \multicolumn{7}{c}{DID} & \multicolumn{1}{c}{\multirow{3}{*}{Architecture}} & \multicolumn{1}{c}{\multirow{3}{*}{Technique}} & \multicolumn{1}{c}{\multirow{3}{*}{Params (M)}} & \multicolumn{1}{c}{\multirow{3}{*}{GFLOPs}} & \multicolumn{1}{c}{\multirow{3}{*}{Runtime (Res, GPU)}} \\

    & & &
    \multicolumn{3}{c}{w/o HM} &
    \multicolumn{3}{c}{w/ HM} &
    \multirow{2}{*}{MABD $\downarrow$} &
    \multicolumn{3}{c}{w/o HM} &
    \multicolumn{3}{c}{w/ HM} &
    \multirow{2}{*}{MABD $\downarrow$} & & & & & \\

    \cmidrule(lr){4-9} 
    \cmidrule(lr){11-16} 

    & & &
    PSNR$\uparrow$ & SSIM$\uparrow$ & LPIPS$\downarrow$ &
    PSNR$\uparrow$ & SSIM$\uparrow$ & LPIPS$\downarrow$ &
    &
    PSNR$\uparrow$ & SSIM$\uparrow$ & LPIPS$\downarrow$ &
    PSNR$\uparrow$ & SSIM$\uparrow$ & LPIPS$\downarrow$ & & & & & & \\
    \midrule

    
     {SwinIR$^*$ \cite{SwinIR:2021}} & 2021 &  {Image} &  {29.11} &  {0.891} &  {0.275} &  {32.22} &  {0.9326} &  {0.136} &  {--} &  {19.62} &  {0.6334} &  {0.387} &  {34.982} &  {0.9334} &  {0.064} &  {--} &  {Transformer} &  {Supervised learning} & 11.716 & 23088.61 & 4471.1 ms ($1200\times700$, RTX 4090) \\
    
     {Retinexformer$^*$ \cite{retinexformer}} & 2023 &  {Image} &  {32.63} &  {0.906} &  {0.265} &  {--} &  {--} &  {--} &  {--} &  {24.15} &  {0.849} &  {0.216} &  {--} &  {--} &  {--} &  {--} &  {Transformer} &  {Supervised learning} & 1.61 & 15.57 & 32.9 ms ($1200\times700$, RTX 4090) \\
    
     {SDSDNet$^*$ \cite{wang2021seeing}} & 2021 &  {Video} &  {20.69} &  {0.726} &  {0.148} &  {33.15} &  {0.946} &  {0.128} &  {--} &  {21.88} &  {0.834} &  {0.216} &  {34.56} &  {0.937} &  {0.086} &  {--} &  {CNN} &  {Supervised learning} & 4.447 & 1217.91 & 30.5 ms ($1200\times700$, RTX 4090) \\
    
     {BVI-CDM$^*$ \cite{Lin:Low:2024}} & 2024 &  {Video} &  {30.51} &  {0.888} &  {0.089} &  {--} &  {--} &  {--} &  {--} &  {23.88} &  {0.802} &  {0.134} &  {--} &  {--} &  {--} &  {--} &  {DM} &  {Supervised learning} & 22.591 & 2991.56 & 482.8 ms ($1200\times700$, RTX 4090) \\
    
     {BVI-Mamba$^*$ \cite{huang:bvi:2025}} & 2025 &  {Video} &  {31.22} &  {0.912} &  {0.071} &  {--} &  {--} &  {--} &  {--} &  {24.21} &  {0.850} &  {0.168} &  {--} &  {--} &  {--} &  {--} &  {Mamba} &  {Supervised learning} & 10.004 & 8121.14 & 1875.1 ms ($1200\times700$, RTX 4090) \\

    \midrule

    Zero-DCE \cite{guo2020zero} & 2020 & Image
    & 10.54 & 0.430 & 0.528
    & 18.93 & 0.488 & 0.507
    & 16.05
    & 14.08 & 0.686 & 0.404
    & 20.13 & 0.655 & 0.462
    & 10.62 & CNN & Curve adjustment &  0.079 & 84.99 & 3 ms ($1200\times900$, RTX 2080 Ti) \\

    RUAS \cite{liu2021ruas} & 2021 & Image
    & 15.31 & 0.631 & 0.481
    & 18.52 & 0.712 & 0.515
    & 6.70
    & 17.23 & 0.775 & 0.430
    & 22.11 & 0.836 & 0.433
    & 10.33 & GAN & Retinex theory & 0.003 & 5.489 & 8.6ms ($1200\times700$, RTX 2080 Ti) \\

    EnlightenGAN \cite{jiang2021enlightengan} & 2021 & Image
    & 15.49 & 0.518 & 0.515
    & 17.88 & 0.550 & 0.522
    & 12.26
    & \cellcolor{third}19.37 & 0.725 & 0.403
    & 22.93 & 0.754 & 0.402
    & 14.20 & GAN & VGG-based regularization & 8.637 & 273.2 & 8 ms ($1200\times900$, RTX 2080 Ti) \\

    NeRCO \cite{NerCO_2023_ICCV} & 2023 & Image
    & \cellcolor{second}21.05 & 0.705 & 0.431
    & 27.67 & 0.822 & 0.402
    & 7.39
    & \cellcolor{second}20.90 & \cellcolor{second}0.826 & 0.386
    & \cellcolor{third}28.47 & \cellcolor{second}0.879 & 0.372
    & 10.32 & GAN & Vision-language priors & 23.295 & 4882.65 & 401.6 ms ($1200\times700$, RTX 4090) \\

    CoLIE \cite{Colie2024} & 2024 & Image
    & 18.21 & 0.583 & 0.410
    & 24.15 & 0.646 & 0.431
    & 11.50
    & 18.80 & 0.784 & 0.356
    & 26.36 & 0.825 & 0.348
    & 12.44 & CNN & Retinex theory & 0.133 & 17.31 & 663.3 ms ($1200\times700$, RTX 4090) \\

    Zero-IG \cite{shi2024zero} & 2024 & Image
    & 19.37 & 0.639 & 0.398
    & \cellcolor{third}27.84 & \cellcolor{third}0.834 & 0.370
    & \cellcolor{best}5.27
    & 14.25 & 0.785 & 0.380
    & 25.84 & \cellcolor{third}0.854 & 0.386
    & \cellcolor{second}7.03 & CNN & Retinex theory & 0.087 & 268.53 & 18.2 ms ($1200\times700$, RTX 4090) \\

    Wakeup-Darkness \cite{zhang2025Wakeup-Darkness} & 2025 & Image
    & \cellcolor{third}19.83 & \cellcolor{third}0.734 & \cellcolor{second}0.342
    & \cellcolor{second}28.43 & 0.774 & \cellcolor{second}0.348
    & 8.84
    & \cellcolor{best}20.93 & 0.801 & \cellcolor{second}0.323
    & \cellcolor{second}30.12 & 0.852 & \cellcolor{second}0.324
    & 11.11 & CNN & Vision-language priors & 0.001 & 1.93 & 6.5 ms ($1200\times700$, RTX 4090) \\
    \hline
    SGZSL \cite{zhengSemanticGuidedZeroShotLearning2022} & 2022 & Video
    & 13.42 & 0.577 & 0.420
    & 24.03 & 0.723 & 0.380
    & 12.17
    & 15.55 & 0.733 & 0.363
    & 25.05 & 0.836 & \cellcolor{third}0.346
    & 11.68 & CNN & Curve adjustment &  0.011 & 0.120 & 1 ms ($1200\times900$, RTX 2080 Ti) \\
    Zero-TIG \cite{li2025zerotig} & 2025 & Video
    & 19.34 & \cellcolor{second}0.790 & \cellcolor{third}0.360
    & 28.05 & \cellcolor{second}0.854 & \cellcolor{third}0.368
    & \cellcolor{third}6.27
    & 16.79 & \cellcolor{third}0.818 & \cellcolor{third}0.373
    & 25.96 & 0.836 & 0.410
    & \cellcolor{best}6.78 & CNN & Retinex theory & 0.093 & 139.99 & 30.4 ms ($1200\times700$, RTX 4090) \\

    TempRetinex \cite{Li2025TempRetinex} & 2026 & Video
    & \cellcolor{best}25.11 & \cellcolor{best}0.829 & \cellcolor{best}0.338
    & \cellcolor{best}30.27 & \cellcolor{best}0.877 & \cellcolor{best}0.327
    & \cellcolor{second}5.29
    & 19.27 & \cellcolor{best}0.849 & \cellcolor{best}0.314
    & \cellcolor{best}32.24 & \cellcolor{best}0.908 & \cellcolor{best}0.283
    & \cellcolor{third}9.65 & CNN & Retinex theory & 0.130 & 486.662 & 46.1 ms ($1200\times700$, RTX 4090) \\

    \bottomrule
  \end{tabular}}
  }
\end{table}
\end{landscape}

\subsection{Dehazing}
\label{ssec:dehazing}

We evaluate the performance of existing video dehazing methods on two representative benchmarks: the REVIDE dataset~\cite{Zhang_2021_CVPR} and the HazeWorld dataset~\cite{xuVideoDehazingMultiRange2023}. REVIDE consists of paired hazy and clean videos captured using a robotic arm under a controlled haze generation system. It includes 47 distinct scenes, with a total of 1,698 video pairs for training and 284 video pairs for testing. HazeWorld is a large-scale synthetic dataset comprising 3,588 training videos and 1,496 testing videos, covering a wide range of outdoor scenes with diverse environmental conditions.

To date, only three unsupervised methods have been proposed for video dehazing, whereas supervised approaches have been studied for a much longer period and are considerably more mature. Interestingly, the results show that VDPrior can achieve performance comparable to, or even exceeding, that of supervised methods. This approach decomposes videos into multiple layers, inspired by professional video editing workflows. Further experiments are required to better understand its limitations and generalization capabilities. Similar to denoising and low-light enhancement tasks, current unsupervised video dehazing methods are still predominantly based on CNN architectures.

Several underwater enhancement methods build upon dehazing techniques, as light attenuation in underwater environments shares similarities with haze formation in atmospheric conditions (see Section~\ref{ssec:synlightatt}), albeit with additional challenges related to wavelength-dependent absorption and color distortion~\cite{Zhou2023Review, Huang2025Underwater}. Consequently, most existing AI-based underwater enhancement approaches rely on supervised learning with paired or synthetic training data. While a limited number of unsupervised methods have been proposed, these are largely confined to image-based enhancement, including CycleGAN-based frameworks~\cite{An2024CycleGAN}, GAN models with learnable gamma correction~\cite{10423039}, and contrastive learning–based approaches~\cite{10398257}. To date, such unsupervised strategies have not been extended to video-based underwater enhancement, leaving temporal consistency and motion-aware restoration largely unexplored in this setting.

\begin{table*}[t]
\centering
\caption{Quantitative comparison of dehazing methods on the HazeWorld and REVIDE datasets. {Empty cells indicate that the corresponding metrics were not reported by the original authors under these specific evaluation protocols.}$^*$ denotes supervised methods}
\label{tab:haze_revide}
\resizebox{\textwidth}{!}{%
{%
\begin{tabular}{lcc|cc|cc|c|c|c|c|c}
\toprule
Method & Year & Type 
& \multicolumn{2}{c}{HazeWorld} & \multicolumn{2}{c}{REVIDE} & Architecture & Technique & Params (M) & GFLOPs & Runtime (Res, GPU) \\
\cmidrule(lr){4-5} \cmidrule(lr){6-7}
& & 
& PSNR $\uparrow$ & SSIM $\uparrow$ & PSNR $\uparrow$ & SSIM $\uparrow$ & & & & & \\
\midrule

DM2F$^*$ \cite{Deng2019DM2F}              & 2019 & Image & 24.54 & 0.9130 & 22.23 & 0.8651 & CNN & Supervised & -- & -- & -- (--, --) \\
MSBDN$^*$ \cite{Dong2020MSBDN}            & 2020 & Image & 23.70 & 0.8858 & 22.01 & 0.8759 & CNN & Supervised & 31.35 & 24.53 & 145 ms ($1280\times720$, TITAN RTX) \\
Dehamer$^*$ \cite{Guo2022Transformer}            & 2021 & Image & 22.92 & 0.9044 & 18.90 & 0.7827 & Transformer & Supervised & 132.4 & 48.26 & 202 ms ($1280\times720$, TITAN RTX) \\
DehazeFormer$^*$ \cite{Song2023ViT}      & 2023 & Image & 25.44 & 0.9286 & 19.70 & 0.8687 & Transformer & Supervised & 4.6 & 47.32 & 427 ms ($1280\times720$, TITAN RTX) \\
CG-IDN$^*$ \cite{Zhang_2021_CVPR}          & 2021 & Video & 25.25 & 0.9155 & 23.21 & 0.8836 & CNN & Supervised & 23 & -- & -- (--, --) \\
NCFL$^*$ \cite{Huang2022Neural}             & 2022 & Video & 24.33 & 0.9253 & 23.63 & 0.8925 & CNN & Supervised & 16 & 771 & -- (--, --) \\
BasicVSR++$^*$ \cite{chan2022basicvsr++}      & 2022 & Video & 26.06 & 0.9207 & 21.68 & 0.8726 & CNN & Supervised & 7.3 & 28.10 & 96 ms ($1280\times720$, TITAN RTX) \\
Phase$^*$ \cite{Liu2022Phase}    & 2022 & Video & --    & --     & 23.83 & 0.8950 & Fourier CNN & Supervised & -- & -- & -- (--, --) \\
MAPNet$^*$ \cite{xuVideoDehazingMultiRange2023}              & 2023 & Video & 27.12 & 0.9349 & 24.16 & 0.9043 & Attn CNN & Supervised & 28.8 & 8.21 & 101 ms ($1280\times720$, TITAN RTX) \\
Triplane$^*$ \cite{Ren2025TriplaneSmoothed}      & 2025   & Video & {28.72} & {0.9518} & {24.59} & {0.9072} & Triplane & Supervised & -- & -- & -- (--, --) \\
ASTP-net$^*$ \cite{Zhen2025Adaptive} & 2025 & Video & 29.55 & 0.9587 & 24.18 & 0.9046 & Transformer & Supervised  & -- & -- & -- (--, --) \\
\hline
IC-Dehazing \cite{10292884} & 2024 & Image & \cellcolor{red!25}24.13 & \cellcolor{red!25}0.9077 & \cellcolor{yellow!25}21.87 & \cellcolor{orange!25}0.9102  & Transformer & Retinex & -- & -- & -- (--, --) \\
DUVD \cite{DUVD:2024} & 2024 & Video & -- & -- & \cellcolor{orange!25}21.96 & \cellcolor{yellow!25}0.9090 & CNN & Depth-aware & -- & -- & -- (--, --) \\
VDPrior \cite{Shrivastava2024Prior} & 2024 & Video & -- & -- & \cellcolor{red!25}24.83 & \cellcolor{red!25}0.9164 & CNN & Layer decomp & -- & -- (--, --) \\
UnpairedVD \cite{patilUnpairedRecurrentLearning2025} & 2025 & Video & -- & -- & 19.89 & 0.8057 & Attn CNN & Recurrent & -- & -- & -- (--, --) \\

\bottomrule
\end{tabular}%
}
}
\end{table*}

\section{Challenges and Open Problems}
\label{sec:challenges}
Unsupervised video enhancement has made significant progress, yet several critical challenges remain unresolved, limiting its use in uncontrolled environments. Future research must address the following key areas:

{\noindent\textbf{Temporal Consistency and Motion modeling}}

Maintaining temporal consistency remains the primary challenge, especially in the situations with complex or rapid motion, occlusion or abrupt scene change. Without clean reference frames to guide them, unsupervised methods can get motion wrong, causing visual problems. For example, if the camera moves quickly, the method might mistakenly add details from a moving foreground object into the background after it has left the frame. This can lead to ghosting, faint traces of the object still showing, or warping, where the background looks oddly stretched or distorted. Also, unsupervised methods rely on latent structure in the input video, such as spatial redundancy or temporal coherence, to generate effective learning signals. In extremely low signal-to-noise ratio (SNR) conditions, where noise dominates the signal, these assumptions no longer hold. Temporal consistency becomes unreliable, leading to over-smoothed results or hallucinated textures. {Future research must develop robust temporal regularization techniques that address temporal consistency and real-world degradation modeling reliably even under these extreme conditions.}  

{\noindent\textbf{Cross-Domain and Cross-Camera Generalization}}
Unsupervised models are often implicitly tuned to the statistics of their training data and can degrade significantly under domain shifts. {A fundamental limitation in objective evaluation is that synthetic distortions (such as simplified additive noise, linear low-light curves, or idealized blur kernels) frequently fail to faithfully represent complex, real-world degradations. This domain gap is particularly pronounced for intricate low-light sensor noise, underwater scattering, atmospheric turbulence, and mixed, co-occurring artifacts. For example, models developed for clean synthetic settings frequently fail in real-world underwater or turbulent environments, where wavelength-dependent absorption, backscatter, and complex phase fluctuations violate standard mathematical approximations. Such mismatches lead to severe color distortion, contrast over-enhancement, and temporal instability during deployment. Establishing robust cross-domain and cross-camera generalization mechanisms is therefore required to resolve these limitations and bridge the gap between synthetic validation proxies and real-world variance.}


{\noindent\textbf{Mixed Real-World Degradations}}

Handling multiple, co-occurring degradations also remains challenging. For instance, low-light conditions are often accompanied by increased noise and reduced contrast in dark regions. This issue is particularly pronounced in user-generated content {(UGC)} videos, which are typically captured by non-professionals and therefore tend to suffer from low visual quality. Common problems include camera shake, poor illumination, and compression artifacts {\cite{Safonov_2025_CVPR}}. Most existing unsupervised enhancement methods are designed to address a single degradation type in isolation. Some attempts have been made to restore multiple degradations in images, for example, jointly handling blur and noise {\cite{NEURIPS2022_bc12914d}} or removing blur, noise, and compression artifacts via unpaired learning {\cite{Poirier_2023_CVPR}}. However, such approaches have not yet been successfully extended to video. {Developing unified blind restoration models capable of disentangling and mitigating mixed degradations without paired data remains a significant open problem.}

{\noindent\textbf{Halluination Risks and Generative Priors}}

While generative models provide powerful prior, unconstrained GANs and standard diffusion models risk injecting false semantic details into video enhancement. Investigating deterministic generative frameworks, such as flow matching, could provide more stable and controllable synthesis trajectories, reducing hallucination risks while maintaining high perceptual quality without explicit supervision.

{\noindent\textbf{Evaluation Without Paired References and Perceptual-Video Quality Metrics}}

Although unsupervised models are most often trained on real world data, there is always a risk that they can overfit their training data. Additionally, the lack of groundtruth complicates robust evaluation of unsupervised approaches. One of the alternatives to evaluation on paired data is using no-reference video quality metrics. Current no-reference metrics may be biased towards certain types of degradations, have insufficient correlation with human perception or be sensitive to adversarial attacks {\cite{pinson_2022_nrmetric}}. Having said that, there aren't many no-reference metrics specifically for video quality available. For instance, Video-BLIINDS {\cite{Saad:Blind:2012}} addresses this by leveraging a natural scene statistics {(NSS)} model of video Discrete Cosine Transform {(DCT)} coefficients alongside a temporal model of motion coherency. The current approach assesses the quality of each frame (then averages it), and the quality of temporal consistency separately {\cite{Lai_2018_ECCV}}. The well-known no-reference metrics for images are NIQE (Naturalness image quality evaluator) {\cite{mittal2013making}}, BRISQUE (Blind/Referenceless image spatial quality evaluator) {\cite{BRISQUE}}, PIQE (Perception-based image quality evaluator) {\cite{venkatanath2015blind}}, and recently the learning-based method, CLIP-IQA {\cite{wang2022exploring}}. For temporal quality, the simplest approach is to measure temporal stability by warping the previous frame to the current one and computing the mean squared error in non-occluded areas. {However, a major gap in the current literature is the inconsistent reporting of such temporal consistency metrics across different restoration tasks. While metrics like warping error are rarely reported in unsupervised deblurring, denoising, and dehazing studies, evaluations rely heavily on single-frame spatial metrics that fail to penalize temporal jitter. Developing objective metrics that accurately penalize temporal flickering and hallucinated textures is essential for benchmarking existing and future video enhancement techniques.}

{\noindent\textbf{Computational Efficiency and Real-Time Deployment Constraints}}

Finally, applicability in real-time scenarios is crucial for many practical video applications, especially in downstream machine vision tasks such as object detection, recognition, and tracking. This is particularly critical in domains like autonomous driving, where decisions must be made instantaneously, or in medical imaging, where enhanced real-time feedback can support diagnosis or intervention. In such settings, video restoration and enhancement methods must not only produce high-quality outputs but also operate with minimal latency to ensure timely and reliable responses. However, achieving this balance between visual quality and computational efficiency remains a significant challenge. Therefore, the development of lightweight, efficient, and robust unsupervised models capable of handling diverse degradation types while meeting real-time constraints is a pressing need. Future research should explore strategies such as model distillation, adaptive inference, or spatiotemporal redundancy exploitation to bridge this gap between quality and speed. {Additionally, these enhancements must explicitly address real-time deployment constraints and downstream-task-oriented enhancement, ensuring outputs improve operational reliability rather than solely perceptual quality. 
As highlighted in our evaluation tables, a significant proportion of unsupervised video enhancement literature entirely omits critical computational metrics such as parameter counts, GFLOPs, and specific runtime environments. This widespread lack of standardized reporting heavily affects the objective assessment of these models for real-world deployment. Establishing consistent efficiency benchmarks is therefore just as urgent as improving visual quality, ensuring that future unsupervised models can reliably meet the minimal latency constraints required by downstream tasks.}

\section{Conclusion}
This survey traces the progression of video restoration and enhancement techniques, starting from traditional methods based on physical models and handcrafted priors, through the rise of supervised deep learning methods that rely on paired training data, to the contemporary unsupervised, self-supervised and zero-shot strategies that offer solutions for video degradations in conditions with no access go ground truth. We synthesized recent approaches and architectural design principles in unsupervised video denoising, deblurring, dehazing, and low-light enhancement, highlighting several key directions: domain translation, self-supervised signal design, consistency-based methods, degradation-aware approaches and methods relying on  explicit or learned implicit priors. Furthermore, we proposed a multi-dimensional design space to factorize these approaches into their constituent supervision signals , temporal modeling strategies, and learning paradigms, providing a unified framework for analyzing hybrid architectures. We also looked into types of objective functions used for video enhancement, including losses for reference fidelity, data-consistency losses, prior-based losses and losses used in adversarial methods.

Finally, we addressed the role of synthetic datasets in supporting both supervised training and objective evaluation of unsupervised video enhancement approaches. While unsupervised methods do not rely ground-truth data during training stage, metrics such as PSNR, SSIM, and LPIPS still require paired references. Synthetic distortions, such as blur, noise, low-light conditions, light attenuation, and atmospheric turbulence, provide controlled settings for training and benchmarking, and thus remain essential the progress in video restoration.


\bibliographystyle{IEEEtran}
\bibliography{sn-bibliography}

@inproceedings{su2017deep,
  title={Deep video deblurring for hand-held cameras},
  author={Su, Shuochen and Delbracio, Mauricio and Wang, Jue and Sapiro, Guillermo and Heidrich, Wolfgang and Wang, Oliver},
  booktitle={Proceedings of the IEEE conference on computer vision and pattern recognition},
  pages={1279--1288},
  year={2017}
}

@inproceedings{shi2024zero,
  title={{Zero-IG}: zero-shot illumination-guided joint denoising and adaptive enhancement for low-light images},
  author={Shi, Yiqi and Liu, Duo and Zhang, Liguo and Tian, Ye and Xia, Xuezhi and Fu, Xiaojing},
  booktitle={Proceedings of the IEEE/CVF conference on computer vision and pattern recognition},
  pages={3015--3024},
  year={2024}
}

@article{foi2008practical,
  title={Practical Poissonian-Gaussian noise modeling and fitting for single-image raw-data},
  author={Foi, Alessandro and Trimeche, Mejdi and Katkovnik, Vladimir and Egiazarian, Karen},
  journal={IEEE transactions on image processing},
  volume={17},
  number={10},
  pages={1737--1754},
  year={2008},
  publisher={IEEE}
}

@inproceedings{blanksby1997noise,
  title={Noise performance of a color CMOS photogate image sensor},
  author={Blanksby, Andrew J and Loinaz, Marc J and Inglis, DA and Ackland, Bryan D},
  booktitle={International Electron Devices Meeting. IEDM Technical Digest},
  pages={205--208},
  year={1997},
  organization={IEEE}
}

@article{hill2025deep,
  author    = {Paul Hill and Nantheera Anantrasirichai and Alin Achim and David Bull},
  title     = {Deep Learning Techniques for Atmospheric Turbulence Removal: A Review},
  journal   = {Artificial Intelligence Review},
  volume    = {58},
  pages     = {101},
  year      = {2025},
  doi       = {10.1007/s10462-024-11086-6}
}

@inproceedings{ho2020denoising,
  author    = {Jonathan Ho and Ajay Jain and Pieter Abbeel},
  title     = {Denoising Diffusion Probabilistic Models},
  booktitle = {Advances in Neural Information Processing Systems (NeurIPS)},
  volume    = {33},
  pages     = {6840--6851},
  year      = {2020}
}

@inproceedings{cordts2016cityscapes,
  title={The cityscapes dataset for semantic urban scene understanding},
  author={Cordts, Marius and Omran, Mohamed and Ramos, Sebastian and Rehfeld, Timo and Enzweiler, Markus and Benenson, Rodrigo and Franke, Uwe and Roth, Stefan and Schiele, Bernt},
  booktitle={Proceedings of the IEEE conference on computer vision and pattern recognition},
  pages={3213--3223},
  year={2016}
}

@article{ghazal2008structure,
  title={Structure-oriented multidirectional wiener filter for denoising of image and video signals},
  author={Ghazal, Mohammed and Amer, Aishy and Ghrayeb, Ali},
  journal={IEEE Transactions on circuits and systems for video technology},
  volume={18},
  number={12},
  pages={1797--1802},
  year={2008},
  publisher={IEEE}
}

@article{hsia2015high,
  title={High-efficiency TV video noise reduction through adaptive spatial--temporal frame filtering},
  author={Hsia, Shih-Chang and Hsu, Wei-Chih and Tsai, Cheng-Liang},
  journal={Journal of Real-Time Image Processing},
  volume={10},
  pages={561--572},
  year={2015},
  publisher={Springer}
}

@article{dabov2007image,
  title={Image denoising by sparse 3-D transform-domain collaborative filtering},
  author={Dabov, Kostadin and Foi, Alessandro and Katkovnik, Vladimir and Egiazarian, Karen},
  journal={IEEE Transactions on image processing},
  volume={16},
  number={8},
  pages={2080--2095},
  year={2007},
  publisher={IEEE}
}

@INPROCEEDINGS{vbm3d,
  author={Dabov, Kostadin and Foi, Alessandro and Egiazarian, Karen},
  booktitle={2007 15th European Signal Processing Conference}, 
  title={Video denoising by sparse 3D transform-domain collaborative filtering}, 
  year={2007},
  volume={},
  number={},
  pages={145-149},
  doi={}}

@inproceedings{li2019enhanced,
  title={An enhanced V-BM3D algorithm for VideoSAR denoising combined with temporal information},
  author={Li, Zihan and Dong, Zhen and Yu, Anxi and He, Zhihua and Yi, Tianzhu},
  booktitle={2019 IEEE 4th International Conference on Signal and Image Processing (ICSIP)},
  pages={994--998},
  year={2019},
  organization={IEEE}
}

@article{zhang2011video,
  title={Video dehazing with spatial and temporal coherence},
  author={Zhang, Jiawan and Li, Liang and Zhang, Yi and Yang, Guoqiang and Cao, Xiaochun and Sun, Jizhou},
  journal={The Visual Computer},
  volume={27},
  pages={749--757},
  year={2011},
  publisher={Springer}
}

@article{he2010single,
  title={Single image haze removal using dark channel prior},
  author={He, Kaiming and Sun, Jian and Tang, Xiaoou},
  journal={IEEE transactions on pattern analysis and machine intelligence},
  volume={33},
  number={12},
  pages={2341--2353},
  year={2010},
  publisher={IEEE}
}

@inproceedings{adidela2021single,
  title={Single image and video Dehazing: A dark channel prior (DCP)-based approach},
  author={Adidela, Sajitha and Singh, Sakshi and Sahu, Tina and Mishra, Amrita},
  booktitle={2021 IEEE 18th India Council International Conference (INDICON)},
  pages={1--6},
  year={2021},
  organization={IEEE}
}

@article{lee2016review,
  title={A review on dark channel prior based image dehazing algorithms},
  author={Lee, Sungmin and Yun, Seokmin and Nam, Ju-Hun and Won, Chee Sun and Jung, Seung-Won},
  journal={EURASIP Journal on Image and Video Processing},
  volume={2016},
  pages={1--23},
  year={2016},
  publisher={Springer}
}

@article{park2018fast,
  title={Fast execution schemes for dark-channel-prior-based outdoor video dehazing},
  author={Park, Yongmin and Kim, Tae-Hwan},
  journal={IEEE Access},
  volume={6},
  pages={10003--10014},
  year={2018},
  publisher={IEEE}
}

@inproceedings{song2015improved,
  title={An improved image dehazing and enhancing method using dark channel prior},
  author={Song, Yingchao and Luo, Haibo and Hui, Bing and Chang, Zheng},
  booktitle={The 27th Chinese Control and Decision Conference (2015 CCDC)},
  pages={5840--5845},
  year={2015},
  organization={IEEE}
}

@article{lin2013dehazing,
  title={Dehazing for image and video using guided filter},
  author={Lin, Zheqi and Wang, Xuansheng},
  journal={Open Journal of Applied Sciences},
  volume={2},
  number={4},
  pages={123--127},
  year={2013},
  publisher={Scientific Research Publishing}
}

@INPROCEEDINGS{lv2010realdehazing,
  author={Lv, Xingyong and Chen, Wenbin and Shen, I-fan},
  booktitle={2010 18th Pacific Conference on Computer Graphics and Applications}, 
  title={Real-Time Dehazing for Image and Video}, 
  year={2010},
  volume={},
  number={},
  pages={62-69},
  doi={10.1109/PacificGraphics.2010.16}}

@article{maggioni2012video,
  title={Video denoising, deblocking, and enhancement through separable 4-D nonlocal spatiotemporal transforms},
  author={Maggioni, Matteo and Boracchi, Giacomo and Foi, Alessandro and Egiazarian, Karen},
  journal={IEEE Transactions on image processing},
  volume={21},
  number={9},
  pages={3952--3966},
  year={2012},
  publisher={IEEE}
}

@article{yahya2015videonoise,
author = {Yahya, Ali and Tan, Jieqing and Li, Lian},
year = {2015},
month = {10},
pages = {1-10},
title = {Video Noise Reduction Method Using Adaptive Spatial-Temporal Filtering},
volume = {2015},
journal = {Discrete Dynamics in Nature and Society},
doi = {10.1155/2015/351763}
}

@article{dubois2003noise,
  title={Noise reduction in image sequences using motion-compensated temporal filtering},
  author={Dubois, Eric and Sabri, Shaker},
  journal={IEEE transactions on communications},
  volume={32},
  number={7},
  pages={826--831},
  year={2003},
  publisher={IEEE}
}

@article{lee1980digital,
  title={Digital image enhancement and noise filtering by use of local statistics},
  author={Lee, Jong-Sen},
  journal={IEEE transactions on pattern analysis and machine intelligence},
  number={2},
  pages={165--168},
  year={1980},
  publisher={IEEE}
}

@article{richardson1972bayesian,
  title={Bayesian-based iterative method of image restoration},
  author={Richardson, William Hadley},
  journal={Journal of the optical society of America},
  volume={62},
  number={1},
  pages={55--59},
  year={1972},
  publisher={Optical Society of America}
}

@article{biswas2015deblurring,
  title={Deblurring images using a wiener filter},
  author={Biswas, Prodip and Sarkar, Abu Sufian and Mynuddin, Mohammed},
  journal={International Journal of Computer Applications},
  volume={109},
  number={7},
  pages={36--38},
  year={2015},
  publisher={Citeseer}
}

@article{lou2013video,
  title={Video stabilization of atmospheric turbulence distortion},
  author={Lou, Yifei and Kang, Sung Ha and Soatto, Stefano and Bertozzi, Andrea L},
  journal={Inverse Problems and Imaging},
  volume={7},
  number={3},
  pages={839--861},
  year={2013},
  publisher={Inverse Problems and Imaging}
}

@inproceedings{aubailly2009automated,
  title={Automated video enhancement from a stream of atmospherically-distorted images: the lucky-region fusion approach},
  author={Aubailly, Mathieu and Vorontsov, Mikhail A and Carhart, Gary W and Valley, Michael T},
  booktitle={Atmospheric Optics: Models, Measurements, and Target-in-the-Loop Propagation III},
  volume={7463},
  pages={104--113},
  year={2009},
  organization={SPIE}
}

@inproceedings{zhu2011stabilizing,
  title={Stabilizing and deblurring atmospheric turbulence},
  author={Zhu, Xiang and Milanfar, Peyman},
  booktitle={2011 IEEE International Conference on Computational Photography (ICCP)},
  pages={1--8},
  year={2011},
  organization={IEEE}
}

@article{jobson1997properties,
  title={Properties and performance of a center/surround retinex},
  author={Jobson, Daniel J and Rahman, Zia-ur and Woodell, Glenn A},
  journal={IEEE transactions on image processing},
  volume={6},
  number={3},
  pages={451--462},
  year={1997},
  publisher={IEEE}
}

@article{jobson1997multiscale,
  title={A multiscale retinex for bridging the gap between color images and the human observation of scenes},
  author={Jobson, Daniel J and Rahman, Zia-ur and Woodell, Glenn A},
  journal={IEEE Transactions on Image processing},
  volume={6},
  number={7},
  pages={965--976},
  year={1997},
  publisher={IEEE}
}

@article{kim1998contrast,
  title={Contrast enhancement system using spatially adaptive histogram equalization with temporal filtering},
  author={Kim, Tae Keun and Paik, Joon Ki and Kang, Bong Soon},
  journal={IEEE Transactions on Consumer Electronics},
  volume={44},
  number={1},
  pages={82--87},
  year={1998},
  publisher={IEEE}
}

@article{zuiderveld1994contrast,
  title={Contrast limited adaptive histogram equalization.},
  author={Zuiderveld, Karel J and others},
  journal={Graphics gems},
  volume={4},
  number={1},
  pages={474--485},
  year={1994},
  publisher={Academic Press, Boston, MA, USA}
}

@article{anantrasirichai2013atmospheric,
  title={Atmospheric turbulence mitigation using complex wavelet-based fusion},
  author={Anantrasirichai, Nantheera and Achim, Alin and Kingsbury, Nick G and Bull, David R},
  journal={IEEE Transactions on Image Processing},
  volume={22},
  number={6},
  pages={2398--2408},
  year={2013},
  publisher={IEEE}
}

@article{cai2009blind,
  title={Blind motion deblurring using multiple images},
  author={Cai, Jian-Feng and Ji, Hui and Liu, Chaoqiang and Shen, Zuowei},
  journal={Journal of computational physics},
  volume={228},
  number={14},
  pages={5057--5071},
  year={2009},
  publisher={Elsevier}
}

@inproceedings{wang2021seeing,
  title={Seeing dynamic scene in the dark: A high-quality video dataset with mechatronic alignment},
  author={Wang, Ruixing and Xu, Xiaogang and Fu, Chi-Wing and Lu, Jiangbo and Yu, Bei and Jia, Jiaya},
  booktitle={Proceedings of the IEEE/CVF international conference on computer vision},
  pages={9700--9709},
  year={2021}
}

@article{zhu2021detection,
  title={Detection and tracking meet drones challenge},
  author={Zhu, Pengfei and Wen, Longyin and Du, Dawei and Bian, Xiao and Fan, Heng and Hu, Qinghua and Ling, Haibin},
  journal={IEEE transactions on pattern analysis and machine intelligence},
  volume={44},
  number={11},
  pages={7380--7399},
  year={2021},
  publisher={IEEE}
}

@article{wen2020ua,
  title={UA-DETRAC: A new benchmark and protocol for multi-object detection and tracking},
  author={Wen, Longyin and Du, Dawei and Cai, Zhaowei and Lei, Zhen and Chang, Ming-Ching and Qi, Honggang and Lim, Jongwoo and Yang, Ming-Hsuan and Lyu, Siwei},
  journal={Computer Vision and Image Understanding},
  volume={193},
  pages={102907},
  year={2020},
  publisher={Elsevier}
}

@InProceedings{Chen_2025_ICCV,
    author    = {Chen, Zhikai and Long, Fuchen and Qiu, Zhaofan and Yao, Ting and Zhou, Wengang and Luo, Jiebo and Mei, Tao},
    title     = {Aligning Global Semantics and Local Textures in Generative Video Enhancement},
    booktitle = {Proceedings of the IEEE/CVF International Conference on Computer Vision (ICCV)},
    month     = {October},
    year      = {2025},
    pages     = {17087-17096}
}

@inproceedings{guizilini20203d,
  title={3d packing for self-supervised monocular depth estimation},
  author={Guizilini, Vitor and Ambrus, Rares and Pillai, Sudeep and Raventos, Allan and Gaidon, Adrien},
  booktitle={Proceedings of the IEEE/CVF conference on computer vision and pattern recognition},
  pages={2485--2494},
  year={2020}
}

@InProceedings{Wang_2025_CVPR,
    author    = {Wang, Ruiyi and Zheng, Yushuo and Zhang, Zicheng and Li, Chunyi and Liu, Shuaicheng and Zhai, Guangtao and Liu, Xiaohong},
    title     = {Learning Hazing to Dehazing: Towards Realistic Haze Generation for Real-World Image Dehazing},
    booktitle = {Proceedings of the Computer Vision and Pattern Recognition Conference (CVPR)},
    month     = {June},
    year      = {2025},
    pages     = {23091-23100}
}

@article{Anantrasirichai:AI:2022,
  title={Artificial intelligence in the creative industries: a review},
  author={Anantrasirichai, N. and Bull, D.},
  journal={Artificial Intelligence Review},
  volume={55},
  pages={589--656},
  year={2022},
  publisher={Springer},
  doi={10.1007/s10462-021-10039-7}
}

@article{Yao:video:2020,
author = {Yao, Rui and Lin, Guosheng and Xia, Shixiong and Zhao, Jiaqi and Zhou, Yong},
title = {Video Object Segmentation and Tracking: A Survey},
year = {2020},
issue_date = {August 2020},
publisher = {Association for Computing Machinery},
address = {New York, NY, USA},
volume = {11},
number = {4},
doi = {10.1145/3391743},
journal = {ACM Trans. Intell. Syst. Technol.}
}

@article{Yi:Comprehensive:2024,
  title={A Comprehensive Study of Object Tracking in Low-Light Environments},
  author={Yi, A and Anantrasirichai, N},
  journal={Sensors},
   volume={24},
  number={14},
  year={2024},
  url={https://arxiv.org/pdf/2312.16250.pdf}
}

@incollection{AJAFERNANDEZ2024145,
title = {Chapter 6 - Image filtering: enhancement and restoration},
editor = {Alejandro F. Frangi and Jerry L. Prince and Milan Sonka},
booktitle = {Medical Image Analysis},
publisher = {Academic Press},
pages = {145-176},
year = {2024},
series = {The MICCAI Society book Series},
isbn = {978-0-12-813657-7},
doi = {https://doi.org/10.1016/B978-0-12-813657-7.00019-4},
author = {Santiago Aja-Fernández and Ariel H. Curiale and Jerry L. Prince}
}

@article{du2024end,
  author    = {Dazhao Du and Enhan Li and Lingyu Si and Fanjiang Xu and Jianwei Niu},
  title     = {End-To-End Underwater Video Enhancement: Dataset and Model},
  journal   = {arXiv preprint arXiv:2403.11506},
  year      = {2024},
  url       = {https://arxiv.org/abs/2403.11506}
}

@inproceedings{yu2020image,
  author    = {Guoshen Yu and Guillermo Sapiro},
  title     = {Image Enhancement and Restoration: Traditional Approaches},
  booktitle = {Computer Vision},
  year      = {2020},
  doi       = {\doi{10.1007/978-3-030-03243-2\_233-1}},
  url       = {\url{https://doi.org/10.1007/978-3-030-03243-2\_233-1}}
  }

@book{celebi2015color,
  editor    = {M.~Emre Celebi and Michela Lecca and Bogdan Smolka},
  title     = {Color Image and Video Enhancement},
  publisher = {Springer International Publishing, Cham},
  year      = {2015},
  doi       = {10.1007/978-3-319-09363-5},
  isbn      = {978-3-319-09363-5}
}

@inproceedings{hill2025automatic,
  author    = {Paul Hill and Adrian Achim and David Bull and Nantheera Anantrasirichai},
  title     = {Automatic Object Detection in Atmospheric Turbulence-Affected Environments},
  booktitle = {Proceedings of SPIE -- Automatic Target Recognition XXXV},
  volume    = {13463},
  pages     = {145--147},
  year      = {2025},
  doi       = {10.1117/12.3053981}
}

@ARTICLE{Li:low:2022,
  author={Li, Chongyi and Guo, Chunle and Han, Linghao and Jiang, Jun and Cheng, Ming-Ming and Gu, Jinwei and Loy, Chen Change},
  journal={IEEE Transactions on Pattern Analysis and Machine Intelligence}, 
  title={Low-Light Image and Video Enhancement Using Deep Learning: A Survey}, 
  year={2022},
  volume={44},
  number={12},
  pages={9396-9416},
  doi={10.1109/TPAMI.2021.3126387}}

@INPROCEEDINGS{Morris:dabit:2025,
  author={Morris, Crispian and Anantrasirichai, Nantheera and Zhang, Fan and Bull, David},
  booktitle={2025 IEEE/CVF Winter Conference on Applications of Computer Vision Workshops (WACVW)}, 
  title={{DaBiT: Depth} and Blur Informed Transformer for Video Focal Deblurring}, 
  year={2025},
  volume={},
  number={},
  pages={278-288},
  doi={10.1109/WACVW65960.2025.00038}}

@article{zhang2026unleashing,
  title={Unleashing the power of self-supervised image denoising: A comprehensive review},
  author={Zhang, Dan and Zhou, Fangfang and Jiang, Yuwen and Albu, Felix and Wei, Yuanzhou and Yang, Xiao and Li, Qiang},
  journal={Computer Vision and Image Understanding},
  pages={104786},
  year={2026},
  publisher={Elsevier}
}

@INPROCEEDINGS{Yang:DepthAnything:2024,
  author={Yang, Lihe and Kang, Bingyi and Huang, Zilong and Xu, Xiaogang and Feng, Jiashi and Zhao, Hengshuang},
  booktitle={2024 IEEE/CVF Conference on Computer Vision and Pattern Recognition (CVPR)}, 
  title={{Depth Anything: Unleashing} the Power of Large-Scale Unlabeled Data}, 
  year={2024},
  volume={},
  number={},
  pages={10371-10381},
  doi={10.1109/CVPR52733.2024.00987}}

@InProceedings{Zhang:learning:2021,
  author={Zhang, Fan and Li, Yu and You, Shaodi and Fu, Ying},
  booktitle={2021 IEEE/CVF Conference on Computer Vision and Pattern Recognition (CVPR)}, 
  title={Learning Temporal Consistency for Low Light Video Enhancement from Single Images}, 
  year={2021},
  volume={},
  number={},
  pages={4965-4974},
  doi={10.1109/CVPR46437.2021.00493}}

@article{xiang2025deep,
  author    = {Yizhou Xiang and Huan Zhou and Chenyang Li and others},
  title     = {Deep Learning in Motion Deblurring: Current Status, Benchmarks and Future Prospects},
  journal   = {The Visual Computer},
  volume    = {41},
  pages     = {3801--3827},
  year      = {2025},
  doi       = {10.1007/s00371-024-03632-8},
  url       = {https://doi.org/10.1007/s00371-024-03632-8}
}

@book{richards2022remote,
  author    = {John A. Richards},
  title     = {Remote Sensing Digital Image Analysis},
  publisher = {Springer, Cham},
  year      = {2022},
  doi       = {10.1007/978-3-030-82327-6},
  url       = {https://doi.org/10.1007/978-3-030-82327-6}
}

@article{hu2022overview,
  author    = {Kai Hu and Chuang Weng and Yuhuan Zhang and Jiawei Jin and Qun Xia},
  title     = {An Overview of Underwater Vision Enhancement: From Traditional Methods to Recent Deep Learning},
  journal   = {Journal of Marine Science and Engineering},
  volume    = {10},
  number    = {2},
  pages     = {241},
  year      = {2022},
  doi       = {10.3390/jmse10020241},
  url       = {https://doi.org/10.3390/jmse10020241}
}

@article{rota2023video,
  author    = {Cristian Rota and Matteo Buzzelli and Simone Bianco and Raimondo Schettini},
  title     = {Video Restoration Based on Deep Learning: A Comprehensive Survey},
  journal   = {Artificial Intelligence Review},
  volume    = {56},
  pages     = {5317--5364},
  year      = {2023},
  doi       = {10.1007/s10462-022-10302-5},
  url       = {https://doi.org/10.1007/s10462-022-10302-5}
}

@InProceedings{Wang_2019_CVPR,
author = {Wang, Jiangliu and Jiao, Jianbo and Bao, Linchao and He, Shengfeng and Liu, Yunhui and Liu, Wei},
title = {Self-Supervised Spatio-Temporal Representation Learning for Videos by Predicting Motion and Appearance Statistics},
booktitle = {Proceedings of the IEEE/CVF Conference on Computer Vision and Pattern Recognition (CVPR)},
month = {June},
year = {2019}
}

@article{liu2025appearance,
  author    = {Rong Liu and Yizhou Zhu and Guobao B. Luo},
  title     = {Appearance Consistency and Motion Coherence Learning for Internal Video Inpainting},
  journal   = {CAAI Transactions on Intelligence Technology},
  pages     = {1--15},
  year      = {2025},
  doi       = {10.1049/cit2.12405},
  url       = {https://doi.org/10.1049/cit2.12405}
}

@article{GOYAL2024102151,
title = {Recent advances in image dehazing: Formal analysis to automated approaches},
journal = {Information Fusion},
volume = {104},
pages = {102151},
year = {2024},
issn = {1566-2535},
doi = {https://doi.org/10.1016/j.inffus.2023.102151},
author = {Bhawna Goyal and Ayush Dogra and Dawa Chyophel Lepcha and Vishal Goyal and Ahmed Alkhayyat and Jasgurpreet Singh Chohan and Vinay Kukreja}
}

@article{anantrasirichai2025artificial,
  title={Advances in artificial intelligence: a review for the creative industries},
  author={Anantrasirichai, Nantheera and Zhang, Fan and Bull, David},
  journal={Artificial Intelligence Review},
  volume={59},
  pages={89},
  year={2026}
}

@inproceedings{nayar1999vision,
  author    = {S. K. Nayar and S. G. Narasimhan},
  title     = {Vision in Bad Weather},
  booktitle = {Proceedings of the IEEE International Conference on Computer Vision (ICCV)},
  volume    = {2},
  pages     = {820--827},
  year      = {1999},
  doi       = {10.1109/ICCV.1999.790306}
}

@misc{bvilowlight2023,
doi = {10.21227/mzny-8c77},
url = {https://dx.doi.org/10.21227/mzny-8c77},
author = {Nantheera Anantrasirichai and Alexandra Malyugina and Rachel Lin and David Bull},
publisher = {IEEE Dataport},
title = {BVI-Lowlight: Fully registered datasets for low-light image and video enhancement},
year = {2023} }

@inproceedings{malyugina2025marine,
  title={{Marine Snow Removal Using Internally Generated Pseudo Ground Truth}},
  author={Malyugina, A and Huang, G and Ruiz, E and Leslie, B and Anantrasirichai, N},
  booktitle={33rd European Signal Processing Conference (EUSIPCO)},
  year={2025}
}

@article{malyugina2024topological,
  title     = {A topological loss function for image denoising on a new {BVI}-lowlight dataset},
  author    = {Malyugina, A. and Anantrasirichai, N. and Bull, D.},
  journal   = {Signal Processing},
  volume    = {211},
  pages     = {109081},
  year      = {2024},
  publisher = {Elsevier},
  doi       = {10.1016/j.sigpro.2024.109081}
}

@inproceedings{cao2025zeroshot,
  author    = {Cheng Cao and Heng Yue and Xiaoyu Liu and Jian Yang},
  title     = {Zero-shot Video Restoration and Enhancement Using Pre-Trained Image Diffusion Model},
  booktitle = {Proceedings of the AAAI Conference on Artificial Intelligence},
  volume    = {39},
  number    = {2},
  pages     = {1935--1943},
  year      = {2025},
  doi       = {10.1609/aaai.v39i2.32189},
  url       = {https://doi.org/10.1609/aaai.v39i2.32189}
}

@InProceedings{Huang_2021_CVPR,
    author    = {Huang, Tao and Li, Songjiang and Jia, Xu and Lu, Huchuan and Liu, Jianzhuang},
    title     = {Neighbor2Neighbor: Self-Supervised Denoising From Single Noisy Images},
    booktitle = {Proceedings of the IEEE/CVF Conference on Computer Vision and Pattern Recognition (CVPR)},
    month     = {June},
    year      = {2021},
    pages     = {14781-14790}
}

@inproceedings{ronneberger2015u,
  title={U-net: Convolutional networks for biomedical image segmentation},
  author={Ronneberger, Olaf and Fischer, Philipp and Brox, Thomas},
  booktitle={Medical image computing and computer-assisted intervention--MICCAI 2015: 18th international conference, Munich, Germany, October 5-9, 2015, proceedings, part III 18},
  pages={234--241},
  year={2015},
  organization={Springer}
}

@inproceedings{yue2020supervised,
  title={Supervised raw video denoising with a benchmark dataset on dynamic scenes},
  author={Yue, Huanjing and Cao, Cong and Liao, Lei and Chu, Ronghe and Yang, Jingyu},
  booktitle={Proceedings of the IEEE/CVF conference on computer vision and pattern recognition},
  pages={2301--2310},
  year={2020}
}

@inproceedings{chan2021basicvsr,
  title={Basicvsr: The search for essential components in video super-resolution and beyond},
  author={Chan, Kelvin CK and Wang, Xintao and Yu, Ke and Dong, Chao and Loy, Chen Change},
  booktitle={Proceedings of the IEEE/CVF conference on computer vision and pattern recognition},
  pages={4947--4956},
  year={2021}
}

@inproceedings{chan2022basicvsr++,
  title={Basicvsr++: Improving video super-resolution with enhanced propagation and alignment},
  author={Chan, Kelvin CK and Zhou, Shangchen and Xu, Xiangyu and Loy, Chen Change},
  booktitle={Proceedings of the IEEE/CVF conference on computer vision and pattern recognition},
  pages={5970--5979},
  year={2022}
}

@inproceedings{yang2019frame,
  title={Frame-consistent recurrent video deraining with dual-level flow},
  author={Yang, Wenhan and Liu, Jiaying and Feng, Jiashi},
  booktitle={Proceedings of the IEEE/CVF conference on computer vision and pattern recognition},
  pages={1661--1670},
  year={2019}
}

@article{ayoub2024dehazingreview,
  title={Review of dehazing techniques: Challenges and future trends},
  author={Ayoub, Abeer and El-Shafai, Walid and El-Samie, Fathi E Abd and Hamad, Ehab KI and EL-Rabaie, El-Sayed M},
  journal={Multimedia Tools and Applications},
  pages={1--29},
  year={2024},
  publisher={Springer}
}

@article{zheng2022lowlightreview,
  title={Low-light image and video enhancement: A comprehensive survey and beyond},
  author={Zheng, Shen and Ma, Yiling and Pan, Jinqian and Lu, Changjie and Gupta, Gaurav},
  journal={arXiv preprint arXiv:2212.10772},
  year={2022}
}

@article{ye2024lowlightsurvey,
  title={A survey on learning-based low-light image and video enhancement},
  author={Ye, Jing and Qiu, Changzhen and Zhang, Zhiyong},
  journal={Displays},
  volume={81},
  pages={102614},
  year={2024},
  publisher={Elsevier}
}

@article{lhiadi2025deepehnhreview,
author = {Lhiadi, Redouane and Jaddar, Abdessamad and Kaaouachi, Abdelali},
year = {2025},
month = {04},
pages = {1518},
title = {Deep learning-based techniques for video enhancement, compression and restoration},
volume = {14},
journal = {IAES International Journal of Artificial Intelligence (IJ-AI)},
doi = {10.11591/ijai.v14.i2.pp1518-1530}
}

@article{wang2022derainingsurvey,
  title={Survey on rain removal from videos or a single image},
  author={Wang, Hong and Wu, Yichen and Li, Minghan and Zhao, Qian and Meng, Deyu},
  journal={Science China Information Sciences},
  volume={65},
  number={1},
  pages={111101},
  year={2022},
  publisher={Springer}
}

@inproceedings{sheeba2019denoisingreview,
  title={A review on video denoising methods},
  author={Sheeba, MC and Christopher, C Seldev},
  booktitle={2019 International Conference on Recent Advances in Energy-efficient Computing and Communication (ICRAECC)},
  pages={1--6},
  year={2019},
  organization={IEEE}
}

@InProceedings{Nah_2017_CVPR,
  author = {Nah, Seungjun and Kim, Tae Hyun and Lee, Kyoung Mu},
  title = {Deep Multi-Scale Convolutional Neural Network for Dynamic Scene Deblurring},
  booktitle = {CVPR},
  month = {July},
  year = {2017}
}

@ARTICLE{pinson_2022_nrmetric,
  author={Pinson, Margaret H.},
  journal={IEEE Transactions on Broadcasting}, 
  title={Why No Reference Metrics for Image and Video Quality Lack Accuracy and Reproducibility}, 
  year={2023},
  volume={69},
  number={1},
  pages={97-117},
  doi={10.1109/TBC.2022.3191059}}

@InProceedings{li2025zerotig,
      title={{Zero-TIG: Temporal} Consistency-Aware Zero-Shot Illumination-Guided Low-light Video Enhancement}, 
      author={Yini Li and Nantheera Anantrasirichai},
      year={2025},
      booktitle={European Signal Processing Conference} 
}

@INPROCEEDINGS{Srinath:undive:2025,
  author={Srinath, Suhas and Chandrasekar, Aditya and Jamadagni, Hemang and Soundararajan, Rajiv and A P, Prathosh},
  booktitle={2025 IEEE/CVF Winter Conference on Applications of Computer Vision (WACV)}, 
  title={{UnDIVE: Generalized} Underwater Video Enhancement Using Generative Priors}, 
  year={2025},
  volume={},
  number={},
  pages={9001-9012},
  doi={10.1109/WACV61041.2025.00872}}

@ARTICLE{LiUnderwater:2020,
  author={Li, Chongyi and Guo, Chunle and Ren, Wenqi and Cong, Runmin and Hou, Junhui and Kwong, Sam and Tao, Dacheng},
  journal={IEEE Transactions on Image Processing}, 
  title={An Underwater Image Enhancement Benchmark Dataset and Beyond}, 
  year={2020},
  volume={29},
  number={},
  pages={4376-4389},
  doi={10.1109/TIP.2019.2955241}}

@InProceedings{Zhang_2021_CVPR,
    author    = {Zhang, Xinyi and Dong, Hang and Pan, Jinshan and Zhu, Chao and Tai, Ying and Wang, Chengjie and Li, Jilin and Huang, Feiyue and Wang, Fei},
    title     = {Learning To Restore Hazy Video: A New Real-World Dataset and a New Method},
    booktitle = {Proceedings of the IEEE/CVF Conference on Computer Vision and Pattern Recognition (CVPR)},
    month     = {June},
    year      = {2021},
    pages     = {9239-9248}
}

@InProceedings{Ye_2022_CVPR,
    author    = {Ye, Tian and Chen, Sixiang and Liu, Yun and Ye, Yi and Chen, Erkang and Li, Yuche},
    title     = {Underwater Light Field Retention: Neural Rendering for Underwater Imaging},
    booktitle = {Proceedings of the IEEE/CVF Conference on Computer Vision and Pattern Recognition (CVPR) Workshops},
    month     = {June},
    year      = {2022},
    pages     = {488-497}
}

@INPROCEEDINGS{Zhang:HazeRD:2027,
  author={Zhang, Yanfu and Ding, Li and Sharma, Gaurav},
  booktitle={2017 IEEE International Conference on Image Processing (ICIP)}, 
  title={{HazeRD: An} outdoor scene dataset and benchmark for single image dehazing}, 
  year={2017},
  volume={},
  number={},
  pages={3205-3209},
  doi={10.1109/ICIP.2017.8296874}}

@INPROCEEDINGS{Contextual:2021,
  author={Anantrasirichai, N. and Bull, David},
  booktitle={2021 IEEE International Conference on Image Processing (ICIP)}, 
  title={Contextual Colorization and Denoising for Low-Light Ultra High Resolution Sequences}, 
  year={2021},
  volume={},
  number={},
  pages={1614-1618},
  doi={10.1109/ICIP42928.2021.9506694}}

@article{li2023diffusion,
  title={Diffusion Models for Image Restoration and Enhancement: A Comprehensive Survey},
  author={Xin Li and Yulin Ren and Xin Jin and Cuiling Lan and Xingrui Wang and Wenjun Zeng and Xinchao Wang and Zhibo Chen},
  journal={International Journal of Computer Vision},
  volume={133},
  pages={8078--8108},
  year={2025},
  doi={10.1007/s11263-025-02570-9}
}

@INPROCEEDINGS{9286636,
  author={Naoto Chiche, Benjamin and Frontera-Pons, Joana and Woiselle, Arnaud and Starck, Jean-Luc},
  booktitle={Tenth International Conference on Image Processing Theory, Tools and Applications (IPTA)}, 
  title={Deep Unrolled Network for Video Super-Resolution}, 
  year={2020},
  volume={},
  number={},
  pages={1-6},
  doi={10.1109/IPTA50016.2020.9286636}}

@ARTICLE{392335,
  author={Geman, D. and Chengda Yang},
  journal={IEEE Transactions on Image Processing}, 
  title={Nonlinear image recovery with half-quadratic regularization}, 
  year={1995},
  volume={4},
  number={7},
  pages={932-946},
  doi={10.1109/83.392335}}

@ARTICLE{10479079,
  author={De Weerdt, Brent and Eldar, Yonina C. and Deligiannis, Nikos},
  journal={IEEE Transactions on Signal Processing}, 
  title={Deep Unfolding Transformers for Sparse Recovery of Video}, 
  year={2024},
  volume={72},
  number={},
  pages={1782-1796},
  doi={10.1109/TSP.2024.3381749}}

@inproceedings{Wen:unpaired:2024,
author = {Wen, Yuanbo and Gao, Tao and Chen, Ting},
title = {Unpaired Photo-realistic Image Deraining with Energy-informed Diffusion Model},
year = {2024},
doi = {10.1145/3664647.3680560},
booktitle = {Proceedings of the 32nd ACM International Conference on Multimedia},
pages = {360–369},
numpages = {10},
location = {Melbourne VIC, Australia},
series = {MM '24}
}

@inproceedings{radford2021learning,
  title={Learning transferable visual models from natural language supervision},
  author={Radford, Alec and Kim, Jong Wook and Hallacy, Christine and Ramesh, Aditya and Goh, Gabriel and Agarwal, Sandhini and Sastry, Saurabh and Askell, Amanda and Mishkin, Pamela and Clark, Jack and Krueger, Gretchen and Sutskever, Ilya},
  booktitle={Proceedings of the 38th International Conference on Machine Learning (ICML)},
  year={2021},
  volume={139},
  pages={8748--8763},
  organization={PMLR}
}

@article{land1971lightness,
  author = {Edwin H. Land and John J. McCann},
  title = {Lightness and Retinex Theory},
  journal = {Journal of the Optical Society of America},
  volume = {61},
  number = {1},
  pages = {1-11},
  year = {1971},
  doi = {10.1364/JOSA.61.000001}
}

@inproceedings{mcglamery1980computer,
  author    = {B. L. McGlamery},
  title     = {A Computer Model for Underwater Camera Systems},
  booktitle = {Ocean Optics VI},
  year      = {1980}
}

@article{jaffe1990computer,
  title={Computer modeling and the design of optimal underwater imaging systems},
  author={Jaffe, Jules S},
  journal={IEEE Journal of Oceanic Engineering},
  volume={15},
  number={2},
  pages={101--111},
  year={1990},
  publisher={IEEE}
}

@article{hardie2017simulation,
  author    = {Russell C. Hardie and John D. Power and David A. LeMaster and Darrell R. D. SGS Bose-Pillai},
  title     = {Simulation of Anisoplanatic Imaging Through Optical Turbulence Using Numerical Wave Propagation with New Validation Analysis},
  journal   = {Optical Engineering},
  volume    = {56},
  number    = {7},
  pages     = {071502},
  year      = {2017},
  doi       = {10.1117/1.OE.56.7.071502},
  url       = {https://doi.org/10.1117/1.OE.56.7.071502}
}

@article{Li:WaterGAN:2017,
  title = {{{WaterGAN}}: {{Unsupervised Generative Network}} to {{Enable Real-time Color Correction}} of {{Monocular Underwater Images}}},
  shorttitle = {{{WaterGAN}}},
  author = {Li, Jie and Skinner, Katherine A. and Eustice, Ryan M. and {Johnson-Roberson}, Matthew},
  year = {2017},
  journal = {IEEE Robotics and Automation Letters},
  eprint = {1702.07392},
  primaryclass = {cs},
  pages = {1--1},
  issn = {2377-3766, 2377-3774},
  doi = {10.1109/LRA.2017.2730363},
  urldate = {2025-02-26},
  archiveprefix = {arXiv}
}

@article{chimitt2020simulating,
  author    = {Nicholas Chimitt and Stanley H. Chan},
  title     = {Simulating Anisoplanatic Turbulence by Sampling Intermodal and Spatially Correlated Zernike Coefficients},
  journal   = {Optical Engineering},
  volume    = {59},
  number    = {8},
  pages     = {083101},
  year      = {2020},
  doi       = {10.1117/1.OE.59.8.083101},
  url       = {https://doi.org/10.1117/1.OE.59.8.083101}
}

@inproceedings{ehret2019model,
  title={Model-blind video denoising via frame-to-frame training},
  author={Ehret, Thibaud and Davy, Axel and Morel, Jean-Michel and Facciolo, Gabriele and Arias, Pablo},
  booktitle={Proceedings of the IEEE/CVF conference on computer vision and pattern recognition},
  pages={11369--11378},
  year={2019}
}

@INPROCEEDINGS{9711075,
  author={Mao, Zhiyuan and Chimitt, Nicholas and Chan, Stanley H.},
  booktitle={2021 IEEE/CVF International Conference on Computer Vision (ICCV)}, 
  title={Accelerating Atmospheric Turbulence Simulation via Learned Phase-to-Space Transform}, 
  year={2021},
  volume={},
  number={},
  pages={14739-14748},
  doi={10.1109/ICCV48922.2021.01449}}

@InProceedings{Mao_2022_ECCV,
author = {Zhiyuan Mao and Ajay Jaiswal and Zhangyang Wang and Stanley H. Chan},
title = {Single Frame Atmospheric Turbulence Mitigation: A Benchmark Study and A New Physics-Inspired Transformer Model},
booktitle = {European Conference on Computer Vision},
month = {October},
year = {2022}
}

@inproceedings{hirsch2010efficient,
  author    = {Michael Hirsch and S. Sra and Bernhard Sch{\"o}lkopf and Stefan Harmeling},
  title     = {Efficient Filter Flow for Space-Variant Multiframe Blind Deconvolution},
  booktitle = {Proceedings of the IEEE Conference on Computer Vision and Pattern Recognition (CVPR)},
  pages     = {607--614},
  year      = {2010},
  publisher = {IEEE}
}

@article{chak2021subsampled,
  author    = {Wai Ho Chak and Choi Ping Lau and Lok Ming Lui},
  title     = {Subsampled Turbulence Removal Network},
  journal   = {Mathematics and Computation in Geometry and Data},
  volume    = {1},
  number    = {1},
  pages     = {1--33},
  year      = {2021}
}

@ARTICLE{10023498,
  author={Mei, Kangfu and Patel, Vishal M.},
  journal={IEEE Journal of Selected Topics in Signal Processing}, 
  title={LTT-GAN: Looking Through Turbulence by Inverting GANs}, 
  year={2023},
  volume={17},
  number={3},
  pages={587-598},
  doi={10.1109/JSTSP.2023.3238552}}

@article{kim2013optimized,
  title={Optimized contrast enhancement for real-time image and video dehazing},
  author={Kim, Jin-Hwan and Jang, Won-Dong and Sim, Jae-Young and Kim, Chang-Su},
  journal={Journal of Visual Communication and Image Representation},
  volume={24},
  number={3},
  pages={410--425},
  year={2013},
  publisher={Elsevier}
}

@inproceedings{ren2017video,
  title={Video desnowing and deraining based on matrix decomposition},
  author={Ren, Weihong and Tian, Jiandong and Han, Zhi and Chan, Antoni and Tang, Yandong},
  booktitle={Proceedings of the IEEE conference on computer vision and pattern recognition},
  pages={4210--4219},
  year={2017}
}

@inproceedings{chenReblur2DeblurDeblurringVideos2018,
  title = {{{Reblur2Deblur}}: {{Deblurring}} Videos via Self-Supervised Learning},
  shorttitle = {{{Reblur2Deblur}}},
  booktitle = {2018 {{IEEE International Conference}} on {{Computational Photography}} ({{ICCP}})},
  author = {Chen, Huaijin and Gu, Jinwei and Gallo, Orazio and Liu, Ming-Yu and Veeraraghavan, Ashok and Kautz, Jan},
  date = {2018-05},
  year={{2018}},
  pages = {1--9},
  issn = {2472-7636},
  doi = {10.1109/ICCPHOT.2018.8368468},
  url = {https://ieeexplore.ieee.org/document/8368468/},
  urldate = {  2025-05-15},
  eventtitle = {2018 {{IEEE International Conference}} on {{Computational Photography}} ({{ICCP}})}
}

@inproceedings{dewilSelfsupervisedTrainingBlind2021,
  title = {Self-Supervised Training for Blind Multi-Frame Video Denoising},
  booktitle = {2021 {{IEEE Winter Conference}} on {{Applications}} of {{Computer Vision}} ({{WACV}})},
  author = {Dewil, Valery and Anger, Jeremy and Davy, Axel and Ehret, Thibaud and Facciolo, Gabriele and Arias, Pablo},
  date = {2021-01},
  year = {{2021}},
  pages = {2723--2733},
  publisher = {IEEE},
  location = {Waikoloa, HI, USA},
  doi = {10.1109/WACV48630.2021.00277},
  url = {https://ieeexplore.ieee.org/document/9423377/},
  urldate = {   2025-05-19},
  eventtitle = {2021 {{IEEE Winter Conference}} on {{Applications}} of {{Computer Vision}} ({{WACV}})},
  isbn = {978-1-6654-0477-8},
  langid = {english}
}

@article{fuoliEfficientRecurrentAdversarial2023,
  title = {An {{Efficient Recurrent Adversarial Framework}} for {{Unsupervised Real-Time Video Enhancement}}},
  author = {Fuoli, Dario and Huang, Zhiwu and Paudel, Danda Pani and Van Gool, Luc and Timofte, Radu},
  date = {2023-04},
  year = {{2023}},
  journal = {{International Journal of Computer Vision}},
  shortjournal = {Int J Comput Vis},
  volume = {131},
  number = {4},
  pages = {1042--1059},
  issn = {0920-5691, 1573-1405},
  doi = {10.1007/s11263-022-01735-0},
  url = {https://link.springer.com/10.1007/s11263-022-01735-0},
  urldate = { 2025-05-26},
  langid = {english}
}

@inproceedings{fuTemporalPluginUnsupervised2024,
  title = {Temporal {{As}} a~{{Plugin}}: {{Unsupervised Video Denoising}} with~{{Pre-trained Image Denoisers}}},
  shorttitle = {Temporal {{As}} a~{{Plugin}}},
  booktitle = {Computer {{Vision}} – {{ECCV}} 2024: 18th {{European Conference}}, {{Milan}}, {{Italy}}, {{September}} 29–{{October}} 4, 2024, {{Proceedings}}, {{Part LVI}}},
  author = {Fu, Zixuan and Guo, Lanqing and Wang, Chong and Wang, Yufei and Li, Zhihao and Wen, Bihan},
  date = {2024-10-30},
  year={{2024}},
  pages = {349--367},
  publisher = {Springer-Verlag},
  location = {Berlin, Heidelberg},
  doi = {10.1007/978-3-031-72992-8\_20},
  url = {https://doi.org/10.1007/978-3-031-72992-8\_20},
  urldate = {  2025-05-21},
  isbn = {978-3-031-72991-1}
}

@inproceedings{leeRestoreRestoredVideo2021,
  title = {Restore from {{Restored}}: {{Video Restoration}} with {{Pseudo Clean Video}}},
  shorttitle = {Restore from {{Restored}}},
  booktitle = {2021  IEEE/CVF Conference on Computer Vision and  Pattern Recognition (CVPR)},
  author = {Lee, Seunghwan and Cho, Donghyeon and Kim, Jiwon and Kim, Tae Hyun},
  date = {2021-06},
  year = {{2021}},
  pages = {3536--3545},
  publisher = {IEEE},
  location = {Nashville, TN, USA},
  doi = {10.1109/CVPR46437.2021.00354},
  url = {https://ieeexplore.ieee.org/document/9578402/},
  urldate = {   2025-05-19},
  eventtitle = {2021  IEEE/CVF Conference on Computer Vision and  Pattern Recognition (CVPR)},
  isbn = {978-1-6654-4509-2},
  langid = {english}
}

@inproceedings{linUnsupervisedFlowAlignedSequencetoSequence,
  title = {Unsupervised {{Flow-Aligned Sequence-to-Sequence Learning}} for {{Video Restoration}}},
  author={Lin, Jing  and Hu, Xiaowan and Cai, Yuanhao and Wang, Haoqian and Yan, Youliang and Zou, Xueyi and Zhang, Yulun and Van Gool, Luc},
  booktitle={International Conference on Machine Learning},
  year={2022}
}

@ARTICLE{10428075,
  author={Bai, Haoran and Pan, Jinshan},
  journal={IEEE Transactions on Pattern Analysis and Machine Intelligence}, 
  title={Self-Supervised Deep Blind Video Super-Resolution}, 
  year={2024},
  volume={46},
  number={7},
  pages={4641-4653},
  doi={10.1109/TPAMI.2024.3361168}}

@inproceedings{Wang2019EDVR,
  author    = {Wang, Xintao and Yu, Kun and Wu, Shixiang and Gu, Chao and Liu, Yi and Dong, Yue and Qiao, Yu and Loy, Chen Change},
  title     = {{EDVR}: Video Restoration With Enhanced Deformable Convolutional Networks},
  booktitle = {Proceedings of the IEEE/CVF Conference on Computer Vision and Pattern Recognition Workshops (CVPRW)},
  year      = {2019},
  pages     = {0--0},
  organization = {IEEE}
}

@article{lvUnsupervisedLowLightVideo2023,
  title = {Unsupervised {{Low-Light Video Enhancement With Spatial-Temporal Co-Attention Transformer}}},
  author = {Lv, Xiaoqian and Zhang, Shengping and Wang, Chenyang and Zhang, Weigang and Yao, Hongxun and Huang, Qingming},
  year = {{2023}},
  journal = {IEEE Transactions on Image Processing},
  shortjournal = {IEEE Trans. on Image Process.},
  volume = {32},
  pages = {4701--4715},
  issn = {1057-7149, 1941-0042},
  doi = {10.1109/TIP.2023.3301332},
  url = {https://ieeexplore.ieee.org/document/10210621/},
  urldate = {2025-05-26},
  langid = {english}
}

@inproceedings{guo2020zero,
  title={Zero-reference deep curve estimation for low-light image enhancement},
  author={Guo, Chunle and Li, Chongyi and Guo, Jichang and Loy, Chen Change and Hou, Junhui and Kwong, Sam and Cong, Runmin},
  booktitle={Proceedings of the IEEE/CVF conference on computer vision and pattern recognition},
  pages={1780--1789},
  year={2020}
}

@article{patilUnpairedRecurrentLearning2025,
  title = {Unpaired Recurrent Learning for Real-World Video de-Hazing},
  author = {Patil, Prashant W. and Randive, Santosh Nagnath and Gupta, Sunil and Rana, Santu and Venkatesh, Svetha and Murala, Subrahmanyam},
  date = {2025-10-01},
  year = {{2025}},
  journal = {{Pattern Recognition}},
  shortjournal = {Pattern Recognition},
  volume = {166},
  pages = {111698},
  issn = {0031-3203},
  doi = {10.1016/j.patcog.2025.111698},
  url = {https://www.sciencedirect.com/science/article/pii/S0031320325003589},
  urldate = { 2025-05-26}
}

@article{lyu2024enhancing,
  title={Enhancing low-light light field images with a deep compensation unfolding network},
  author={Lyu, Xianqiang and Hou, Junhui},
  journal={IEEE Transactions on Image Processing},
  year={2024},
  publisher={IEEE}
}

@inproceedings{wu2022uretinex,
  title={Uretinex-net: Retinex-based deep unfolding network for low-light image enhancement},
  author={Wu, Wenhui and Weng, Jian and Zhang, Pingping and Wang, Xu and Yang, Wenhan and Jiang, Jianmin},
  booktitle={Proceedings of the IEEE/CVF conference on computer vision and pattern recognition},
  pages={5901--5910},
  year={2022}
}

@article{li2023progressive,
  title={Progressive deep video dehazing without explicit alignment estimation},
  author={Li, Runde and Chen, Lei},
  journal={Applied Intelligence},
  volume={53},
  number={10},
  pages={12437--12447},
  year={2023},
  publisher={Springer}
}

@inproceedings{shethUnsupervisedDeepVideo2021,
  title = {Unsupervised {{Deep Video Denoising}}},
  booktitle = {2021 IEEE/CVF International Conference on Computer Vision (ICCV)IEEE/CVF International Conference on Computer Vision (ICCV)},
  author = {Sheth, Dev Yashpal and Mohan, Sreyas and Vincent, Joshua L. and Manzorro, Ramon and Crozier, Peter A. and Khapra, Mitesh M. and Simoncelli, Eero P. and Fernandez-Granda, Carlos},
  date = {2021-10},
  year={{2021}},
  pages = {1739--1748},
  publisher = {IEEE},
  location = {Montreal, QC, Canada},
  doi = {10.1109/ICCV48922.2021.00178},
  url = {https://ieeexplore.ieee.org/document/9711126/},
  urldate = {2025-05-21},
  eventtitle = {2021 IEEE/CVF International Conference on Computer Vision (ICCV)IEEE/CVF International Conference on Computer Vision (ICCV)},
  isbn = {978-1-6654-2812-5},
  langid = {english}
}

@InProceedings{Varghese_2023_ICCV,
    author    = {Varghese, Nisha and Kumar, Ashish and Rajagopalan, A. N.},
    title     = {Self-supervised Monocular Underwater Depth Recovery, Image Restoration, and a Real-sea Video Dataset},
    booktitle = {Proceedings of the IEEE/CVF International Conference on Computer Vision (ICCV)},
    month     = {October},
    year      = {2023},
    pages     = {12248-12258}
}

@article{goodfellow2020generative,
  title={Generative adversarial networks},
  author={Goodfellow, Ian and Pouget-Abadie, Jean and Mirza, Mehdi and Xu, Bing and Warde-Farley, David and Ozair, Sherjil and Courville, Aaron and Bengio, Yoshua},
  journal={Communications of the ACM},
  volume={63},
  number={11},
  pages={139--144},
  year={2020},
  publisher={ACM New York, NY, USA}
}

@INPROCEEDINGS{7351548,
  author={Anantrasirichai, N. and Burn, J. and Bull, David R.},
  booktitle={2015 IEEE International Conference on Image Processing (ICIP)}, 
  title={Robust texture features based on undecimated dual-tree complex wavelets and local magnitude binary patterns}, 
  year={2015},
  volume={},
  number={},
  pages={3957-3961},
  doi={10.1109/ICIP.2015.7351548}}

@inproceedings{wangRecurrentSelfSupervisedVideo2023,
  title = {Recurrent {{Self-Supervised Video Denoising}} with {{Denser Receptive Field}}},
  booktitle = {Proceedings of the 31st {{ACM International Conference}} on {{Multimedia}}},
  author = {Wang, Zichun and Zhang, Yulun and Zhang, Debing and Fu, Ying},
  date = {2023-10-26},
  year = {{2023}},
  pages = {7363--7372},
  publisher = {ACM},
  location = {Ottawa ON Canada},
  doi = {10.1145/3581783.3612228},
  url = {https://dl.acm.org/doi/10.1145/3581783.3612228},
  urldate = {  2025-05-15},
  eventtitle = {{{MM}} '23: {{The}} 31st {{ACM International Conference}} on {{Multimedia}}},
  isbn = {979-8-4007-0108-5},
  langid = {english}
}

@inproceedings{chen2025spatiotemporal,
  title     = {Spatiotemporal Blind-Spot Network with Calibrated Flow Alignment for Self-Supervised Video Denoising},
  author    = {Chen, Zhen and Jiang, Tao and Hu, Xintao and Zhang, Wei and Li, Hao and Wang, Hongwei},
  booktitle = {Proceedings of the AAAI Conference on Artificial Intelligence},
  year      = {2025},
  volume    = {39},
  number    = {3},
  pages     = {2411--2419},
  doi       = {10.1609/aaai.v39i3.32242},
  url       = {https://doi.org/10.1609/aaai.v39i3.32242}
}

@inproceedings{Shrivastava2024Prior,
  author    = {Shrivastava, Gaurav and Lim, Ser-Nam and Shrivastava, Abhinav},
  title     = {Video Decomposition Prior: Editing Videos Layer by Layer},
  booktitle = {Proceedings of the International Conference on Learning Representations (ICLR)},
  year      = {2024}
}

@inproceedings{xuVideoDehazingMultiRange2023,
  title = {Video {{Dehazing}} via a {{Multi-Range Temporal Alignment Network}} with {{Physical Prior}}},
  booktitle = {2023  IEEE/CVF Conference on Computer Vision and  Pattern Recognition (CVPR)},
  author = {Xu, Jiaqi and Hu, Xiaowei and Zhu, Lei and Dou, Qi and Dai, Jifeng and Qiao, Yu and Heng, Pheng-Ann},
  date = {2023-06},
  year      = {{2023}},
  pages = {18053--18062},
  publisher = {IEEE},
  location = {Vancouver, BC, Canada},
  doi = {10.1109/CVPR52729.2023.01731},
  url = {https://ieeexplore.ieee.org/document/10203534/},
  urldate = {  2025-05-19},
  eventtitle = {2023  IEEE/CVF Conference on Computer Vision and  Pattern Recognition (CVPR)},
  isbn = {979-8-3503-0129-8},
  langid = {english}
}

@article{Zhou2023Review,
  author    = {Zhou, Jun and Yang, Tianyu and Zhang, Wei},
  title     = {Underwater Vision Enhancement Technologies: A Comprehensive Review, Challenges, and Recent Trends},
  journal   = {Applied Intelligence},
  volume    = {53},
  pages     = {3594--3621},
  year      = {2023},
  publisher = {Springer},
  doi       = {10.1007/s10489-022-03767-y},
  url       = {https://doi.org/10.1007/s10489-022-03767-y}
}

@INPROCEEDINGS{Lin2025Dynamic,
  author={Lin, Ruirui and Huang, Guoxi and Anantrasirichai, Nantheera},
  booktitle={2026 IEEE International Conference on Image Processing (ICIP)}, 
  title={Dynamic Weight-Based Temporal Aggregation for Low-Light Video Enhancement Under Extreme Noise}, 
  year={2026},
  volume={},
  number={},
  pages={1-6},
  doi={10.1109/ICIP61757.2026.11630319}}

@online{yehDiffIR2VRZeroZeroShotVideo2025,
  title = {{{DiffIR2VR-Zero}}: {{Zero-Shot Video Restoration}} with {{Diffusion-based Image Restoration Models}}},
  shorttitle = {{{DiffIR2VR-Zero}}},
  author = {Yeh, Chang-Han and Lin, Chin-Yang and Wang, Zhixiang and Hsiao, Chi-Wei and Chen, Ting-Hsuan and Shiu, Hau-Shiang and Liu, Yu-Lun},
  date = {2025-03-25},
  year = {{2025}},
  eprint = {2407.01519},
  eprinttype = {arXiv},
  eprintclass = {cs},
  doi = {10.48550/arXiv.2407.01519},
  url = {http://arxiv.org/abs/2407.01519},
  urldate = { 2025-05-26},
  langid = {english},
  pubstate = {prepublished}
}

@inproceedings{Li2025TempRetinex,
  title   = {{TempRetinex}: Retinex-based Unsupervised Enhancement for Low-light Video Under Diverse Lighting Conditions},
  author  = {Li, Yini and Louis Forster and  David Bull and Anantrasirichai, Nantheera},
  booktitle = {IEEE International Conference on Multimedia and Expo (ICME)},
  year    = {2026}
}

@article{Lin:BVIRLV:2024,
  title={{BVI-RLV: A} Fully Registered Dataset and Benchmarks for Low-Light Video Enhancement},
  author={R. Lin and N. Anantrasirichai and G. Huang and J. Lin and Q. Sun and A. Malyugina and D. Bull},
  journal={arXiv:2401.10166},
  year={2024}
}

@inproceedings{Lin:Low:2024,
author = {R. Lin and Q. Sun and N. Anantrasirichai},
title = {Low-light Video Enhancement with Conditional Diffusion Models and Wavelet Interscale Attentions},
year = {2024},
doi = {10.1145/3697294.3697304},
booktitle = {ACM SIGGRAPH CVMP}
}

@InProceedings{Fu_2023_ICCV,
    author    = {H. Fu and W. Zheng and X. Wang and J. Wang and H. Zhang and H. Ma},
    title     = {Dancing in the Dark: A Benchmark towards General Low-light Video Enhancement},
    booktitle = {ICCV},
    month     = {},
    year      = {2023},
    pages     = {12877-12886}
}

@inproceedings{zhengSemanticGuidedZeroShotLearning2022,
  title = {Semantic-{{Guided Zero-Shot Learning}} for {{Low-Light Image}}/{{Video Enhancement}}},
  booktitle = {2022 IEEE/CVF Winter Conference on Applications of Computer Vision (WACV)},
  author = {Zheng, Shen and Gupta, Gaurav},
  date = {2022-01},
  year={{2022}},
  pages = {581--590},
  publisher = {IEEE},
  location = {Waikoloa, HI, USA},
  doi = {10.1109/WACVW54805.2022.00064},
  url = {https://ieeexplore.ieee.org/document/9707578/},
  urldate = {  2025-05-15},
  eventtitle = {2022 IEEE/CVF Winter Conference on Applications of Computer Vision (WACV)},
  isbn = {978-1-6654-5824-5},
  langid = {english}
}

@article{zhang2025Wakeup-Darkness,
  title={{Wakeup-Darkness}: When Multimodal Meets Unsupervised Low-light Image Enhancement},
  author={X. Zhang and Z. Xu and H. Tang and C. Gu and W. Chen},
  journal={TOMM},
  year={2025}
}

@InProceedings{NerCO_2023_ICCV,
    author    = {S. Yang and M. Ding and Y. Wu and Z. Li and J. Zhang},
    title     = {Implicit Neural Representation for Cooperative Low-light Image Enhancement},
    booktitle = {ICCV},
    month     = {October},
    year      = {2023},
    pages     = {12918-12927}
}

@InProceedings{Colie2024,
  title = {Fast Context-Based Low-Light Image Enhancement via Neural Implicit Representations},
  booktitle = {ECCV},
  author = {T. Chobola and Y. Liu and H. Zhang and J. Schnabel and T. Peng},
  year = {2024},
  pages = {413–430}
}

@article{jiang2021enlightengan,
  title={Enlightengan: Deep light enhancement without paired supervision},
  author={Y. Jiang and X. Gong and D. Liu and Y. Cheng and C. Fang and X. Shen and J. Yang and P. Zhou and Z. Wang},
  journal={TIP},
  volume={30},
  pages={2340--2349},
  year={2021},
  publisher={IEEE}
}

@inproceedings{liu2021ruas,
title = {Retinex-inspired Unrolling with Cooperative Prior Architecture Search for Low-light Image Enhancement},
author = {R. Liu and L. Ma and J. Zhang and X. Fan and Z. Luo},
booktitle = {CVPR},
year = {2021}
}

@inproceedings{retinexformer,
  title={Retinexformer: One-stage Retinex-based Transformer for Low-light Image Enhancement},
  author={Y. Cai and H. Bian and J. Lin and H. Wang and R. Timofte and Y. Zhang},
  booktitle={ICCV},
  year={2023}
}

@inproceedings{huang:bvi:2025,
title = "BVI-Mamba: Video Enhancement Using a Visual State-Space Model for Low-Light and Underwater Environments",
author = "G. Huang and R. Lin and Y. Li and D. Bull and N. Anantrasirichai",
year = "2025",
booktitle = "Machine Learning from Challenging Data 2025",
}

@inproceedings{NEURIPS2023_6ba85c6f,
 author = {Shrivastava, Gaurav and Lim, Ser Nam and Shrivastava, Abhinav},
 booktitle = {Advances in Neural Information Processing Systems},
 editor = {A. Oh and T. Naumann and A. Globerson and K. Saenko and M. Hardt and S. Levine},
 pages = {34228--34246},
 publisher = {Curran Associates, Inc.},
 title = {Video Dynamics Prior: An Internal Learning Approach for Robust Video Enhancements},
 url = {https://proceedings.neurips.cc/paper\_files/paper/2023/file/6ba85c6f1c7656a6a647bc4d63b90bf0-Paper-Conference.pdf},
 volume = {36},
 year = {2023}
}

@article{zhengUnsupervisedDeepVideo2023,
  title = {Unsupervised {{Deep Video Denoising}} with {{Untrained Network}}},
  author = {Zheng, Huan and Pang, Tongyao and Ji, Hui},
  date = {2023-06-26},
  year = {{2023}},
  journal = {Proceedings of the AAAI Conference on Artificial Intelligence},
  shortjournal = {AAAI},
  volume = {37},
  number = {3},
  pages = {3651--3659},
  issn = {2374-3468, 2159-5399},
  doi = {10.1609/aaai.v37i3.25476},
  url = {https://ojs.aaai.org/index.php/AAAI/article/view/25476},
  urldate = { 2025-05-26},
  langid = {english}
}

@inproceedings{Liang2022RVRT,
  title     = {Recurrent Video Restoration Transformer with Guided Deformable Attention},
  author    = {Liang, Jingyun and Fan, Yuchen and Xiang, Xiaoyu and Ranjan, Rakesh and Ilg, Eddy and Green, Simon and Cao, Jiezhang and Zhang, Kai and Timofte, Radu and Van Gool, Luc},
  booktitle = {Advances in Neural Information Processing Systems 35},
  year      = {2022},
  url       = {https://proceedings.neurips.cc/paper/2022/hash/02687e7b22abc64e651be8da74ec610e-Abstract.html}
}

@ARTICLE{10462902,
  author={Liang, Jingyun and Cao, Jiezhang and Fan, Yuchen and Zhang, Kai and Ranjan, Rakesh and Li, Yawei and Timofte, Radu and Van Gool, Luc},
  journal={IEEE Transactions on Image Processing}, 
  title={VRT: A Video Restoration Transformer}, 
  year={2024},
  volume={33},
  number={},
  pages={2171-2182},
  doi={10.1109/TIP.2024.3372454}}

@InProceedings{Vaksman_2021_ICCV,
    author    = {Vaksman, Gregory and Elad, Michael and Milanfar, Peyman},
    title     = {Patch Craft: Video Denoising by Deep Modeling and Patch Matching},
    booktitle = {Proceedings of the IEEE/CVF International Conference on Computer Vision (ICCV)},
    month     = {October},
    year      = {2021},
    pages     = {2157-2166}
}

@INPROCEEDINGS{10647922,
  author={Aiyetigbo, Mary and Ravichandran, Dineshchandar and Chalhoub, Reda and Kalivas, Peter and Luo, Feng and Li, Nianyi},
  booktitle={2024 IEEE International Conference on Image Processing (ICIP)}, 
  title={Unsupervised Coordinate-Based Video Denoising}, 
  year={2024},
  volume={},
  number={},
  pages={1438-1444},
  doi={10.1109/ICIP51287.2024.10647922}}

@INPROCEEDINGS{li2022unidirectional,
  title={Unidirectional Video Denoising by Mimicking Backward Recurrent Modules with Look-ahead Forward Ones},
  author={Li, Junyi and Wu, Xiaohe and Niu, Zhenxing and Zuo, Wangmeng},
  booktitle={ECCV},
  year={2022}
}

@article{pandey2025comprehensive,
  title={Comprehensive review of single image defogging techniques: enhancement, prior, and learning based approaches},
  author={Pandey, Pranav and Gupta, Rakesh and Goel, Nishant},
  journal={Artificial Intelligence Review},
  volume={58},
  number={116},
  year={2025},
  publisher={Springer},
  doi={10.1007/s10462-024-11034-4},
  url={https://doi.org/10.1007/s10462-024-11034-4}
}

@INPROCEEDINGS{SwinIR:2021,
  author={Liang, Jingyun and Cao, Jiezhang and Sun, Guolei and Zhang, Kai and Van Gool, Luc and Timofte, Radu},
  booktitle={2021 IEEE/CVF International Conference on Computer Vision Workshops (ICCVW)}, 
  title={{SwinIR}: Image Restoration Using Swin Transformer}, 
  year={2021},
  volume={},
  number={},
  pages={1833-1844},
  doi={10.1109/ICCVW54120.2021.00210}}

@INPROCEEDINGS{9156652,
  author={Tassano, Matias and Delon, Julie and Veit, Thomas},
  booktitle={2020 IEEE/CVF Conference on Computer Vision and Pattern Recognition (CVPR)}, 
  title={{FastDVDnet}: Towards Real-Time Deep Video Denoising Without Flow Estimation}, 
  year={2020},
  volume={},
  number={},
  pages={1351-1360},
  doi={10.1109/CVPR42600.2020.00143}}

@INPROCEEDINGS{10657840,
  author={Zhang, Huicong and Xie, Haozhe and Yao, Hongxun},
  booktitle={2024 IEEE/CVF Conference on Computer Vision and Pattern Recognition (CVPR)}, 
  title={Blur-Aware Spatio-Temporal Sparse Transformer for Video Deblurring}, 
  year={2024},
  volume={},
  number={},
  pages={3616-3626},
  doi={10.1109/CVPR52733.2024.00258}}

@inproceedings{li2023simple,
  title     = {A Simple Baseline for Video Restoration with Grouped Spatial-Temporal Shift},
  author    = {Li, Dasong and Shi, Xiaoyu and Zhang, Yi and Cheung, Ka Chun and See, Simon and Wang, Xiaogang and Qin, Hongwei and Li, Hongsheng},
  booktitle = {Proceedings of the IEEE/CVF Conference on Computer Vision and Pattern Recognition (CVPR)},
  year      = {2023}
}

@InProceedings{Chen_2024_CVPR,
    author    = {Chen, Lufei and Tian, Xiangpeng and Xiong, Shuhua and Lei, Yinjie and Ren, Chao},
    title     = {Unsupervised Blind Image Deblurring Based on Self-Enhancement},
    booktitle = {Proceedings of the IEEE/CVF Conference on Computer Vision and Pattern Recognition (CVPR)},
    month     = {June},
    year      = {2024},
    pages     = {25691-25700}
}

@INPROCEEDINGS{8237572,
  author={Yi, Zili and Zhang, Hao and Tan, Ping and Gong, Minglun},
  booktitle={2017 IEEE International Conference on Computer Vision (ICCV)}, 
  title={{DualGAN}: Unsupervised Dual Learning for Image-to-Image Translation}, 
  year={2017},
  volume={},
  number={},
  pages={2868-2876},
  doi={10.1109/ICCV.2017.310}}

@ARTICLE{10292884,
  author={Gui, Jie and Cong, Xiaofeng and He, Lei and Tang, Yuan Yan and Kwok, James Tin-Yau},
  journal={IEEE Transactions on Multimedia}, 
  title={Illumination Controllable Dehazing Network based on Unsupervised Retinex Embedding}, 
  year={2024},
  volume={26},
  number={},
  pages={4819-4830},
  doi={10.1109/TMM.2023.3326881}}

@article{Zhen2025Adaptive,
  author    = {Zhen, Wei and Li, Yanli and Xu, Guanyu and others},
  title     = {Adaptive Spatiotemporal Partitioning for Efficient Video Dehazing},
  journal   = {The Visual Computer},
  volume    = {41},
  pages     = {12055--12070},
  year      = {2025},
  publisher = {Springer},
  doi       = {10.1007/s00371-025-04144-9},
  url       = {https://doi.org/10.1007/s00371-025-04144-9}
}

@ARTICLE{10540131,
  author={Shukla, Ankit and Upadhyay, Avinash and Sharma, Manoj and Saini, Anil and Fatema, Nuzhat and Malik, Hasmat and Afthanorhan, Asyraf and Hossaini, Mohammad Asef},
  journal={IEEE Access}, 
  title={A Novel Zero-Shot Real World Spatio-Temporal Super-Resolution (ZS-RW-STSR) Model for Video Super-Resolution}, 
  year={2024},
  volume={12},
  number={},
  pages={123969-123984},
  doi={10.1109/ACCESS.2024.3406476}}

@ARTICLE{DUVD:2024,
  author={Yang, Yang and Guo, Chun-Le and Guo, Xiaojie},
  journal={IEEE Transactions on Image Processing}, 
  title={Depth-Aware Unpaired Video Dehazing}, 
  year={2024},
  volume={33},
  number={},
  pages={2388-2403},
  doi={10.1109/TIP.2024.3378472}}

@inproceedings{Deng2019DM2F,
  author    = {Deng, Zhe and Zhu, Lei and Hu, Xiaowei and others},
  title     = {Deep Multi-Model Fusion for Single-Image Dehazing},
  booktitle = {Proceedings of the IEEE/CVF International Conference on Computer Vision (ICCV)},
  year      = {2019},
  pages     = {2453--2462},
  publisher = {IEEE}
}

@inproceedings{Dong2020MSBDN,
  author    = {Dong, Hao and Pan, Jinshan and Xiang, Lei and others},
  title     = {Multi-Scale Boosted Dehazing Network with Dense Feature Fusion},
  booktitle = {Proceedings of the IEEE/CVF Conference on Computer Vision and Pattern Recognition (CVPR)},
  year      = {2020},
  pages     = {2157--2167},
  publisher = {IEEE}
}

@inproceedings{Guo2022Transformer,
  author    = {Guo, Chun-Lei and Yan, Qiang and Anwar, Salman and others},
  title     = {Image Dehazing Transformer with Transmission-Aware 3D Position Embedding},
  booktitle = {Proceedings of the IEEE/CVF Conference on Computer Vision and Pattern Recognition (CVPR)},
  year      = {2022},
  pages     = {5812--5820},
  publisher = {IEEE}
}

@article{Song2023ViT,
  author    = {Song, Yifan and He, Zhi and Qian, Haifeng and others},
  title     = {Vision Transformers for Single Image Dehazing},
  journal   = {IEEE Transactions on Image Processing},
  volume    = {32},
  pages     = {1927--1941},
  year      = {2023},
  publisher = {IEEE}
}

@inproceedings{Huang2022Neural,
  author    = {Huang, Chao and Li, Jian and Li, Bo and others},
  title     = {Neural Compression-Based Feature Learning for Video Restoration},
  booktitle = {Proceedings of the IEEE/CVF Conference on Computer Vision and Pattern Recognition (CVPR)},
  year      = {2022},
  pages     = {5872--5881},
  publisher = {IEEE}
}

@inproceedings{Liu2022Phase,
  author    = {Liu, Yifan and Wan, Lei and Fu, Hongbo and others},
  title     = {Phase-Based Memory Network for Video Dehazing},
  booktitle = {Proceedings of the 30th ACM International Conference on Multimedia (ACM MM)},
  year      = {2022},
  pages     = {5427--5435},
  publisher = {ACM}
}

@article{Ren2025TriplaneSmoothed,
  author    = {Ren, Jianrui and Chen, Hao and Ye, Tianyu and others},
  title     = {Triplane-Smoothed Video Dehazing with CLIP-Enhanced Generalization},
  journal   = {International Journal of Computer Vision},
  volume    = {133},
  pages     = {475--488},
  year      = {2025},
  publisher = {Springer},
  doi       = {10.1007/s11263-024-02161-0},
  url       = {https://doi.org/10.1007/s11263-024-02161-0}
}

@INPROCEEDINGS{10204475,
  author={Tang, Xiaole and Zhao, Xile and Liu, Jun and Wang, Jianli and Miao, Yuchun and Zeng, Tieyong},
  booktitle={2023 IEEE/CVF Conference on Computer Vision and Pattern Recognition (CVPR)}, 
  title={Uncertainty-Aware Unsupervised Image Deblurring with Deep Residual Prior}, 
  year={2023},
  volume={},
  number={},
  pages={9883-9892},
  doi={10.1109/CVPR52729.2023.00953}}

@ARTICLE{8935440,
  author={Lu, Boyu and Chen, Jun-Cheng and Chellappa, Rama},
  journal={IEEE Transactions on Biometrics, Behavior, and Identity Science}, 
  title={UID-GAN: Unsupervised Image Deblurring via Disentangled Representations}, 
  year={2020},
  volume={2},
  number={1},
  pages={26-39},
  doi={10.1109/TBIOM.2019.2959133}}

@inproceedings{tao2018srndeblur,
  title={Scale-recurrent Network for Deep Image Deblurring},
  author={Tao, Xin and Gao, Hongyun and Shen, Xiaoyong and Wang, Jue and Jia, Jiaya},
  booktitle={IEEE Conference on Computer Vision and Pattern Recognition (CVPR)},
  year={2018}
}

@inproceedings{Chen2022SimpleBaselines,
  title     = {Simple Baselines for Image Restoration},
  author    = {Chen, Liangyu and Chu, Xiaojie and Zhang, Xiangyu and Sun, Jian},
  booktitle = {Proceedings of the European Conference on Computer Vision (ECCV)},
  year      = {2022}
}

@inproceedings{FCLGAN:2022,
author = {Zhao, Suiyi and Zhang, Zhao and Hong, Richang and Xu, Mingliang and Yang, Yi and Wang, Meng},
title = {{FCL-GAN}: A Lightweight and Real-Time Baseline for Unsupervised Blind Image Deblurring},
year = {2022},
isbn = {9781450392037},
publisher = {Association for Computing Machinery},
address = {New York, NY, USA},
url = {https://doi.org/10.1145/3503161.3548113},
doi = {10.1145/3503161.3548113},
booktitle = {Proceedings of the 30th ACM International Conference on Multimedia},
pages = {6220–6229},
numpages = {10},
location = {Lisboa, Portugal},
series = {MM '22}
}

@inproceedings{PontTuset2017DAVIS,
  title     = {The 2017 {DAVIS} Challenge on Video Object Segmentation},
  author    = {Pont-Tuset, Jordi and Perazzi, Federico and Caelles, Sergi and Arbel{\'a}ez, Pablo and Sorkine-Hornung, Alexander and Van Gool, Luc},
  booktitle = {CVPR Workshops},
  pages     = {149--157},
  year      = {2017}
}

@INPROCEEDINGS{gu2023mamba,
  title={{Mamba: Linear}-time sequence modeling with selective state spaces},
  author={Gu, Albert and Dao, Tri},
  booktitle={Conference on Language Modeling},
  year={2024}
}

@incollection{zhuUnrolledDecomposedUnpaired2025,
  title = {Unrolled {{Decomposed Unpaired Learning}} for {{Controllable Low-Light Video Enhancement}}},
  booktitle = {Computer {{Vision}} – {{ECCV}} 2024},
  author = {Zhu, Lingyu and Yang, Wenhan and Chen, Baoliang and Zhu, Hanwei and Ni, Zhangkai and Mao, Qi and Wang, Shiqi},
  editor = {Leonardis, Aleš and Ricci, Elisa and Roth, Stefan and Russakovsky, Olga and Sattler, Torsten and Varol, Gül},
  date = {2025},
  year={{2025}},
  volume = {15081},
  pages = {329--347},
  publisher = {Springer Nature Switzerland},
  location = {Cham},
  doi = {10.1007/978-3-031-73337-6\_19},
  url = {https://link.springer.com/10.1007/978-3-031-73337-6\_19},
  urldate = {2025-05-19},
  isbn = {978-3-031-73336-9 978-3-031-73337-6},
  langid = {english}
}

@inproceedings{NEURIPS2022_95504595,
 author = {Kawar, Bahjat and Elad, Michael and Ermon, Stefano and Song, Jiaming},
 booktitle = {Advances in Neural Information Processing Systems},
 editor = {S. Koyejo and S. Mohamed and A. Agarwal and D. Belgrave and K. Cho and A. Oh},
 pages = {23593--23606},
 publisher = {Curran Associates, Inc.},
 title = {Denoising Diffusion Restoration Models},
 url = {\url{https://proceedings.neurips.cc/paper_files/paper/2022/file/95504595b6169131b6ed6cd72eb05616-Paper-Conference.pdf}},
 volume = {35},
 year = {2022}
}

@ARTICLE{Pan:exploiting:2022,
  author={Pan, Xingang and Zhan, Xiaohang and Dai, Bo and Lin, Dahua and Loy, Chen Change and Luo, Ping},
  journal={IEEE Transactions on Pattern Analysis and Machine Intelligence}, 
  title={Exploiting Deep Generative Prior for Versatile Image Restoration and Manipulation}, 
  year={2022},
  volume={44},
  number={11},
  pages={7474-7489},
  doi={10.1109/TPAMI.2021.3115428}}

@inproceedings{zhuUnpairedImageToImageTranslation2017,
  title = {Unpaired {{Image-To-Image Translation Using Cycle-Consistent Adversarial Networks}}},
  author = {Zhu, Jun-Yan and Park, Taesung and Isola, Phillip and Efros, Alexei A.},
  year = {{2017}},
  pages = {2223--2232},
  urldate = {2025-05-30},
  booktitle = {{Proceedings of IEEE International Conference on Computer Vision}},
  doi = {10.1109/ICCV.2017.244}
  
}

@inproceedings{ulyanov2018deep,
  title={Deep image prior},
  author={Ulyanov, Dmitry and Vedaldi, Andrea and Lempitsky, Victor},
  booktitle={Proceedings of the IEEE conference on computer vision and pattern recognition},
  pages={9446--9454},
  year={2018}
}

@inproceedings{lehtinen2018noise2noise,
  title={{Noise2Noise: Learning} image restoration without clean data},
  author={Lehtinen, Jaakko and Munkberg, Jacob and Hasselgren, Jon and Laine, Samuli and Karras, Tero and Aittala, Miika and Aila, Timo},
  booktitle={International Conference on Machine
Learning},
  year={2018}
}

@inproceedings{zamir2020cycleisp,
  title={Cycleisp: Real image restoration via improved data synthesis},
  author={Zamir, Syed Waqas and Arora, Aditya and Khan, Salman and Hayat, Munawar and Khan, Fahad Shahbaz and Yang, Ming-Hsuan and Shao, Ling},
  booktitle={Proceedings of the IEEE/CVF conference on computer vision and pattern recognition},
  pages={2696--2705},
  year={2020}
}

@inproceedings{zhou2022lednet,
  title={Lednet: Joint low-light enhancement and deblurring in the dark},
  author={Zhou, Shangchen and Li, Chongyi and Change Loy, Chen},
  booktitle={European conference on computer vision},
  pages={573--589},
  year={2022},
  organization={Springer}
}

@inproceedings{nah2019ntire,
  title={Ntire 2019 challenge on video deblurring and super-resolution: Dataset and study},
  author={Nah, Seungjun and Baik, Sungyong and Hong, Seokil and Moon, Gyeongsik and Son, Sanghyun and Timofte, Radu and Mu Lee, Kyoung},
  booktitle={Proceedings of the IEEE/CVF conference on computer vision and pattern recognition workshops},
  pages={0--0},
  year={2019}
}

@ARTICLE{10423039,
  author={Liang, Dan and Chu, Jiale and Cui, Yuguo and Zhai, Zhanhu and Wang, Dingcai},
  journal={IEEE Transactions on Geoscience and Remote Sensing}, 
  title={{NPT-UL}: An Underwater Image Enhancement Framework Based on Nonphysical Transformation and Unsupervised Learning}, 
  year={2024},
  volume={62},
  number={},
  pages={1-19},
  doi={10.1109/TGRS.2024.3363037}}

@article{An2024CycleGAN,
  author    = {An, Yongli and Zhang, Wenpeng and Ji, Zhanlin},
  title     = {Unsupervised Underwater Image Enhancement with Improved CycleGAN},
  journal   = {Engineering Research Express},
  volume    = {6},
  number    = {3},
  pages     = {035241},
  year      = {2024},
  publisher = {IOP Publishing},
  doi       = {10.1088/2631-8695/ad6f52},
  url       = {https://doi.org/10.1088/2631-8695/ad6f52}
}

@ARTICLE{10398257,
  author={Yin, Jiankai and Wang, Yan and Guan, Bowen and Zeng, Xianchao and Guo, Lei},
  journal={IEEE Transactions on Geoscience and Remote Sensing}, 
  title={Unsupervised Underwater Image Enhancement Based on Disentangled Representations via Double-Order Contrastive Loss}, 
  year={2024},
  volume={62},
  number={},
  pages={1-15},
  doi={10.1109/TGRS.2024.3353371}}

@InProceedings{Lin2026ELVIS,
    author    = {Lin, Joanne and Lin, Ruirui and Li, Yini and Bull, David and Anantrasirichai, Nantheera},
    title     = {{ELVIS}: Enhance Low-Light for Video Instance Segmentation in the Dark},
    booktitle = {Proceedings of the IEEE/CVF Conference on Computer Vision and Pattern Recognition (CVPR)},
    month     = {June},
    year      = {2026},
    pages     = {25011-25021}
}

@article{Huang2025Underwater,
  author  = {Huang, Guoxi and Wang, Haoran and Seymour, Brett and Kovacs, Evan and Ellerbrock, John and Blackham, Dave and Anantrasirichai, Nantheera},
  title   = {Visual Enhancement and 3D Representation for Underwater Scenes: A Review},
  journal = {Artificial Intelligence Review},
  year    = {2026},
  volume  = {59},
  number={206}
}

@article{wei2021eld,
  title={Physics-based noise modeling for extreme low-light photography},
  author={Wei, Kaixuan and Fu, Ying and Zheng, Yinqiang and Yang, Jiaolong},
  journal={IEEE Transactions on Pattern Analysis and Machine Intelligence},
  volume={44},
  number={11},
  pages={8520--8537},
  year={2021},
  publisher={IEEE}
}

@InProceedings{Zhang2023generalrawnoise,
    author    = {Zhang, Feng and Xu, Bin and Li, Zhiqiang and Liu, Xinran and Lu, Qingbo and Gao, Changxin and Sang, Nong},
    title     = {Towards General Low-Light Raw Noise Synthesis and Modeling},
    booktitle = {Proceedings of the IEEE/CVF International Conference on Computer Vision (ICCV)},
    month     = {October},
    year      = {2023},
    pages     = {10820-10830}
}

@inproceedings{li2025dualdn,
  title={Dualdn: Dual-domain denoising via differentiable isp},
  author={Li, Ruikang and Wang, Yujin and Chen, Shiqi and Zhang, Fan and Gu, Jinwei and Xue, Tianfan},
  booktitle={European Conference on Computer Vision},
  pages={160--177},
  year={2025},
  organization={Springer}
}

@InProceedings{zheng2024senmvae,
    author    = {Zheng, Dihan and Zou, Yihang and Zhang, Xiaowen and Bao, Chenglong},
    title     = {SeNM-VAE: Semi-Supervised Noise Modeling with Hierarchical Variational Autoencoder},
    booktitle = {Proceedings of the IEEE/CVF Conference on Computer Vision and Pattern Recognition (CVPR)},
    month     = {June},
    year      = {2024},
    pages     = {25889-25899}
}

@InProceedings{Kousha2022sRGBflow,
    author    = {Kousha, Shayan and Maleky, Ali and Brown, Michael S. and Brubaker, Marcus A.},
    title     = {Modeling sRGB Camera Noise With Normalizing Flows},
    booktitle = {Proceedings of the IEEE/CVF Conference on Computer Vision and Pattern Recognition (CVPR)},
    month     = {June},
    year      = {2022},
    pages     = {17463-17471}
}

@InProceedings{Fu2023sRGBncnoise,
    author    = {Fu, Zixuan and Guo, Lanqing and Wen, Bihan},
    title     = {sRGB Real Noise Synthesizing With Neighboring Correlation-Aware Noise Model},
    booktitle = {Proceedings of the IEEE/CVF Conference on Computer Vision and Pattern Recognition (CVPR)},
    month     = {June},
    year      = {2023},
    pages     = {1683-1691}
}

@InProceedings{Monakhova2022starlight,
    author    = {Monakhova, Kristina and Richter, Stephan R. and Waller, Laura and Koltun, Vladlen},
    title     = {Dancing Under the Stars: Video Denoising in Starlight},
    booktitle = {Proceedings of the IEEE/CVF Conference on Computer Vision and Pattern Recognition (CVPR)},
    month     = {June},
    year      = {2022},
    pages     = {16241-16251}
}

@InProceedings{lin2025generallowlight,
      title={Towards a General-Purpose Zero-Shot Synthetic Low-Light Image and Video Pipeline}, 
      author={Joanne Lin and Crispian Morris and Ruirui Lin and Fan Zhang and David Bull and Nantheera Anantrasirichai},
      year={2025},
      booktitle={the 3rd International Workshop on Multimedia Content Generation and Evaluation: New Methods and Practice},
      url={https://dl.acm.org/doi/10.1145/3746278.3759376}, 
}

@article{lv2021agllie,
 title={Attention guided low-light image enhancement with a large scale low-light simulation dataset},
 author={Lv, Feifan and Li, Yu and Lu, Feng},
 journal={International Journal of Computer Vision},
 volume={129},
 number={7},
 pages={2175--2193},
 year={2021}
}

@ARTICLE{yang2021lolv2synth,
  author={Yang, Wenhan and Wang, Wenjing and Huang, Haofeng and Wang, Shiqi and Liu, Jiaying},
  journal={IEEE Transactions on Image Processing}, 
  title={Sparse Gradient Regularized Deep Retinex Network for Robust Low-Light Image Enhancement}, 
  year={2021},
  volume={30},
  number={},
  pages={2072-2086},
  doi={10.1109/TIP.2021.3050850}}

@InProceedings{triantafyllidou2020sidgan,
author="Triantafyllidou, Danai
and Moran, Sean
and McDonagh, Steven
and Parisot, Sarah
and Slabaugh, Gregory",
editor="Vedaldi, Andrea
and Bischof, Horst
and Brox, Thomas
and Frahm, Jan-Michael",
title="Low Light Video Enhancement Using Synthetic Data Produced with an Intermediate Domain Mapping",
booktitle="Computer Vision -- ECCV 2020",
year="2020",
publisher="Springer International Publishing",
address="Cham",
pages="103--119",
isbn="978-3-030-58601-0"
}

@InProceedings{chen2019smid,
author = {Chen, Chen and Chen, Qifeng and Do, Minh N. and Koltun, Vladlen},
title = {Seeing Motion in the Dark},
booktitle = {Proceedings of the IEEE/CVF International Conference on Computer Vision (ICCV)},
month = {October},
year = {2019}
}

@ARTICLE{7744574,
  author={Chan, Stanley H. and Wang, Xiran and Elgendy, Omar A.},
  journal={IEEE Transactions on Computational Imaging}, 
  title={Plug-and-Play ADMM for Image Restoration: Fixed-Point Convergence and Applications}, 
  year={2017},
  volume={3},
  number={1},
  pages={84-98},
  doi={10.1109/TCI.2016.2629286}}

@ARTICLE{10004791,
  author={Kamilov, Ulugbek S. and Bouman, Charles A. and Buzzard, Gregery T. and Wohlberg, Brendt},
  journal={IEEE Signal Processing Magazine}, 
  title={Plug-and-Play Methods for Integrating Physical and Learned Models in Computational Imaging: Theory, algorithms, and applications}, 
  year={2023},
  volume={40},
  number={1},
  pages={85-97},
  doi={10.1109/MSP.2022.3199595}}

@inproceedings{ryu2019plug,
  title     = {Plug-and-Play Methods Provably Converge with Properly Trained Denoisers},
  author    = {Ryu, Ernest and Liu, Jingwei and Wang, Sunyoung and Chen, Xuejun and Wang, Zhaoran and Yin, Wotao},
  booktitle = {Proceedings of the 36th International Conference on Machine Learning (ICML)},
  year      = {2019},
  pages     = {5546-5557}
}

@ARTICLE{9454311,
  author={Zhang, Kai and Li, Yawei and Zuo, Wangmeng and Zhang, Lei and Van Gool, Luc and Timofte, Radu},
  journal={IEEE Transactions on Pattern Analysis and Machine Intelligence}, 
  title={Plug-and-Play Image Restoration With Deep Denoiser Prior}, 
  year={2022},
  volume={44},
  number={10},
  pages={6360-6376},
  doi={10.1109/TPAMI.2021.3088914}}

@ARTICLE{10401873,
  author={Zerva, Matina Ch. and Kondi, Lisimachos P.},
  journal={IEEE Access}, 
  title={Video Super-Resolution Using Plug-and-Play Priors}, 
  year={2024},
  volume={12},
  number={},
  pages={11963-11971},
  doi={10.1109/ACCESS.2024.3355195}}

@ARTICLE{7839189,
  author={Zhang, Kai and Zuo, Wangmeng and Chen, Yunjin and Meng, Deyu and Zhang, Lei},
  journal={IEEE Transactions on Image Processing}, 
  title={Beyond a Gaussian Denoiser: Residual Learning of Deep CNN for Image Denoising}, 
  year={2017},
  volume={26},
  number={7},
  pages={3142-3155},
  doi={10.1109/TIP.2017.2662206}}

@ARTICLE{9495194,
  author={Yuan, Xin and Liu, Yang and Suo, Jinli and Durand, Frédo and Dai, Qionghai},
  journal={IEEE Transactions on Pattern Analysis and Machine Intelligence}, 
  title={Plug-and-Play Algorithms for Video Snapshot Compressive Imaging}, 
  year={2022},
  volume={44},
  number={10},
  pages={7093-7111},
  doi={10.1109/TPAMI.2021.3099035}}

@ARTICLE{9913822,
  author={Nair, Pravin and Chaudhury, Kunal N.},
  journal={IEEE Transactions on Image Processing}, 
  title={Plug-and-Play Regularization Using Linear Solvers}, 
  year={2022},
  volume={31},
  number={},
  pages={6344-6355},
  doi={10.1109/TIP.2022.3211473}}

@inproceedings{10.1145/2964284.2967188,
author = {Guo, Xiaojie},
title = {LIME: A Method for Low-light IMage Enhancement},
year = {2016},
isbn = {9781450336031},
publisher = {Association for Computing Machinery},
address = {New York, NY, USA},
url = {https://doi.org/10.1145/2964284.2967188},
doi = {10.1145/2964284.2967188},
booktitle = {Proceedings of the 24th ACM International Conference on Multimedia},
pages = {87–91},
numpages = {5},
location = {Amsterdam, The Netherlands},
series = {MM '16}
}

@article{article,
author = {Reddy, Sai},
year = {2024},
month = {05},
pages = {1-5},
title = {Advancements in Video Deblurring: A Comprehensive Review},
volume = {08},
journal = {INTERANTIONAL JOURNAL OF SCIENTIFIC RESEARCH IN ENGINEERING AND MANAGEMENT},
doi = {10.55041/IJSREM32759}
}

@article{mittal2013making,
  title     = {Making a Completely Blind Image Quality Analyzer},
  author    = {Mittal, Anish and Soundararajan, Rajiv and Bovik, Alan C.},
  journal   = {IEEE Signal Processing Letters},
  volume    = {22},
  number    = {3},
  pages     = {209--212},
  year      = {2013},
  month     = {March},
  doi       = {10.1109/LSP.2013.2295092}
}

@InProceedings{Lai_2018_ECCV,
author = {Lai, Wei-Sheng and Huang, Jia-Bin and Wang, Oliver and Shechtman, Eli and Yumer, Ersin and Yang, Ming-Hsuan},
title = {Learning Blind Video Temporal Consistency},
booktitle = {Proceedings of the European Conference on Computer Vision (ECCV)},
month = {September},
year = {2018}
}

@inproceedings{wang2022exploring,
    author = {Wang, Jianyi and Chan, Kelvin CK and Loy, Chen Change},
    title = {Exploring CLIP for Assessing the Look and Feel of Images},
    booktitle = {AAAI},
    year = {2023}
}

@inproceedings{venkatanath2015blind,
  title     = {Blind Image Quality Evaluation Using Perception Based Features},
  author    = {Venkatanath, N. and Praneeth, D. and Chandrasekhar, Bh. M. and Channappayya, S. S. and Medasani, S. S.},
  booktitle = {Proceedings of the 21st National Conference on Communications (NCC)},
  year      = {2015},
  publisher = {IEEE},
  address   = {Piscataway, NJ}
}

@INPROCEEDINGS{Saad:Blind:2012,
  author={Saad, Michele A. and Bovik, Alan C.},
  booktitle={2012 Conference Record of the Forty Sixth Asilomar Conference on Signals, Systems and Computers (ASILOMAR)}, 
  title={Blind quality assessment of videos using a model of natural scene statistics and motion coherency}, 
  year={2012},
  volume={},
  number={},
  pages={332-336},
  doi={10.1109/ACSSC.2012.6489018}}

@INPROCEEDINGS{BRISQUE,
  author={Mittal, Anish and Moorthy, Anush K. and Bovik, Alan C.},
  booktitle={2011 Conference Record of the Forty Fifth Asilomar Conference on Signals, Systems and Computers (ASILOMAR)}, 
  title={Blind/Referenceless Image Spatial Quality Evaluator}, 
  year={2011},
  volume={},
  number={},
  pages={723-727},
  doi={10.1109/ACSSC.2011.6190099}}

@InProceedings{Poirier_2023_CVPR,
    author    = {Poirier-Ginter, Yohan and Lalonde, Jean-Fran\c{c}ois},
    title     = {Robust Unsupervised StyleGAN Image Restoration},
    booktitle = {Proceedings of the IEEE/CVF Conference on Computer Vision and Pattern Recognition (CVPR)},
    month     = {June},
    year      = {2023},
    pages     = {22292-22301}
}

@inproceedings{NEURIPS2022_bc12914d,
 author = {Zhang, Zhilu and Xu, RongJian and Liu, Ming and Yan, Zifei and Zuo, Wangmeng},
 booktitle = {Advances in Neural Information Processing Systems},
 editor = {S. Koyejo and S. Mohamed and A. Agarwal and D. Belgrave and K. Cho and A. Oh},
 pages = {29179--29191},
 publisher = {Curran Associates, Inc.},
 title = {Self-Supervised Image Restoration with Blurry and Noisy Pairs},
 url = {\url{https://proceedings.neurips.cc/paper_files/paper/2022/file/bc12914d66b41b6bfc2d3a5decdb498b-Paper-Conference.pdf}},
 volume = {35},
 year = {2022}
}

@InProceedings{Safonov_2025_CVPR,
    author    = {Safonov, Nickolay and Bryntsev, Alexey and Moskalenko, Andrey and Kulikov, Dmitry and Vatolin, Dmitriy and Timofte, Radu and Lei, Haibo and Gao, Qifan and Luo, Qing and Li, Yaqing and Song, Jie and Hao, Shaozhe and Zheng, Meisong and Xu, Jingyi and Wu, Chengbin and Liu, Jiahui and Chen, Ying and Deng, Xin and Xu, Mai and Liang, Peipei and Ma, Jie and Jin, Junjie and Pang, Yingxue and Luo, Fangzhou and Chen, Kai and Zhao, Shijie and Wu, Mingyang and Li, Renjie and Zuo, Yushen and Tu, Zhengzhong and Zhong, Shengyun},
    title     = {NTIRE 2025 Challenge on UGC Video Enhancement: Methods and Results},
    booktitle = {Proceedings of the Computer Vision and Pattern Recognition Conference (CVPR) Workshops},
    month     = {June},
    year      = {2025},
    pages     = {1503-1513}
}

\newpage
\begin{IEEEbiographynophoto}{Alexandra Malyugina}
completed a PhD in Computer Science from the University of Bristol and is currently a researcher in the Visual Information Laboratory at the University of Bristol. Currently she works as a Senior Research Associate in the School of Computer Science at the University of Bristol. Her research interests include image and video denoising and enhancement in challenging visual conditions, and the application of topological data analysis techniques in image processing. 
\end{IEEEbiographynophoto}

\begin{IEEEbiographynophoto}{Yini Li} 
received the BSc degree in Information Countermeasures Technology from Beijing Institute of Technology and the M.Eng. degree in Electronics and Communication from University of Chinese Academy of Sciences. She is currently pursuing the Ph.D. degree in Computer Science at University of Bristol. Her research interests include low-light video enhancement, computational photography, and video processing.
\end{IEEEbiographynophoto}

\begin{IEEEbiographynophoto}{Joanne Lin}
completed her BSc in Computer Science at the University of Bristol and is now pursuing a PhD at the Visual Information Laboratory, University of Bristol. Her PhD focuses on segmentation and tracking in low-light videos. She also has research interests in image/video synthesis, image/video enhancement and machine perception tasks in challenging conditions.
\end{IEEEbiographynophoto}

\begin{IEEEbiographynophoto}{Nantheera Anantrasirichai} (S’04-M’07) received her Ph.D. in Electrical and Electronic Engineering from the University of Bristol. She is currently a Professor in the School of Computer Science at the University of Bristol. Her research interests include image processing and computer vision with challenging data.
\end{IEEEbiographynophoto}



\EOD

\end{document}